\documentclass{article} 
\usepackage{iclr2027_conference,times}

\usepackage{amsmath,amsfonts,bm}

\def\eqref#1{equation~\ref{#1}}

\def\1{\bm{1}}

\DeclareMathAlphabet{\mathsfit}{\encodingdefault}{\sfdefault}{m}{sl}
\SetMathAlphabet{\mathsfit}{bold}{\encodingdefault}{\sfdefault}{bx}{n}

\usepackage{hyperref}
\usepackage{url}

\usepackage[table]{xcolor}
\usepackage{booktabs}
\usepackage{tabularx}
\usepackage{graphicx}

\usepackage{subcaption}

\usepackage{amssymb}
\usepackage{siunitx}
\usepackage{algorithm}
\usepackage{algorithmicx}
\usepackage{algpseudocode}
\title{Directed Temporal Representations for \mbox{Offline} Visual Control}

\author{%
\begin{tabular}[t]{@{}l@{\hspace{3em}}l@{}}
\textbf{Chenyang Yuan}$^{1}$ & \textbf{Haoyu Wang}$^{2}$ \\
\mdseries University of Sheffield & \mdseries Shanghai Jiao Tong University \\
\mdseries\texttt{cyuan7@sheffield.ac.uk} & \mdseries\texttt{create\_arc@sjtu.edu.cn} \\[0.8em]
\textbf{Zhuo Sun}$^{3,\dagger}$ & \textbf{Xiaoyuan Cheng}$^{4,\dagger}$ \\
\mdseries Shanghai University of Finance and Economics & \mdseries University College London \\
\mdseries\texttt{zhuosunreid@outlook.com} & \mdseries\texttt{ucesxc4@ucl.ac.uk}
\end{tabular}}

\iclrfinalcopy

\begin{document}

\maketitle
\begingroup
\renewcommand{\thefootnote}{\fnsymbol{footnote}}
\footnotetext[2]{Corresponding authors.}
\endgroup
\begin{abstract}

Predictive world models provide compact visual representations for control. Control requires a latent geometry aligned with temporal reachability rather than predictive similarity alone. We introduce Directed Temporal Representations for Control (DTRC), which learns such a geometry from offline visual trajectories on top of frozen LeWorldModel (LeWM) features. DTRC constructs a directed temporal quasimetric over the learned control representation. Short-range temporal offsets calibrate the distance scale. Bootstrapped targets extend temporal reachability across longer horizons. Action-conditioned consistency aligns the representation with local transition dynamics. The resulting distance estimates temporal reaching cost, and its change across a transition defines goal-relative temporal progress. We use this progress signal as a temporal critic for direct goal-conditioned policy learning. Model-assisted targets provide an additional training-time refinement under behavior-support and dynamics-agreement constraints. Across ten visual control tasks, DTRC achieves strong
goal-conditioned control performance relative to planning and
direct-policy baselines.
Held-out diagnostics on the four LeWM tasks show consistent
short-range temporal calibration, task-dependent long-range and
directional structure, and positive transition-level progress.
Temporal supervision improves the same flow-policy parameterization
across all four LeWM tasks, while the resulting policy acts directly
without iterative trajectory search at test time. Our project website provides code and visualizations:
\url{https://chenyang-yuan.github.io/dtrc-project/}.
\end{abstract}

\section{Introduction}

Latent world models compress visual observations into predictive
representations and learn action-conditioned dynamics in latent space,
enabling control through imagined rollouts
\citep{hafner2019planet,hafner2020dreamer,hansen2022tdmpc,hansen2024tdmpc2}.
Visual goal-reaching planners evaluate candidate actions using
distances between predicted latent states and goal representations
\citep{zhou2024dinowm,maes2026lewm}.
Latent geometry therefore directly influences the control objective.
Recent work reshapes latent spaces to align Euclidean distance with
feasible transition structure and improve planning
\citep{wang2026temporalstraighteninglatentplanning}.
Temporal reachability provides a complementary control structure based
on ordered state--goal relations.
These relations are associated with interaction cost and may be
asymmetric across opposite transition directions
\citep{wang2023qrl}.
These properties motivate a control representation aligned with
temporal reachability rather than predictive similarity alone.

Offline trajectories provide direct supervision for this temporal
structure.
Future states serve as achieved goals for learning from offline
experience
\citep{andrychowicz2018hindsightexperiencereplay,
ghosh2020learningreachgoalsiterated}.
Temporal relations between states have also been used to construct
goal-conditioned representations and value structure
\citep{eysenbach2022contrastive,
park2024hiqlofflinegoalconditionedrl,
park2024foundationpolicieshilbertrepresentations,
wang2023qrl}.
For control, temporal reachability requires local scale, directional
structure, and long-horizon consistency.
Short-range temporal offsets establish the local distance scale.
Ordered state--goal relations preserve directional information.
Multistep relations extend temporal reachability across longer horizons
\citep{zheng2026mqe}.
Changes in temporal cost across individual transitions provide a local
measure of progress.
A control representation should preserve these properties in a common
temporal geometry.

We introduce Directed Temporal Representations for Control (DTRC) to
learn this geometry from offline visual trajectories.
Starting from frozen LeWorldModel (LeWM) features
\citep{maes2026lewm}, DTRC learns a control representation equipped
with a directed temporal quasimetric.
A symmetric metric component defines the base geometry.
An order-sensitive residual introduces directional variation in
state--goal reaching cost 
\citep{liu2023metricresidualnetworkssample}.
Short-range temporal offsets calibrate the distance scale.
Bootstrapped targets extend temporal reachability across longer
horizons.
Action-conditioned consistency aligns the representation with local
transition dynamics.
The resulting distance estimates temporal reaching cost.
Its reduction across a transition defines goal-relative temporal
progress.

The directed temporal distance provides a critic for policy learning.
Advantage-weighted methods use value estimates to weight recorded
actions during policy learning
\citep{peng2019awr,kostrikov2022offline}.
DTRC uses goal-relative temporal progress for the same purpose.
Recorded actions receive larger weights for transitions with greater
temporal progress.
Policy-generated actions receive the same temporal evaluation through
their predicted successor states.
Additional model-assisted targets are restricted by behavior support
and dynamics agreement
\citep{yu2020mopo,yu2021combo,sikchi2022learning}.
Only supported candidates with lower predicted temporal cost than the
recorded continuation contribute to the additional policy objective.
Model-assisted targets therefore provide an additional training-time
refinement.
At test time, the goal-conditioned policy acts directly without
iterative trajectory search.

We evaluate DTRC on ten visual goal-conditioned control tasks spanning
navigation, reaching, and manipulation.
We first evaluate end-to-end goal-conditioned control performance.
We then study temporal structure beyond the directly supervised local
range, including task-dependent directionality and transition-level
progress.
Controlled comparisons isolate the effect of temporal supervision on
the same flow-policy parameterization.
Finally, qualitative visualizations examine the organization of the
learned control representation across tasks.
Our main contributions are threefold:
\begin{itemize}
    \item We introduce DTRC, a framework for learning a directed
    temporal control representation on frozen predictive visual
    features. The representation uses a directed quasimetric to 
    represent ordered state--goal relations through temporal reaching cost.

    \item We develop an offline learning objective for temporal
    representation learning. Short-range trajectory offsets establish
    the distance scale, bootstrapped targets propagate temporal
    reachability across longer horizons, and action-conditioned 
    consistency aligns the representation with local transition dynamics. 
    Distance reduction then defines goal-relative temporal progress for policy supervision.

    \item Across ten visual goal-conditioned control tasks, DTRC achieves
    strong performance relative to predictive-model planning and
    goal-conditioned policy baselines.
    Held-out diagnostics on the four LeWM tasks show consistent local
    temporal calibration, task-dependent long-range and directional
    structure, and positive transition-level progress.
    Temporal supervision further improves the same flow-policy
    parameterization across four LeWM tasks.
\end{itemize}

\section{Preliminaries}
\label{sec:preliminaries}

\paragraph{Offline goal-conditioned control.}
We consider finite-horizon goal-conditioned control from visual observations. At time $t$, the agent observes $o_t \in \mathcal{O}$, aims to reach a goal observation $o_g \in \mathcal{O}$ within a remaining horizon $h_t \in \{1,\ldots,H\}$, and selects an action $a_t \in \mathcal{A}$. The learning setting provides a fixed offline dataset $\mathcal{D}$ of unlabeled raw-pixel trajectories, where each trajectory is given by $\tau=(o_0,a_0,o_1,a_1,\ldots,o_T)$. The goal-conditioned control problem is to learn a policy
\begin{equation}
    \pi(a_t \mid o_t,o_g,h_t)
\end{equation}
that maps the current observation, goal observation, and remaining horizon to actions 
that progressively steer the system toward the goal. At test time, control is closed-loop: 
after executing $a_t$, the agent receives the next observation and recomputes its action 
conditioned on the updated state and the same goal.

\paragraph{Frozen predictive visual features.}
We encode the current and goal observations using a pretrained LeWM
encoder $E$~\citep{maes2026lewm},
\begin{equation}
    z_t = E(o_t),
    \qquad
    z_g = E(o_g),
\end{equation}
where $z_t,z_g\in\mathbb{R}^{d_z}$ denote predictive visual features.
LeWM is a joint-embedding predictive world model that supports
action-conditioned prediction and latent-space planning without pixel
reconstruction.
We keep the pretrained encoder fixed throughout training.
DTRC learns the control representation and temporal geometry on top of
these frozen features.

\section{Directed Temporal Representations for Control}
\label{sec:method}

\begin{figure*}[ht]
    \centering
    \includegraphics[width=\textwidth]
    {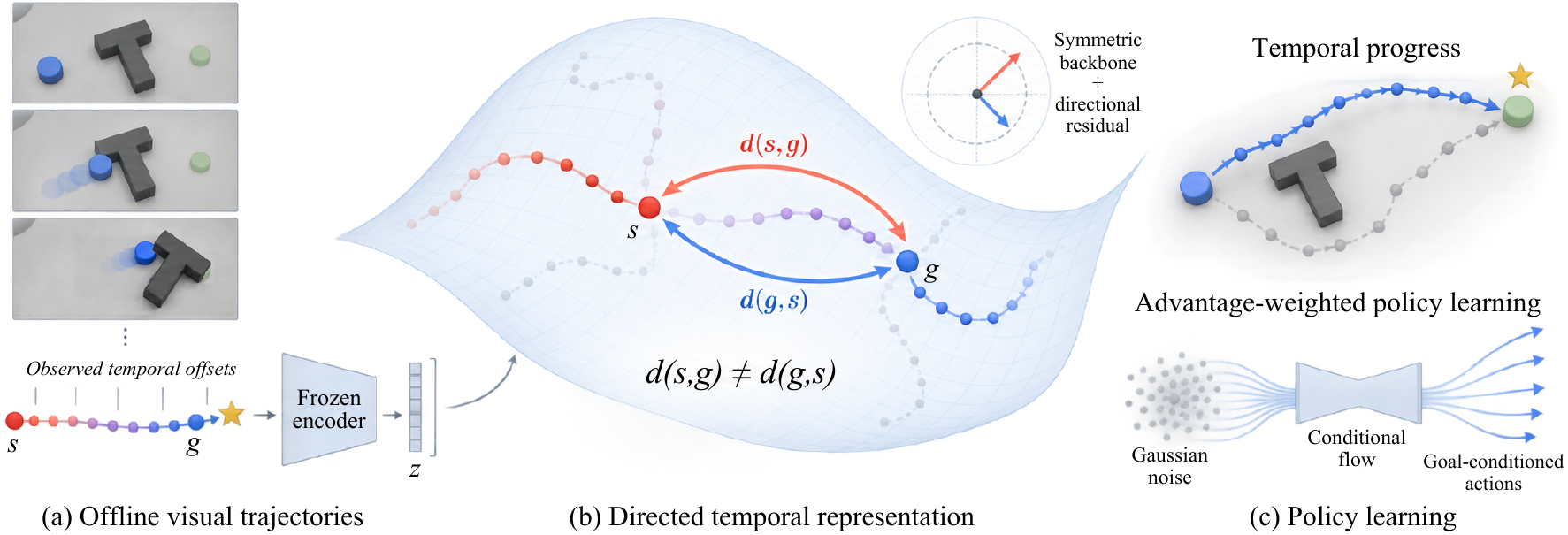}
    \caption{
    \textbf{Overview of directed temporal representation learning and
    policy training.}
    \textbf{(a)} Offline visual trajectories provide temporal supervision
    on top of frozen LeWM features.
    \textbf{(b)} DTRC learns a control representation with a directed
    temporal quasimetric. A symmetric metric component defines the base
    geometry, and an order-sensitive residual represents directional
    state--goal cost.
    \textbf{(c)} Temporal distance reduction measures goal-relative
    progress. The resulting temporal advantage weights recorded
    continuations for direct policy learning.
    }
    \label{fig:method_overview}
\end{figure*}

We introduce Directed Temporal Representations for Control (DTRC),
an offline framework for visual goal-conditioned control.
DTRC learns a directed temporal geometry on top of frozen predictive
visual features.
The method contains three stages.
First, we learn a control representation equipped with a directed
temporal quasimetric
(Section~\ref{sec:directed_temporal_representation}).
Second, offline trajectory supervision calibrates the local temporal
scale and propagates temporal reachability across longer horizons.
Action-conditioned consistency further aligns the representation with
local transition dynamics
(Section~\ref{sec:learning_temporal_geometry}).
Third, temporal distance reduction provides a goal-relative signal for
direct policy learning
(Section~\ref{sec:policy_learning}).
Model-assisted targets provide an additional training-time refinement.

\subsection{Directed Temporal Representation}
\label{sec:directed_temporal_representation}

Temporal reaching cost has directional and compositional structure.
Opposite transition directions can have different reaching costs.
A path through an intermediate state also constrains the direct
state--goal cost
\citep{wang2023qrl,liu2023metricresidualnetworkssample}.
We therefore represent temporal reachability with a directed
quasimetric.

Given a frozen LeWM feature $z$, a representation head
$\psi:\mathbb{R}^{d_z}\rightarrow\mathbb{R}^{d_e}$ produces the control
representation
\begin{equation}
    e=\psi(z).
\end{equation}
A symmetric metric head
$S:\mathbb{R}^{d_e}\rightarrow\mathbb{R}^{d_\phi}$
defines the base geometry
\citep{park2024foundationpolicieshilbertrepresentations}.
We augment this geometry with an order-sensitive map
$A_\theta:\mathbb{R}^{d_e}\rightarrow\mathbb{R}^{K}$
using the metric--residual construction
\citep{liu2023metricresidualnetworkssample}.
For $e_t=\psi(z_t)$ and $e_g=\psi(z_g)$, the directed temporal
distance is
\begin{equation}
    d_{\theta}(e_t,e_g)
    =
    \alpha
    \left[
        \left\|S(e_t)-S(e_g)\right\|_2
        +
        \max_{j\in[K]}
        \left[
            A_{\theta,j}(e_g)-A_{\theta,j}(e_t)
        \right]_+
    \right],
    \label{eq:directed_quasimetric}
\end{equation}
where $[u]_+=\max(u,0)$,
$A_\theta:\mathbb{R}^{d_e}\rightarrow\mathbb{R}^{K}$ defines
the directed residual coordinates.
The first term is symmetric in the state--goal pair.
The residual depends on the ordering of the pair.
Reversing the state--goal order changes the residual and can therefore
produce
$d_\theta(e_t,e_g)\neq d_\theta(e_g,e_t)$.
The parameter $\alpha=\exp(\eta)>0$ is a learned positive scale.

The metric--residual construction also satisfies the triangle inequality.
For any intermediate representation $e_m$, the direct state--goal cost
is bounded by the cost of reaching $e_m$ and continuing from $e_m$ to
the goal:
\begin{equation}
    d_\theta(e_t,e_g)
    \leq
    d_\theta(e_t,e_m)
    +
    d_\theta(e_m,e_g).
    \label{eq:directed_triangle}
\end{equation}
This property gives the distance a compositional structure across
state--goal pairs while retaining directional asymmetry.
The following objectives assign temporal semantics to this structured
distance.

\subsection{Learning Temporal Geometry from Offline Trajectories}
\label{sec:learning_temporal_geometry}

The directed quasimetric defines the structural form of the temporal
geometry.
Offline trajectories assign temporal semantics to this geometry.
Future states provide achieved goals, and their recorded offsets provide
direct temporal supervision
\citep{andrychowicz2018hindsightexperiencereplay,
ghosh2020learningreachgoalsiterated}.
We first calibrate the local distance scale and then propagate temporal
reachability across longer horizons.

\paragraph{Local temporal calibration.}
Short-range state pairs provide direct temporal anchors.
For a recorded offset
$\Delta\in\{1,\ldots,J\}$, we supervise
$d_\theta(e_t,e_{t+\Delta})$ using the observed separation $\Delta$.
We minimize
\begin{equation}
    \mathcal L_{\mathrm{local}}
    =
    \mathbb E_{t,\Delta}
    \left[
        \ell_{\tau_e}^{2}
        \left(
            \Delta
            -
            d_\theta(e_t,e_{t+\Delta})
        \right)
    \right],
    \label{eq:local_temporal_loss}
\end{equation}
where $J$ denotes the largest directly supervised temporal offset.
This objective establishes the local temporal scale of the learned
distance.

\paragraph{Bootstrapped long-horizon targets.}
For a current representation $e_t$ and goal representation $e_g$,
let $\Delta_{t,g}$ denote their recorded temporal separation for
state--goal pairs from the same trajectory.
Direct temporal supervision is available for short offsets.
Longer-range and cross-trajectory goal pairs use a bootstrapped target.
We combine an observed $n$-step prefix with the estimated remaining
temporal cost:
\begin{equation}
    y_{t,g}
    =
    \begin{cases}
        \Delta_{t,g},
        & 0<\Delta_{t,g}\leq n,
        \\[3pt]
        n+d_{\bar\theta}(\bar e_{t+n},\bar e_g),
        & \text{otherwise}.
    \end{cases}
    \label{eq:temporal_bootstrap}
\end{equation}
Here $\bar e=\bar\psi(z)$ denotes the target representation and
$d_{\bar\theta}$ denotes the corresponding target temporal distance.
The first branch anchors the geometry to observed temporal
separation.
The second branch propagates the same scale to longer state--goal
relations by combining an observed prefix with the estimated
remaining cost.
The bootstrap horizon $n$ controls the length of the observed
prefix.

Multiple trajectories and bootstrapped targets can provide different
temporal costs for related state--goal pairs.
We use lower-expectile regression to emphasize lower costs supported by
the temporal targets
\citep{newey1987asymmetric,kostrikov2022offline}.
The goal-conditioned temporal objective is
\begin{equation}
    \mathcal L_{\mathrm{goal}}
    =
    \mathbb E
    \left[
        \ell_{\tau_e}^{2}
        \left(
            y_{t,g}
            -
            d_\theta(e_t,e_g)
        \right)
    \right],
    \label{eq:goal_temporal_loss}
\end{equation}
where
$\ell_{\tau_e}^{2}(u)
=
\left|\tau_e-\mathbf{1}(u<0)\right|u^2$
denotes the expectile loss and
$u=y_{t,g}-d_\theta(e_t,e_g)$ is the temporal prediction residual.
We use $\tau_e<\frac{1}{2}$.
This choice assigns greater weight to overestimation relative to the
temporal target and shifts the learned distance toward lower supported
costs.

\paragraph{Action-conditioned consistency.}
The temporal objectives constrain state--goal relations.
We additionally require the control representation to remain compatible
with local action-conditioned transitions.
A transition head $T_\xi$ predicts the next control representation
through the residual update
\begin{equation}
    \hat e_{t+1}
    =
    e_t
    +
    T_\xi(e_t,a_t).
    \label{eq:representation_transition}
\end{equation}
We optimize the representation and transition head with
\begin{equation}
    \mathcal L_{\mathrm{cons}}
    =
    \mathbb E
    \left[
        \left\|
            \hat e_{t+1}
            -
            \bar e_{t+1}
        \right\|_2^2
    \right],
    \label{eq:representation_consistency}
\end{equation}
where gradients are stopped through the target representation
$\bar e_{t+1}$.
This loss aligns the learned control representation with local
transition dynamics.
The transition head $T_\xi$ is used for representation learning.
It is separate from the latent dynamics ensemble used for
model-assisted policy augmentation.

\paragraph{Overall temporal objective.}
We jointly optimize the control representation, directed temporal
distance, and transition-consistency head with
\begin{equation}
    \mathcal L_{\mathrm{temp}}
    =
    \mathcal L_{\mathrm{goal}}
    +
    \lambda_{\mathrm{local}}
    \mathcal L_{\mathrm{local}}
    +
    \lambda_{\mathrm{cons}}
    \mathcal L_{\mathrm{cons}}.
    \label{eq:temporal_objective}
\end{equation}
The coefficients $\lambda_{\mathrm{local}}$ and
$\lambda_{\mathrm{cons}}$ control local temporal calibration and
action-conditioned consistency, respectively.

\subsection{Policy Learning from Temporal Progress}
\label{sec:policy_learning}

The learned temporal distance provides a goal-relative signal for
policy learning.
We first use recorded continuations to train a direct
goal-conditioned policy.
Model-assisted alternatives provide additional training targets.

\paragraph{Temporal progress and advantage.}
For a recorded $k$-step continuation from $t$ to $t+k$, we define
goal-relative temporal progress as the reduction in temporal distance:
\begin{equation}
    P_{t,g}^{(k)}
    =
    d_{\bar\theta}(\bar e_t,\bar e_g)
    -
    d_{\bar\theta}(\bar e_{t+k},\bar e_g).
    \label{eq:temporal_progress}
\end{equation}
Positive values indicate a reduction in the estimated remaining
temporal cost.
The recorded continuation incurs a temporal cost of $k$ transitions.
We therefore define its temporal advantage as
\begin{equation}
    \mathcal A_{t,g}^{\mathrm{data}}
    =
    P_{t,g}^{(k)}-k
    =
    d_{\bar\theta}(\bar e_t,\bar e_g)
    -
    \left[
        k+
        d_{\bar\theta}(\bar e_{t+k},\bar e_g)
    \right].
    \label{eq:data_temporal_advantage}
\end{equation}
The subtraction of $k$ accounts for the temporal cost of the recorded
action sequence.
A larger advantage indicates greater temporal progress relative to the
cost of the recorded continuation.

\paragraph{Advantage-weighted policy training.}
We parameterize the goal-conditioned policy with conditional flow
matching
\citep{lipman2023flowmatching,park2025fql}.
This parameterization supports multimodal action distributions under a
fixed state--goal condition.
For a $k$-step action sequence
$\mathbf a_t=(a_t,\ldots,a_{t+k-1})$, the policy condition is
\begin{equation}
    c_t
    =
    [z_t,z_g,e_t,e_g,\operatorname{Emb}(h_t/H)],
\end{equation}
where $\operatorname{Emb}$ denotes the horizon embedding.

The temporal advantage determines the contribution of each recorded
continuation:
\begin{equation}
    w_{t,g}^{\mathrm{data}}
    =
    \min
    \left\{
        w_{\max},
        \exp
        \left(
            \frac{
                \mathcal A_{t,g}^{\mathrm{data}}
            }{\beta}
        \right)
    \right\},
    \label{eq:data_policy_weight}
\end{equation}
where $\beta$ controls the selectivity of advantage weighting and
$w_{\max}$ limits large weights.
The recorded-data policy objective is
\begin{equation}
    \mathcal L_{\mathrm{policy}}^{\mathrm{data}}
    =
    \mathbb E
    \left[
        w_{t,g}^{\mathrm{data}}
        \,
        \ell_{\mathrm{FM}}
        \left(
            \omega;
            \mathbf a_t,
            c_t
        \right)
    \right],
    \label{eq:data_policy_objective}
\end{equation}
where $\ell_{\mathrm{FM}}$ denotes the standard conditional
flow-matching loss.

\paragraph{Model-assisted refinement.}
The same temporal distance can also evaluate policy-generated
alternatives through their predicted successor states.
We retain additional targets that satisfy the behavior-support and
dynamics-agreement criteria and improve the predicted temporal cost
relative to the recorded continuation.
The resulting model-assisted policy objective is denoted by
$\mathcal L_{\mathrm{policy}}^{\mathrm{im}}$.
We provide the candidate construction and filtering procedure in
Appendix~\ref{app:implementation}.

The total policy objective is
\begin{equation}
    \mathcal L_{\mathrm{policy}}
    =
    \mathcal L_{\mathrm{policy}}^{\mathrm{data}}
    +
    \lambda_{\mathrm{im}}
    \mathcal L_{\mathrm{policy}}^{\mathrm{im}},
    \label{eq:full_policy_objective}
\end{equation}
where $\lambda_{\mathrm{im}}$ controls the contribution of the
model-assisted targets.

\section{Experiments}
\label{sec:experiments}

\subsection{Experimental Setup}
\label{sec:experimental-setup}

\paragraph{Environments and datasets.}
We evaluate DTRC on ten visual goal-conditioned control tasks spanning
navigation, reaching, and manipulation. We use the four offline datasets
adopted by LeWM~\citep{maes2026lewm}: PushT, TwoRoom, Reacher, and OGBench-Cube. We additionally evaluate on six visual OGBench
tasks: PointMaze, Cube-Double, Scene, AntMaze-Medium,
HumanoidMaze-Medium, and Puzzle-3x3.
All methods receive current and goal images encoded by pretrained LeWM
encoders.

\paragraph{Baselines.}
We compare against LeWM~\citep{maes2026lewm},
goal-conditioned inverse dynamics model (GC-IDM)~\citep{nguyen2026latentgeometrysearchamortizing},
goal-conditioned behavioral cloning (GCBC)~\citep{ghosh2020learningreachgoalsiterated},
goal-conditioned IQL (GC-IQL)~\citep{kostrikov2022offline,ogbench_park2025},
and a goal-conditioned adaptation of Hilbert foundation policies
(HILP)~\citep{park2024foundationpolicieshilbertrepresentations}.
Baseline and implementation details are provided in
Appendices~\ref{app:baselines}--\ref{app:architectures_hyperparameters}.

\subsection{Goal-Conditioned Control Performance}
\label{sec:control_performance}

We first evaluate end-to-end goal-conditioned control performance.
The four LeWM tasks provide a direct comparison with predictive-model
planning and goal-conditioned policy baselines.
We then evaluate on six additional visual OGBench tasks to assess
performance across a broader set of navigation and manipulation
environments.
\begin{figure*}[t]
    \centering
    \includegraphics[width=\textwidth]
    {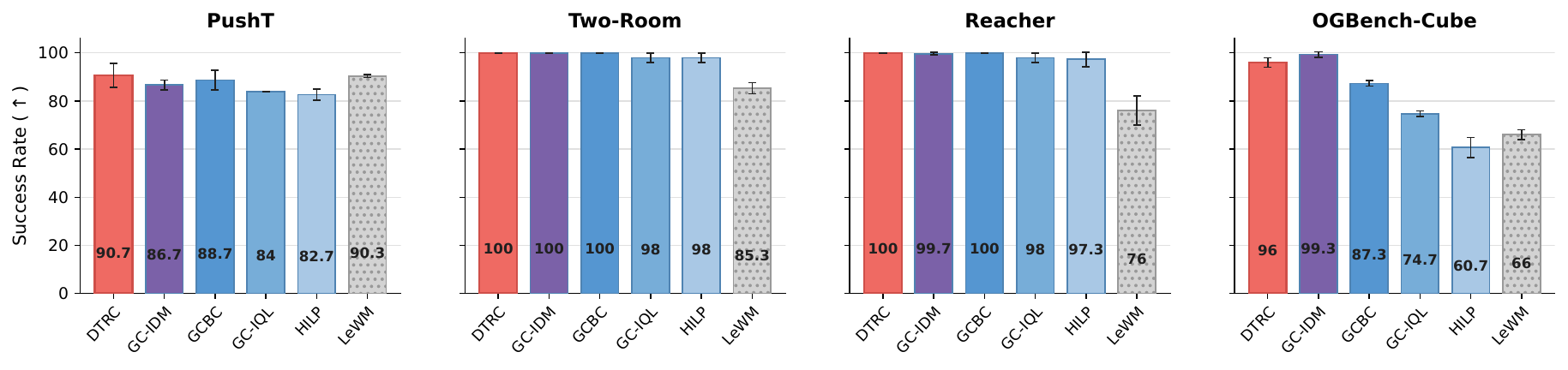}
    \caption{
        \textbf{Goal-conditioned control on the four LeWM tasks.}
        Bars report mean success rate over five evaluation seeds;
        error bars denote sample standard deviation.
    }
    \label{fig:lewm-four-task-comparison}
\end{figure*}
Figure~\ref{fig:lewm-four-task-comparison} compares DTRC with
predictive-model planning and direct-policy baselines on PushT,
TwoRoom, Reacher, and OGBench-Cube.
DTRC achieves the highest or tied-highest success on PushT, TwoRoom,
and Reacher, and the second-highest success on Cube.
Relative to LeWM, DTRC matches its performance on PushT and improves
success by $14.7$, $24.0$, and $30.0$ percentage points on TwoRoom,
Reacher, and Cube, respectively.
The results show that temporal representation learning can support
direct goal-conditioned control without test-time trajectory search.

\begin{table*}[t]
    \centering
    \caption{
        \textbf{Goal-conditioned control on six additional visual
        OGBench tasks.}
        Results report mean success rate (\%) $\pm$ sample standard
        deviation over five evaluation seeds.
        Best results are shown in bold and second-best results are
        underlined.
    }
    \label{tab:ogbench-six-methods}
    \setlength{\tabcolsep}{4.2pt}
    \resizebox{\textwidth}{!}{%
    \begin{tabular}{lcccccc}
        \toprule
        Method
        & PointMaze
        & Cube-Double
        & Scene
        & AntMaze-Medium
        & HumanoidMaze-Medium
        & Puzzle-3x3 \\
        \midrule

        LeWM
        & $13.3 \pm 1.2$
        & \underline{$50.7 \pm 2.3$}
        & \textbf{$56.7 \pm 6.4$}
        & \textbf{$32.0 \pm 6.0$}
        & $82.0 \pm 2.0$
        & $24.7 \pm 1.2$ \\

        GCBC
        & \textbf{$28.7 \pm 3.1$}
        & $44.7 \pm 2.3$
        & $48.7 \pm 5.0$
        & $28.0 \pm 3.5$
        & $83.3 \pm 3.1$
        & $35.3 \pm 3.1$ \\

        GC-IQL
        & $16.0 \pm 2.0$
        & $48.0 \pm 5.3$
        & $51.3 \pm 9.0$
        & $30.0 \pm 5.3$
        & $84.0 \pm 2.0$
        & $31.3 \pm 5.0$ \\

        HILP
        & $22.7 \pm 4.6$
        & $43.3 \pm 5.8$
        & $50.0 \pm 7.2$
        & $28.7 \pm 3.1$
        & $81.3 \pm 4.2$
        & \textbf{$40.0 \pm 6.9$} \\

        GC-IDM
        & \underline{$28.0 \pm 5.3$}
        & $38.7 \pm 4.6$
        & $48.7 \pm 10.3$
        & $30.0 \pm 3.5$
        & \textbf{$88.0 \pm 2.0$}
        & $33.3 \pm 2.3$ \\

        \rowcolor{blue!6}
        DTRC (ours)
        & $11.3 \pm 3.1$
        & \textbf{$54.7 \pm 8.3$}
        & \underline{$53.3 \pm 4.2$}
        & \underline{$31.3 \pm 4.6$}
        & \underline{$84.7 \pm 5.0$}
        & \textbf{$40.0 \pm 5.3$} \\

        \bottomrule
    \end{tabular}%
    }
\end{table*}

Table~\ref{tab:ogbench-six-methods} extends the comparison to six
additional visual OGBench tasks.
DTRC achieves the best or tied-best result on Cube-Double and
Puzzle-3x3, and the second-best result on Scene, AntMaze-Medium, and
HumanoidMaze-Medium.
Performance is less favorable on PointMaze, where DTRC reaches
$11.3\%$ success compared with $28.7\%$ for GCBC.
Overall, DTRC ranks among the top two methods on five of six tasks;
larger goal offsets are evaluated in
Appendix~\ref{app:goal_offset_results}.
We next examine the learned temporal structure and its contribution to
policy learning.

\subsection{Temporal Structure of the Learned Representation}
\label{sec:temporal_structure}

\begin{figure*}[t]
    \centering
    \includegraphics[width=\textwidth]
    {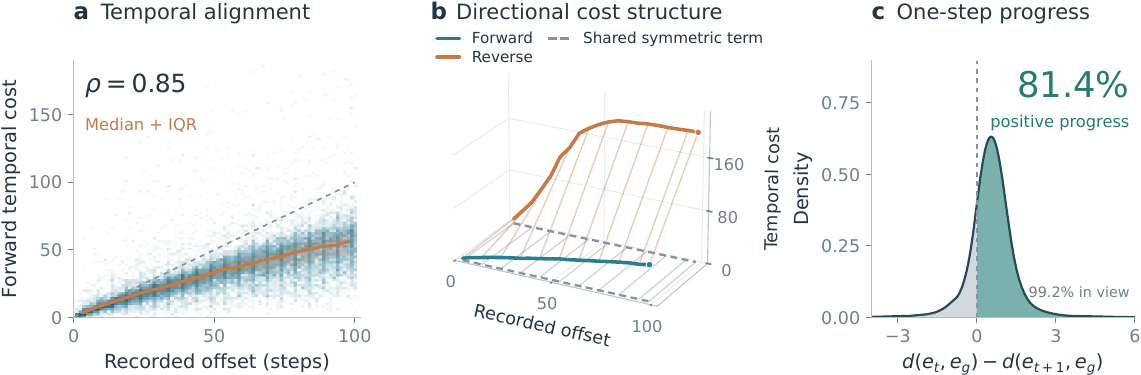}
    \caption{
        \textbf{Temporal structure learned by DTRC on PushT.}
        We evaluate 20,000 held-out state--future-goal pairs, with
        200 pairs at each temporal offset from 1 to 100 transitions.
        \textbf{(a)} Forward temporal distance as a function of recorded
        temporal offset. The solid curve and shaded region show the
        median and interquartile range in 5-step bins; the dashed line
        denotes equality with the recorded offset
        ($\rho=0.845$).
        \textbf{(b)} Forward distance $d(e_t,e_g)$ and reverse distance
        $d(e_g,e_t)$ for the same state--goal pairs. The gray curve
        shows the symmetric contribution of the same DTRC model.
        \textbf{(c)} Distribution of one-step temporal progress
        $P_{t,g}^{(1)}=d(e_t,e_g)-d(e_{t+1},e_g)$.
        Positive progress occurs for $81.41\%$ of the evaluated pairs.
    }
    \label{fig:pusht_temporal_diagnostics}
\end{figure*}

We first evaluate the learned temporal representation independently of
policy performance.
We use the EMA target network at 200,000 updates and sample
state--future-goal pairs from held-out episodes.
For each task, we draw 200 pairs at every temporal offset from 1 to
100 transitions, yielding 20,000 pairs.
All distances are computed in the full learned representation.
Recorded temporal offset is used as a reference for trajectory duration,
rather than as ground-truth optimal reaching cost. Figure~\ref{fig:pusht_temporal_diagnostics} illustrates three properties
on PushT: temporal structure across state--goal offsets, directional
state--goal cost, and transition-level temporal progress.

Figure~\ref{fig:pusht_temporal_diagnostics}a shows that the learned
forward distance retains substantial temporal structure beyond the
directly supervised local range, with Spearman correlation
$\rho=0.845$ over offsets from 1 to 100 transitions.
The median distance increases with recorded offset while deviating from
the equality reference at longer horizons.
This behavior is consistent with the learning objective: short-range
offsets calibrate the local distance scale, while bootstrapped targets
represent longer-horizon state--goal reaching cost rather than recorded
trajectory duration.
Figure~\ref{fig:pusht_temporal_diagnostics}b further shows pronounced
directional structure on PushT.
The reverse cost exceeds the forward cost for $92.76\%$ of the sampled
pairs, while the symmetric contribution is unchanged under reversal.
The asymmetry therefore arises from the order-sensitive component
(see Appendices~\ref{app:quasimetric_decomposition}
and~\ref{app:goal_relative_landscapes}).
Figure~\ref{fig:pusht_temporal_diagnostics}c evaluates the one-step
instance of temporal progress from Eq.~\ref{eq:temporal_progress}.
The learned distance decreases after the recorded transition for
$81.41\%$ of the sampled state--goal pairs.
This result shows that the learned geometry resolves local
goal-relative progress over the balanced offset range.
Temporal progress is subsequently converted into the temporal advantage
used for policy supervision.

We next summarize the same diagnostics across the four LeWM tasks.
Table~\ref{tab:temporal_structure} separates short-range temporal
calibration from longer-range, directional, and transition-level
properties of the learned geometry.
\begin{table}[t]
    \centering
    \caption{
        \textbf{Temporal representation diagnostics across the four
        LeWM tasks.}
        MAE$_{1:8}$ measures the mean absolute difference between
        forward temporal distance and recorded offset over temporal
        offsets 1--8.
        $\rho_{1:100}$ denotes Spearman's rank correlation between
        forward temporal distance and recorded offset over offsets
        1--100.
        Reverse $>$ Forward reports the fraction of state--goal pairs
        satisfying $d(e_g,e_t)>d(e_t,e_g)$.
        Positive progress reports the fraction satisfying
        $P_{t,g}^{(1)}>0$.
    }
    \label{tab:temporal_structure}
    \small
    \setlength{\tabcolsep}{4.5pt}
    \begin{tabular}{lcccc}
        \toprule
        Task
        & MAE$_{1:8}$
        & $\rho_{1:100}$
        & Reverse $>$ Forward
        & Positive progress \\
        & (steps) & & (\%) & (\%) \\
        \midrule
        PushT   & 1.55 & 0.845 & 92.76 & 81.41 \\
        TwoRoom & 1.51 & 0.800 & 57.65 & 65.78 \\
        Reacher & 1.62 & 0.510 & 50.28 & 59.41 \\
        Cube    & 0.74 & 0.221 & 52.46 & 72.35 \\
        \bottomrule
    \end{tabular}
\end{table}
Short-range MAE remains below two transitions on all four tasks, while
long-range association and directional structure vary by task
(Table~\ref{tab:temporal_structure}).
Positive one-step progress remains above $50\%$ on these tasks and above
$60\%$ on all six additional OGBench tasks
(Appendix~\ref{app:additional_temporal_diagnostics}).
Thus, DTRC resolves goal-relative transition progress across all ten
tasks without requiring a common long-range or directional geometry.

\subsection{Effect of Temporal Supervision on Policy Learning}
\label{sec:temporal_policy_effect}

The representation diagnostics above establish temporal structure in
the learned geometry.
We next evaluate whether this structure provides useful supervision for
policy learning.
We compare unweighted flow behavioral cloning with the same flow-policy
parameterization augmented by the learned DTRC control representation
and temporal-advantage weighting.
Model-assisted targets are disabled in this comparison.

\begin{table}[t]
    \centering
    \caption{
        \textbf{Effect of temporal supervision on policy learning.}
        Success rates (\%) are reported as mean $\pm$ sample standard
        deviation over five evaluation seeds.
        Both variants use the same flow-policy parameterization.
        Temporal supervision adds the learned DTRC control representation
        and temporal-advantage weighting without model-assisted targets.
        All variants use the same 50 start--goal pairs per task and seed,
        a goal offset of 25 recorded transitions, and a 50-step
        interaction budget.
    }
    \label{tab:temporal_policy_ablation}
    \small
    \setlength{\tabcolsep}{5pt}
    \begin{tabular}{lcccc}
        \toprule
        Configuration
        & PushT
        & TwoRoom
        & Reacher
        & Cube \\
        \midrule
        Flow BC
        & $74.7 \pm 4.6$
        & $94.7 \pm 4.2$
        & $68.0 \pm 3.5$
        & $91.3 \pm 1.2$ \\
        + temporal supervision
        & $84.0 \pm 5.3$
        & $100.0 \pm 0.0$
        & $100.0 \pm 0.0$
        & $97.3 \pm 3.1$ \\
        \bottomrule
    \end{tabular}
\end{table}

Temporal supervision improves the same flow-policy parameterization on
all four tasks
(Table~\ref{tab:temporal_policy_ablation}).
Success increases by $9.3$ percentage points on PushT, $5.3$ on
TwoRoom, $32.0$ on Reacher, and $6.0$ on Cube.
The largest change occurs on Reacher, where success increases from
$68.0\%$ to $100.0\%$.
Since model-assisted targets are disabled, this comparison isolates the
effect of the learned temporal representation and temporal-advantage
supervision from model-assisted policy refinement.
With the representation diagnostics, these results connect the learned
temporal geometry to policy learning.
Full component ablations, including behavior-support and
directional-residual variants, are reported in
Appendix~\ref{app:component_ablation}.

\subsection{Qualitative Structure of the Learned Representation}
\label{sec:qualitative_representation}
\begin{figure*}[t]
    \centering
    \includegraphics[width=\textwidth]{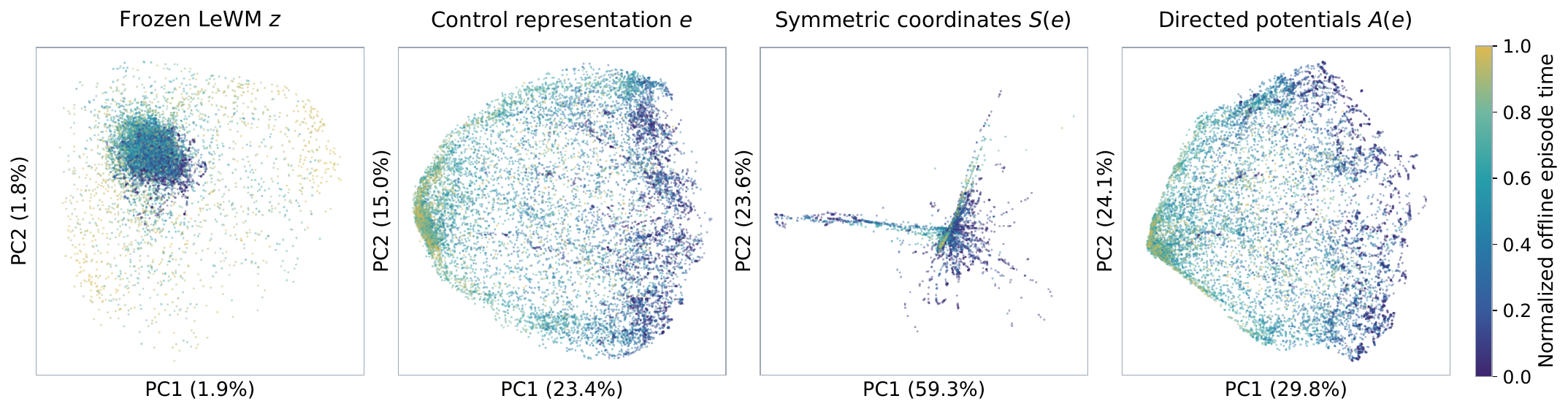}
    \caption{
        \textbf{Qualitative organization of the learned representation
        on PushT.}
        We project frozen LeWM features $z$, learned control
        representations $e$, symmetric coordinates $S(e)$, and directed
        potentials $A(e)$ onto their first two principal components.
        Points are colored by normalized time within the offline
        trajectory.
        The projections are used only for visualization; all temporal
        distances and quantitative diagnostics are computed in the full
        learned representation.
    }
    \label{fig:representation_spaces}
\end{figure*}
Figure~\ref{fig:representation_spaces} provides a qualitative view of
the transformation from frozen predictive features to the learned
temporal representation.
We visualize frozen LeWM features, the learned control representation,
the symmetric coordinates, and the directed potentials using their
leading principal components.
Temporal ordering is less apparent in the leading principal components
of the frozen LeWM features.
After DTRC training, the control representation exhibits a more
structured progression with offline episode time.
The symmetric coordinates and directed potentials further exhibit
distinct organizations, consistent with their separate roles in the
directed quasimetric.
These projections are illustrative; the temporal scale, directional
structure, and transition-level progress are evaluated quantitatively
in Section~\ref{sec:temporal_structure}.

\paragraph{Task-dependent geometry.}
We next visualize recorded trajectories in the learned control
representation across tasks.
\begin{figure*}[t]
    \centering
    \includegraphics[width=0.7\textwidth]
    {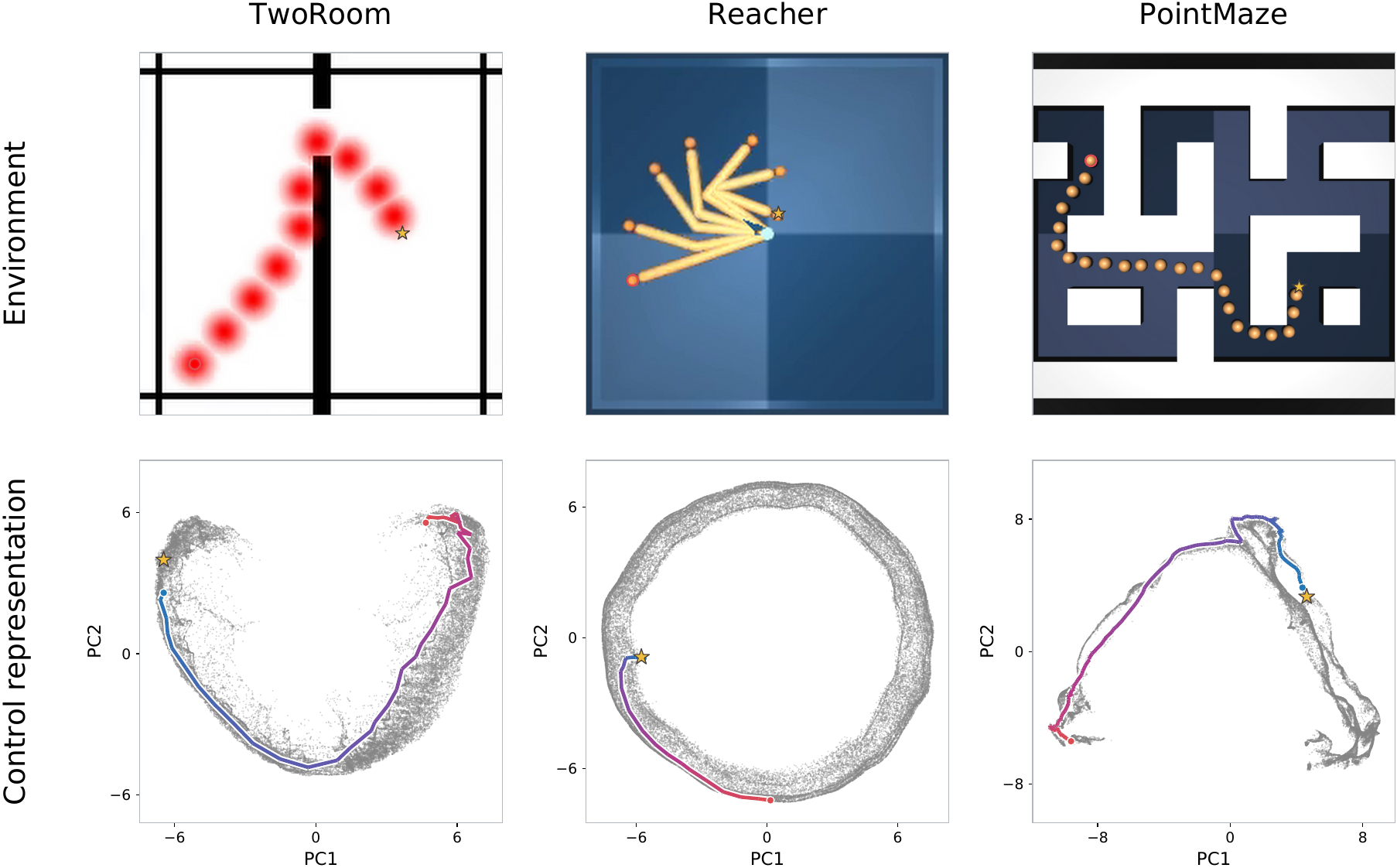}
    \caption{
        \textbf{Task-dependent geometry of the learned control
        representation.}
        Recorded trajectories are shown in the environment (top) and
        in the first two principal components of the DTRC control
        representation (bottom) for TwoRoom, Reacher, and PointMaze.
        Gray points show the offline representation distribution, and
        colored curves show recorded trajectories.
        Stars denote trajectory goals.
        The PCA projections are used only for qualitative visualization.
    }
    \label{fig:cross_task_geometry}
\end{figure*}
Figure~\ref{fig:cross_task_geometry} shows distinct geometric
organizations across control problems.
TwoRoom and PointMaze exhibit curved trajectory structure under
constrained navigation, while Reacher forms an approximately cyclic
organization in the PCA projection.
The learned trajectories therefore remain structured within
task-specific geometries rather than collapsing onto a common
one-dimensional temporal coordinate.
This variation is consistent with the task-dependent long-range and
directional statistics in Table~\ref{tab:temporal_structure}, and
illustrates why recorded trajectory duration need not induce the same
global ordering across tasks. 
Additional representation-space visualizations and their evolution
during training across all ten tasks are provided in
Appendices~\ref{app:representation_spaces}
and~\ref{app:training_dynamics}.

\section{Conclusion}

We introduced Directed Temporal Representations for Control (DTRC),
which learns a directed temporal geometry on top of frozen predictive
visual features.
The learned quasimetric represents goal-relative temporal reaching cost,
while distance reduction provides transition-level progress for direct
policy learning.
Across ten visual goal-conditioned control tasks, DTRC achieves strong
control performance and exhibits task-dependent long-range and
directional structure together with consistent local progress.
Controlled comparisons further show that temporal supervision improves
the same flow-policy parameterization across the four LeWM tasks.
These results support directed temporal geometry as an effective
interface between predictive visual representations and
goal-conditioned control.

\bibliography{iclr2027_conference}
\bibliographystyle{iclr2027_conference}

\clearpage
\appendix
\section{Baselines}
\label{app:baselines}

All baselines operate on LeWM features extracted from current and goal
images. The visual encoders are kept fixed during downstream training
and evaluation. Below, we describe the control models and the
adaptations used in our comparison.

\subsection{LeWM}
\label{app:baseline_lewm}

We use the pretrained LeWM encoder and action-conditioned predictor
with CEM-based model-predictive control~\citep{maes2026lewm}.
Candidate action sequences are evaluated by rolling out the predictor
in latent space and comparing the predicted outcomes with the encoded
goal. Control proceeds in a receding-horizon manner using updated
visual observations.
We retain the original latent-planning formulation without adding
the temporal distance or policy-learning components of DTRC.

\subsection{GC-IDM}
\label{app:baseline_gcidm}

We use the goal-conditioned inverse-dynamics policy of GC-IDM
\citep{nguyen2026latentgeometrysearchamortizing}.
The policy takes the current LeWM feature, goal feature, and remaining
horizon as inputs and predicts the recorded action through a
supervised regression objective.
We retain its MLP architecture and horizon-conditioning mechanism.
For the additional OGBench tasks, the action output dimension is
adjusted to the corresponding environment.
At evaluation time, the policy predicts actions directly from the
updated visual input without latent rollout or trajectory search.

\subsection{GCBC}
\label{app:baseline_gcbc}

GCBC learns a direct goal-conditioned policy by supervised imitation
\citep{ghosh2020learningreachgoalsiterated}.
Future observations from the same trajectory are sampled as goals,
and an MLP predicts the recorded action from the current feature,
goal feature, and remaining horizon.
Training minimizes the mean squared action-prediction error.
This baseline uses neither a value function nor advantage weighting.
It is distinct from the flow-matching behavioral-cloning variant
used in our component comparisons.

\subsection{GC-IQL}
\label{app:baseline_gciql}

GC-IQL learns a goal-conditioned action-value function
$Q(z_t,a_t,z_g)$ and value function $V(z_t,z_g)$
\citep{kostrikov2022offline,ogbench_park2025}.
The action-value function is trained by Bellman regression using a
target value network. The value function is trained by upper-expectile
regression toward the target action-value estimates for recorded
actions.
Goal relabeling supplies a reward of zero at the relabeled goal and
$-1$ otherwise, with bootstrapping terminated at the goal.
We first train the value networks and then extract a direct
goal-conditioned policy using the weighted regression procedure
described below.

\subsection{HILP}
\label{app:baseline_hilp}

We adapt the Hilbert value parameterization of HILP
\citep{park2024foundationpolicieshilbertrepresentations}
to frozen LeWM features:
\begin{equation}
    V(z_t,z_g)
    =
    -\left\|
        \phi_{\mathrm H}(z_t)-\phi_{\mathrm H}(z_g)
    \right\|_2,
    \label{eq:baseline_hilbert_value}
\end{equation}
where $\phi_{\mathrm H}$ is a learned embedding network.
The value function is trained with goal-conditioned expectile
temporal-difference learning.
We then extract a direct goal-conditioned policy using the same
weighted regression procedure as GC-IQL.
This adaptation retains symmetric Hilbert value structure
but replaces its directional skill learning and goal-directed skill
selection with direct policy extraction.
We denote this adaptation by HILP throughout the control results.

\paragraph{Policy extraction for GC-IQL and HILP.}
For both baselines, the policy is conditioned on the current feature,
goal feature, and an embedding of the normalized remaining horizon.
Recorded actions are fitted with a weighted Gaussian
negative log-likelihood.
The weights are computed from the target-value difference
\begin{equation}
    \Delta V_{t,g}
    =
    V_{\mathrm{targ}}(z_{t+1},z_g)
    -
    V_{\mathrm{targ}}(z_t,z_g),
    \qquad
    w_{t,g}
    =
    \min\left\{
        100,\exp(\alpha_{\mathrm{base}}\Delta V_{t,g})
    \right\},
    \label{eq:baseline_policy_weights}
\end{equation}
with $\alpha_{\mathrm{base}}=10$ for GC-IQL and
$\alpha_{\mathrm{base}}=3$ for HILP.
The weights are treated as fixed during policy optimization.
This implementation uses the value difference rather than the
$Q-V$ advantage of the original IQL policy-extraction rule.
At evaluation time, both policies execute their predicted mean action.

\paragraph{Network parameterization.}
GCBC, GC-IQL, and HILP use MLPs with three hidden layers of width 512.
Their policies receive a 64-dimensional sinusoidal embedding of the
normalized remaining horizon.
The HILP embedding $\phi_{\mathrm H}$ has 64 dimensions.
These policies predict a single action at each decision step and
are queried again after receiving the next observation.

\clearpage
\section{Implementation Details}
\label{app:implementation}

\paragraph{Goal sampling.}
For the temporal objective in
Section~\ref{sec:learning_temporal_geometry}, each current state is paired
with a goal sampled from three sources. We use the next state as the goal
with probability $0.1$, sample a future state from the same trajectory with
probability $0.6$, and sample an observation uniformly from the offline
dataset with probability $0.3$. Future states are sampled geometrically with
$\Pr(\Delta=m)=0.02(0.98)^{m-1}$ for $m\geq1$, with offsets beyond the
remaining trajectory clipped to its final state.

\paragraph{Temporal-critic hyperparameters.}
We use a bootstrap horizon of $n=8$ and a local supervision range of $J=8$.
The expectile parameter $\tau_e=0.3$ is used for both
$\mathcal L_{\mathrm{goal}}$ and $\mathcal L_{\mathrm{local}}$.
The loss coefficients are
$\lambda_{\mathrm{local}}=1$ and
$\lambda_{\mathrm{cons}}=0.1$.

\paragraph{Target networks.}
\label{app:target_networks}
We maintain slowly updated target networks for the control representation and
directed temporal distance. After each critic update, each target parameter
$\bar w$ is updated from its online counterpart $w$ by exponential moving average (EMA)
\begin{equation}
    \bar w
    \leftarrow
    0.995\,\bar w
    +
    0.005\,w,
    \label{eq:target_ema}
\end{equation}
following standard parameter averaging
\citep{polyak1992acceleration}.
Bootstrap targets are computed entirely with the target representation and
target temporal distance, with gradients stopped through all target
quantities. Target-distance estimates and the resulting bootstrapped targets
are capped at $d_{\max}=200$ for numerical stability, while online critic
predictions are left unclipped.

\paragraph{Numerical distance evaluation.}
For numerical stability, the Euclidean component of
Eq.~\eqref{eq:directed_quasimetric} is evaluated using the smoothed norm
$\sqrt{r^2+\epsilon^2}-\epsilon$, where
$r=\|S(e_t)-S(e_g)\|_2$ and $\epsilon=10^{-4}$, which preserves zero
self-distance while smoothing the norm near zero.

\paragraph{Latent dynamics.}
To obtain the predicted successor
$\hat z_{t+k}(\mathbf a)$ in
Eq.~\eqref{eq:local_candidate_selection}, we learn an ensemble of residual
dynamics predictors in the frozen LeWM latent space. For the $m$-th ensemble
member, the next latent state is defined as
\begin{equation}
    \hat z_{t+1}^{(m)}
    =
    z_t
    +
    f_m(z_t,a_t),
    \qquad
    m\in\{1,\ldots,M\},
    \label{eq:app_latent_dynamics}
\end{equation}
where $f_m$ predicts the action-conditioned change in the LeWM latent and
$M$ denotes the number of ensemble members. These predictors are trained
separately from the pretrained LeWM predictor. They operate in the visual latent space $z$, whereas the transition head
$T_\xi$ used in the representation-consistency objective operates in the
control representation space.

For multi-step training, each predictor is initialized at
$\hat z_t^{(m)}=z_t$ and recursively applied under the recorded actions:
\begin{equation}
    \hat z_{t+\ell}^{(m)}
    =
    \hat z_{t+\ell-1}^{(m)}
    +
    f_m
    \left(
        \hat z_{t+\ell-1}^{(m)},
        a_{t+\ell-1}
    \right),
    \qquad
    \ell=1,\ldots,H_F.
    \label{eq:app_dynamics_rollout}
\end{equation}
The predicted rollout is matched to the corresponding recorded LeWM latents
with
\begin{equation}
    \mathcal L_{\mathrm{dyn}}
    =
    \mathbb E_{\mathcal D}
    \left[
        \frac{1}{M H_F d_z}
        \sum_{m=1}^{M}
        \sum_{\ell=1}^{H_F}
        \left\|
            \hat z_{t+\ell}^{(m)}
            -
            z_{t+\ell}
        \right\|_2^2
    \right],
    \label{eq:app_dynamics_loss}
\end{equation}
where $H_F$ is the dynamics training horizon and $d_z$ is the LeWM latent
dimension. 
For candidate evaluation, each ensemble member is rolled out for $k$
transitions under the proposed action sequence $\mathbf a$. The successor
used in Eq.~\eqref{eq:local_candidate_selection} is the ensemble-mean
prediction
\begin{equation}
    \hat z_{t+k}(\mathbf a)
    =
    \frac{1}{M}
    \sum_{m=1}^{M}
    \hat z_{t+k}^{(m)}(\mathbf a),
    \label{eq:app_mean_successor}
\end{equation}
which is mapped through the target representation $\bar\psi$ and then
evaluated by the target temporal critic.

\paragraph{Model-assisted policy augmentation.}
As a training-time augmentation to the advantage-weighted policy objective,
we use the policy to generate alternative action targets that remain supported
by the offline behavior and can be predicted consistently by the learned
dynamics. Specifically, under the same policy condition $c_t$, we sample
$N$ candidate action sequences
$\{\mathbf a_t^{(i)}\}_{i=1}^{N}$ from the conditional flow policy.
Before ranking candidates with the temporal critic, we filter them using a
behavior-support error and a dynamics-disagreement score, since a low predicted
temporal cost is informative only for candidates whose actions remain
compatible with the offline data and whose predicted outcomes are consistent across the learned dynamics ensemble.

\textbf{Behavior support.}
We train an auxiliary state-conditioned flow model $v_\nu$ on recorded
actions by minimizing the standard conditional flow-matching loss
\citep{lipman2023flowmatching}, with the current LeWM latent $z_t$ as its
condition. For the $i$-th candidate $\mathbf a_t^{(i)}$, we define the
behavior-support error as
\begin{equation}
    q_{\mathrm{beh}}^{(i)}
    =
    \frac{1}{R k d_a}
    \sum_{r=1}^{R}
    \left\|
        v_\nu
        \left(
            x_{s_r}^{(i,r)},
            s_r,
            z_t
        \right)
        -
        \left(
            \mathbf a_t^{(i)}-x_0^{(r)}
        \right)
    \right\|_2^2,
    \label{eq:behavior_support_score}
\end{equation}
where
$x_0^{(r)}\sim\mathcal N(0,I_{k d_a})$,
$s_r\sim\mathcal U(0.1,0.9)$, and
$x_{s_r}^{(i,r)}
=
(1-s_r)x_0^{(r)}+s_r\mathbf a_t^{(i)}$.
The parameter $R$ denotes the number of flow residuals used to estimate the
error. A candidate satisfying
$q_{\mathrm{beh}}^{(i)}\leq\delta_{\mathrm{beh}}$
is considered sufficiently supported by the recorded behavior.

\textbf{Dynamics agreement.}
For the same candidate $\mathbf a_t^{(i)}$, each dynamics model predicts a
latent trajectory
$\{\hat z_{t+\ell}^{(m,i)}\}_{\ell=1}^{k}$.
We define the dynamics-disagreement score as
\begin{equation}
    q_{\mathrm{dyn}}^{(i)}
    =
    \frac{1}{k d_z}
    \sum_{\ell=1}^{k}
    \sum_{j=1}^{d_z}
    \operatorname{Std}_{m=1,\ldots,M}
    \left(
        [\hat z_{t+\ell}^{(m,i)}]_j
    \right),
    \label{eq:dynamics_disagreement}
\end{equation}
where $d_z$ denotes the LeWM latent
dimension. A candidate satisfying
$q_{\mathrm{dyn}}^{(i)}\leq\delta_{\mathrm{dyn}}$
is considered sufficiently consistent across the dynamics ensemble. 

The thresholds $\delta_{\mathrm{beh}}$ and $\delta_{\mathrm{dyn}}$ are
calibrated from the corresponding scores of recorded actions. Every $50$
training updates, we compute the $90$th percentile of each score in the
current data batch and update the corresponding threshold by exponential
moving average with retention coefficient $0.99$.

We denote by $\mathcal V_t$ the candidate action sequences satisfying both
$q_{\mathrm{beh}}^{(i)}\leq\delta_{\mathrm{beh}}$ and
$q_{\mathrm{dyn}}^{(i)}\leq\delta_{\mathrm{dyn}}$.
Among these candidates, we select the sequence with the lowest predicted
temporal cost:
\begin{equation}
    \mathbf a_t^\star
    =
    \arg\min_{\mathbf a\in\mathcal V_t}
    \left[
        k
        +
        d_{\bar\theta}
        \left(
            \bar\psi
            \left(
                \hat z_{t+k}(\mathbf a)
            \right),
            \bar e_g
        \right)
    \right].
    \label{eq:local_candidate_selection}
\end{equation}
If $\mathcal V_t$ is empty, no model-assisted update is applied to that
training sample.

We compare the selected candidate with the recorded continuation through the
predicted improvement
\begin{equation}
    \mathcal A_{t,g}^{\mathrm{im}}
    =
    C_{\mathcal D}
    -
    C_\star,
    \label{eq:imagined_advantage}
\end{equation}
where
$C_{\mathcal D}
=
k+d_{\bar\theta}(\bar e_{t+k},\bar e_g)$
is the predicted temporal cost of the recorded continuation, and
$C_\star
=
k+d_{\bar\theta}
\left(
    \bar\psi(\hat z_{t+k}(\mathbf a_t^\star)),
    \bar e_g
\right)$
is the predicted temporal cost of the selected candidate.
A positive $\mathcal A_{t,g}^{\mathrm{im}}$ indicates that the candidate
is predicted to improve upon the recorded continuation.

Accepted candidates provide additional flow-matching targets with weight
\begin{equation}
    w_{t,g}^{\mathrm{im}}
    =
    \min
    \left\{
        w_{\max},
        \exp
        \left(
            \frac{\mathcal A_{t,g}^{\mathrm{im}}}{\beta}
        \right)
    \right\},
    \label{eq:imagined_weight}
\end{equation}
and the corresponding model-assisted policy objective is
\begin{equation}
   \mathcal L_{\mathrm{policy}}^{\mathrm{im}}
    =
    \mathbb E
    \left[
        \mathbf 1
        \left(
            \mathcal A_{t,g}^{\mathrm{im}}>0
        \right)
        w_{t,g}^{\mathrm{im}}\,
        \ell_{\mathrm{FM}}
        \left(
            \omega;
            \mathbf a_t^\star,
            c_t
        \right)
    \right],
    \label{eq:model_assisted_policy}
\end{equation}
where $\lambda_{\mathrm{im}}$ controls the contribution of the model-assisted
targets. The candidate contributes to the policy update only when
$\mathcal A_{t,g}^{\mathrm{im}}>0$.

\clearpage
\section{Algorithm and Training Procedure}
\label{append:training_algorithm}

\begin{algorithm}[ht]
    \caption{Training Directed Temporal Representations for Control}
    \label{alg:dtrc_training}
    \small
    \begin{algorithmic}[1]
        \Statex \textbf{Input:} offline trajectories $\mathcal D$,
        frozen visual encoder $E$

        \State Initialize
        $\psi,d_\theta,T_\xi,\pi_\omega,
        \{f_m\}_{m=1}^{M},v_\nu$
        and target networks $\bar\psi,d_{\bar\theta}$

        \For{$u=1,\ldots,U$}

            \State Sample $(o_t,\mathbf a_t,o_{t+k})\sim\mathcal D$
            and relabel goal $o_g$

            \State $z_t\leftarrow E(o_t)$, $z_g\leftarrow E(o_g)$;
            $e_t\leftarrow\psi(z_t)$, $e_g\leftarrow\psi(z_g)$

            \State Set
            $c_t\leftarrow
            [z_t,z_g,e_t,e_g,\operatorname{Emb}(h_t/H)]$

            \State Update $\{f_m\}_{m=1}^{M}$ with
            $\mathcal L_{\mathrm{dyn}}$
            and update $v_\nu$ on recorded actions

            \State Compute $y_{t,g}$ with
            Eq.~\ref{eq:temporal_bootstrap}

            \State Compute
            $\mathcal A_{t,g}^{\mathrm{data}}$,
            $w_{t,g}^{\mathrm{data}}$, and
            $\mathcal L_{\mathrm{policy}}^{\mathrm{data}}$

            \State
            $\mathcal L_{\mathrm{policy}}^{\mathrm{im}}
            \leftarrow 0$

            \If{$u\geq U_{\mathrm{im}}$}

                \State Sample
                $\{\mathbf a_t^{(i)}\}_{i=1}^{N}
                \sim\pi_\omega(\cdot\mid c_t)$

                \State Compute
                $\{q_{\mathrm{beh}}^{(i)},
                q_{\mathrm{dyn}}^{(i)}\}_{i=1}^{N}$
                and obtain $\mathcal V_t$

                \If{$\mathcal V_t\neq\varnothing$}

                    \State Select $\mathbf a_t^\star$ with
                    Eq.~\ref{eq:local_candidate_selection}

                    \State Compute
                    $\mathcal A_{t,g}^{\mathrm{im}}$ with
                    Eq.~\ref{eq:imagined_advantage}

                    \If{$\mathcal A_{t,g}^{\mathrm{im}}>0$}

                        \State Compute
                        $\mathcal L_{\mathrm{policy}}^{\mathrm{im}}$
                        with Eq.~\ref{eq:model_assisted_policy}

                        \State Refine eligible bootstrapped temporal targets

                    \EndIf
                \EndIf
            \EndIf

            \State Update $\psi,d_\theta,T_\xi$ by minimizing
            $\mathcal L_{\mathrm{temp}}$

            \State Update $\bar\psi,d_{\bar\theta}$ with Eq.~\ref{eq:target_ema}

            \State Update $\pi_\omega$ by minimizing
            $\mathcal L_{\mathrm{policy}}
            =
            \mathcal L_{\mathrm{policy}}^{\mathrm{data}}
            +
            \lambda_{\mathrm{im}}
            \mathcal L_{\mathrm{policy}}^{\mathrm{im}}$

        \EndFor

        \Statex \textbf{Return:} $\pi_\omega$
    \end{algorithmic}
\end{algorithm}

\clearpage

\section{Architectures and Hyperparameters}
\label{app:architectures_hyperparameters}

\paragraph{Network architectures.}
We implement all trainable modules as multilayer perceptrons.
Hidden layers use GELU activations
\citep{hendrycks2016gaussian}, with LayerNorm applied in the representation,
policy, behavior, and dynamics networks
\citep{ba2016layer}.
The symmetric head $S$, order-sensitive head $A_\theta$, and transition head
$T_\xi$ use GELU without LayerNorm.
The representation head additionally applies LayerNorm to its output.
We use linear output layers throughout and do not use dropout.
The directed-distance scale is initialized to $\alpha=1$.
Table~\ref{tab:app_architectures} reports the corresponding input, hidden,
and output dimensions.

\begin{table}[ht]
    \centering
    \caption{
        Trainable network architectures.
        Hidden widths exclude output layers.
        The policy condition has dimension
        $2d_z+2d_e+64=704$.
        $\operatorname{Emb}(s)$ denotes the $64$-dimensional
        flow-time embedding.
    }
    \label{tab:app_architectures}
    \small
    \setlength{\tabcolsep}{4pt}
    \renewcommand{\arraystretch}{1.13}
    \begin{tabular*}{\linewidth}{@{\extracolsep{\fill}}llcc@{}}
        \toprule
        Network & Input & Hidden widths & Output \\
        \midrule
        Representation $\psi$
        & $z$
        & $512,512$
        & $128$ \\

        Symmetric head $S$
        & $e$
        & $256$
        & $64$ \\

        Order-sensitive head $A_\theta$
        & $e$
        & $256$
        & $64$ \\

        Transition head $T_\xi$
        & $[e,a]$
        & $256$
        & $128$ \\

        Dynamics member $f_m$
        & $[z,a]$
        & $512,512$
        & $192$ \\

        Policy velocity $v_\omega$
        & $[x_s,\operatorname{Emb}(s),c_t]$
        & $512,512,512$
        & $k d_a$ \\

        Behavior velocity $v_\nu$
        & $[x_s,\operatorname{Emb}(s),z_t]$
        & $512,512,512$
        & $k d_a$ \\
        \bottomrule
    \end{tabular*}

    \vspace{3pt}
    \begin{minipage}{\linewidth}
        \footnotesize
        $T_\xi$ predicts a residual update to $e$, and each $f_m$
        predicts a residual update to $z$.
        Dynamics ensemble members have independent parameters and are
        trained on the same minibatches.
    \end{minipage}
\end{table}

\paragraph{Hyperparameters.}
Table~\ref{tab:app_hyperparameters} summarizes the default configuration.

\begin{table}[ht]
    \centering
    \caption{
        Default training and inference hyperparameters for DTRC.
    }
    \label{tab:app_hyperparameters}
    \small
    \setlength{\tabcolsep}{5pt}
    \renewcommand{\arraystretch}{1.05}
    \begin{tabular*}{\linewidth}{@{\extracolsep{\fill}}ll@{}}
        \toprule
        Hyperparameter & Value \\
        \midrule

        \multicolumn{2}{@{}l}{\textit{Optimization}} \\
        Training iterations $U$
        & $200{,}000$ \\
        Minibatch size
        & $1024$ \\
        Optimizer
        & Adam \\
        Learning rate
        & $3\times10^{-4}$ (constant) \\
        Adam moment coefficients
        & $(0.9,0.999)$ \\
        Adam numerical epsilon
        & $10^{-8}$ \\
        Weight decay
        & $0$ \\
        Gradient clipping
        & None \\
        \midrule

        \multicolumn{2}{@{}l}{\textit{Temporal critic and goal sampling}} \\
        Next / future / dataset-wide goal probabilities
        & $0.1/0.6/0.3$ \\
        Future-goal continuation $\gamma_g$
        & $0.98$ \\
        Bootstrap horizon $n$
        & $8$ \\
        Local supervision range $J$
        & $8$ \\
        Expectile $\tau_e$
        & $0.3$ \\
        Local coefficient $\lambda_{\mathrm{local}}$
        & $1$ \\
        Consistency coefficient $\lambda_{\mathrm{cons}}$
        & $0.1$ \\
        Target-network update rate
        & $0.005$ \\
        Target distance / bootstrap cap $d_{\max}$
        & $200$ \\
        Norm-smoothing constant $\epsilon$
        & $10^{-4}$ \\
        \midrule

        \multicolumn{2}{@{}l}{\textit{Flow policy}} \\
        Action horizon $k$
        & $1$ \\
        Horizon-conditioning cap $H$
        & $50$ \\
        Flow sampling solver
        & Forward Euler \\
        
        Flow sampling steps
        & $8$ \\
        Advantage temperature $\beta$
        & $1$ \\
        Maximum weight $w_{\max}$
        & $20$ \\
        Action clip (standardized space)
        & $[-3,3]$ \\
        Test-time policy samples per decision
        & $1$ \\
        \midrule

        \multicolumn{2}{@{}l}{\textit{Model-assisted augmentation}} \\
        Dynamics ensemble size $M$
        & $4$ \\
        Dynamics training horizon $H_F$
        & $4$ \\
        Model-assisted start iteration $U_{\mathrm{im}}$
        & $10{,}000$ \\
        Policy candidates per condition $N$
        & $16$ \\
        Behavior-score samples $R$
        & $2$ \\
        Model-assisted coefficient $\lambda_{\mathrm{im}}$
        & $1$ \\
        \bottomrule
    \end{tabular*}
\end{table}

\clearpage

\section{Additional Results}
\label{app:additional_results}

\subsection{Component Ablations}
\label{app:component_ablation}

\begin{table*}[ht]
    \centering
    \caption{
        \textbf{Ablation study on the four LeWM tasks.}
        Success rates (\%) are reported as mean $\pm$ sample standard deviation
        over five evaluation seeds.
        All variants use the same 50 start--goal pairs per task and seed,
        a goal offset of 25 recorded transitions, and a 50-step interaction
        budget.
    }
    \label{tab:component_ablation}

    \small
    \setlength{\tabcolsep}{6pt}
    \renewcommand{\arraystretch}{1.12}

    \begin{tabular}{lcccc}
        \toprule
        Configuration
        & PushT
        & TwoRoom
        & Reacher
        & Cube \\
        \midrule

        Flow BC
        & $74.7 \pm 4.6$
        & $94.7 \pm 4.2$
        & $68.0 \pm 3.5$
        & $91.3 \pm 1.2$ \\

        Flow BC + temporal supervision
        & $84.0 \pm 5.3$
        & $100.0 \pm 0.0$
        & $100.0 \pm 0.0$
        & $97.3 \pm 3.1$ \\

        \midrule

        DTRC w/o behavior support
        & $88.0 \pm 4.0$
        & $100.0 \pm 0.0$
        & $100.0 \pm 0.0$
        & $95.3 \pm 1.2$ \\

        DTRC w/o directional residual
        & $90.0 \pm 5.3$
        & $100.0 \pm 0.0$
        & $100.0 \pm 0.0$
        & $\mathbf{98.7 \pm 2.3}$ \\

        \textbf{DTRC (full)}
        & $\mathbf{90.7 \pm 5.0}$
        & $\mathbf{100.0 \pm 0.0}$
        & $\mathbf{100.0 \pm 0.0}$
        & $96.0 \pm 2.0$ \\
        \bottomrule
    \end{tabular}
\end{table*}

\paragraph{Configuration details.}
Flow BC uses unweighted conditional flow matching.
Flow BC + temporal supervision adds the learned DTRC control
representation and temporal-advantage weighting, without model-assisted
augmentation.
The behavior-support ablation retains the dynamics-agreement filter.
The directional-residual ablation removes the order-sensitive component
and retains only the symmetric metric component.

\subsection{Performance across Goal Offsets}
\label{app:goal_offset_results}

\begin{table}[ht]
    \centering
    \caption{
        \textbf{Goal-conditioned control across goal offsets.}
        Task-averaged success rates (\%) over six OGBench tasks.
    }
    \label{tab:goal_offset_results}

    \small
    \setlength{\tabcolsep}{5pt}
    \renewcommand{\arraystretch}{1.10}

    \begin{tabular}{lccc}
        \toprule
        & \multicolumn{3}{c}{Goal offset} \\
        \cmidrule(lr){2-4}
        Method
        & 25
        & 50
        & 100 \\
        \midrule

        LeWM
        & $43.2 \pm 1.6$
        & $17.6 \pm 4.1$
        & $10.7 \pm 1.9$\\

        GCBC
        & $44.8 \pm 1.4$
        & $16.4 \pm 4.6$
        & $9.1 \pm 0.8$ \\

        GC-IQL
        & $43.4 \pm 2.2$
        & $16.9 \pm 3.8$
        & $9.8 \pm 1.0$ \\

        HILP
        & $44.3 \pm 2.2$
        & $17.2 \pm 2.7$
        & $9.0 \pm 0.9$ \\

        GC-IDM
        & $44.5 \pm 2.2$
        & $16.8 \pm 3.3$
        & $8.3 \pm 0.3$ \\

        \midrule

        \textbf{DTRC (ours)}
        & $\mathbf{45.9 \pm 2.2}$
        & $\mathbf{20.3 \pm 4.7}$
        & $\mathbf{18.9 \pm 0.7}$\\
        \bottomrule
    \end{tabular}
\end{table}

\paragraph{Evaluation details.}
The six tasks are PointMaze, Cube-Double, Scene, AntMaze-Medium,
HumanoidMaze-Medium, and Puzzle-3x3, with equal weight in the task average.
Within each goal offset, all methods use the same 50 start--goal pairs per
task and evaluation seed.
We first average success rates across tasks for each seed, then report the
mean and sample standard deviation over 5 evaluation seeds.
Goal offsets count recorded trajectory transitions rather than minimum
reaching times.
HILP denotes the Hilbert-value adaptation used in our visual-control
comparison.
Checkpoints are fixed across goal offsets and evaluation seeds, so the
reported deviations do not measure variability across independent training
runs.

\clearpage
\section{Additional Representation Analyses}
\label{app:additional_representation}

\subsection{Representation Spaces across Tasks}
\label{app:representation_spaces}

\begin{figure*}[p]
    \centering

    \begin{subfigure}[t]{0.98\textwidth}
        \centering
        \includegraphics[width=\textwidth]{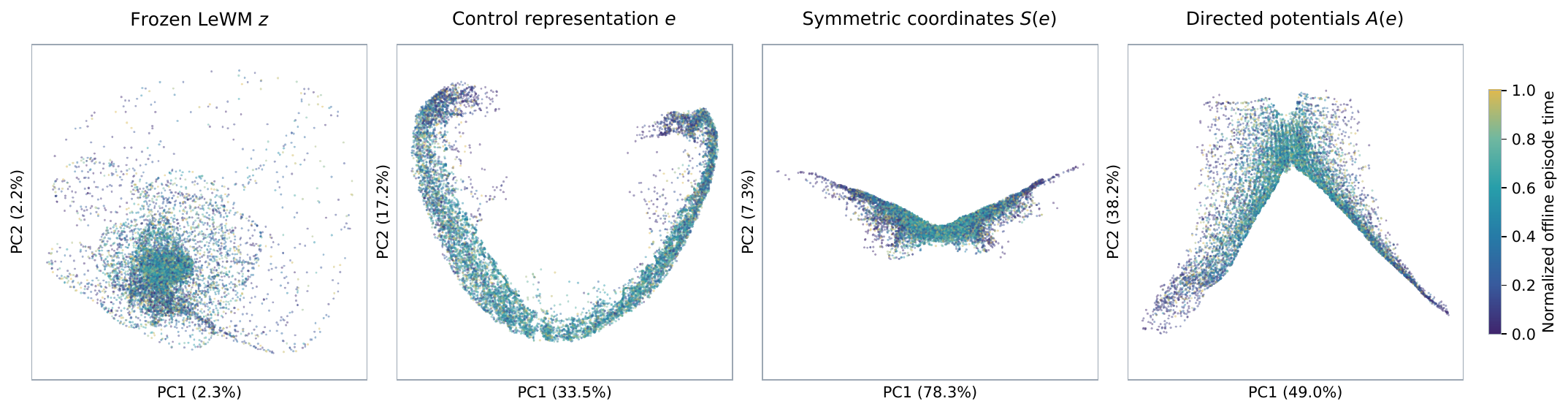}
        \caption{TwoRoom}
        \label{fig:app_four_spaces_tworoom}
    \end{subfigure}

    \vspace{6pt}

    \begin{subfigure}[t]{0.98\textwidth}
        \centering
        \includegraphics[width=\textwidth]{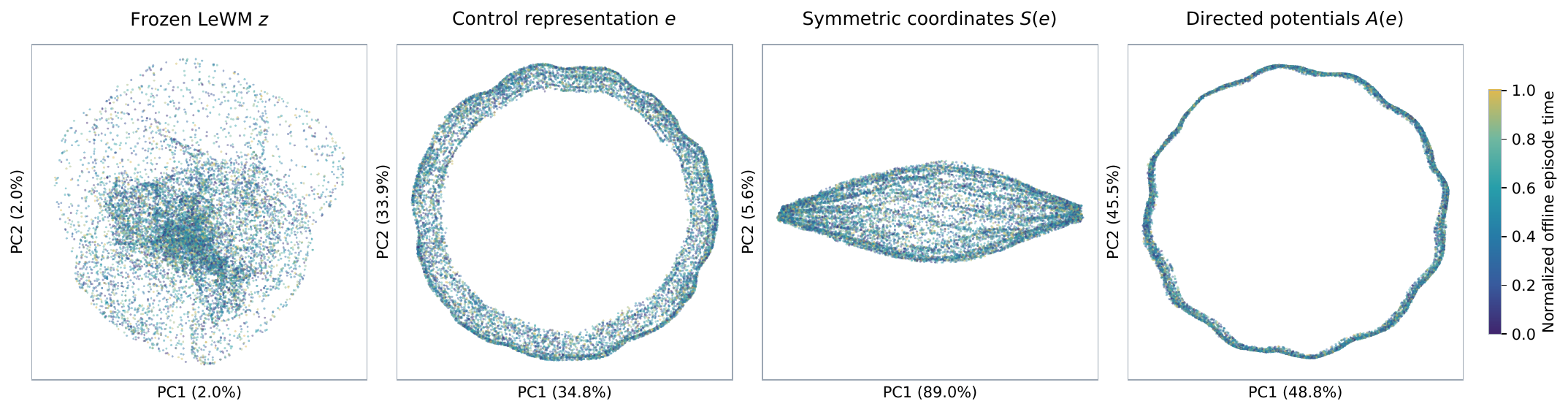}
        \caption{Reacher}
        \label{fig:app_four_spaces_reacher}
    \end{subfigure}

    \caption{
        \textbf{Learned representation spaces on TwoRoom and Reacher.}
        From left to right, each row shows frozen LeWM features $z$,
        the learned control representation $e$, symmetric coordinates
        $S(e)$, and directed potentials $A(e)$.
        Colors denote normalized offline episode time.
        PCA is fitted independently in each space.
    }
    \label{fig:app_four_spaces_tworoom_reacher}
\end{figure*}

\begin{figure*}[p]
    \centering
    \includegraphics[width=\textwidth]{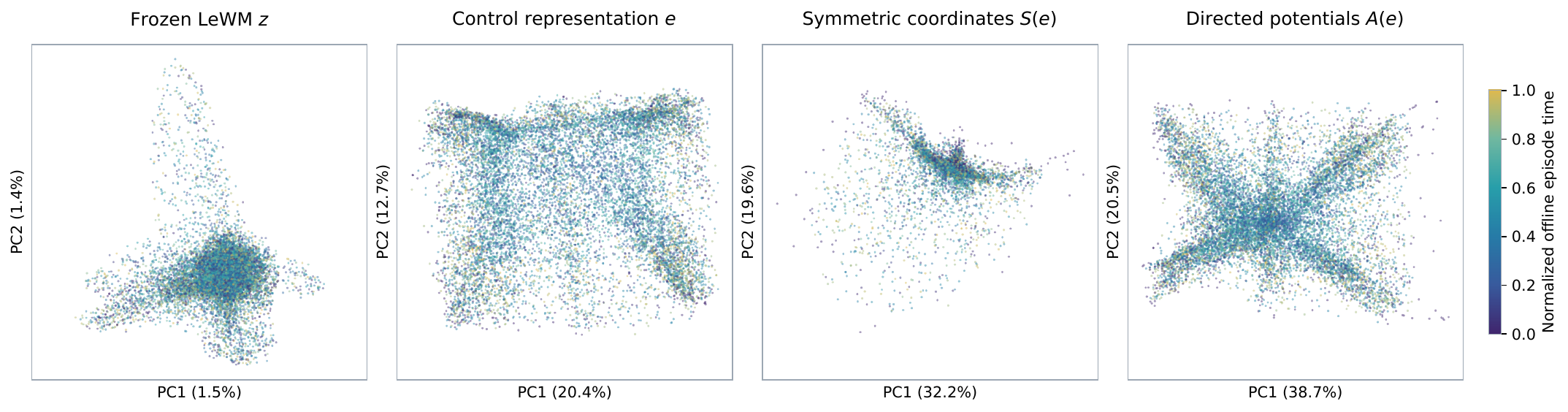}
    \caption{
        \textbf{Learned representation spaces on OGBench-Cube.}
        From left to right, the panels show frozen LeWM features $z$,
        the learned control representation $e$, symmetric coordinates
        $S(e)$, and directed potentials $A(e)$.
        Colors denote normalized offline episode time.
        PCA is fitted independently in each space.
    }
    \label{fig:app_four_spaces_cube}
\end{figure*}

The main text visualizes the transformation from frozen LeWM features
to the learned DTRC representation on PushT.
We provide the corresponding visualizations for the remaining three
LeWM tasks in Figures~\ref{fig:app_four_spaces_tworoom_reacher}
and~\ref{fig:app_four_spaces_cube}.
For each task, we visualize frozen predictive features $z$, the learned
control representation $e$, symmetric coordinates $S(e)$, and directed
potentials $A(e)$.
Each space is projected independently onto its first two principal
components.
Colors indicate normalized time within the offline episode.
The projections are used only for qualitative visualization; all
temporal distances and quantitative diagnostics are computed in the
full learned representation.
PCA orientations and scales are therefore not directly comparable
across spaces or tasks.

The learned spaces exhibit distinct organizations across the three
tasks.
TwoRoom develops a curved control representation with separate
structures in the symmetric and directed heads.
Reacher exhibits an approximately cyclic organization in both the
control representation and directed potentials.
Cube produces a different non-cyclic geometry, with the symmetric and
directed heads again organizing the representation differently.
Together with the PushT visualization in the main text, these examples
show that DTRC does not impose a single low-dimensional geometric form
across tasks.
The relevant common structure is instead defined by the induced
state--goal temporal distance and its control semantics.

\paragraph{Additional OGBench tasks.}
We further visualize the same four representation spaces on the six
additional OGBench tasks.
These tasks exhibit substantially different projected geometries,
providing a broader qualitative view of the task-dependent
representations learned by DTRC.

\begin{figure*}[p]
    \centering

    \begin{subfigure}[t]{0.80\textwidth}
        \centering
        \includegraphics[width=\textwidth]
        {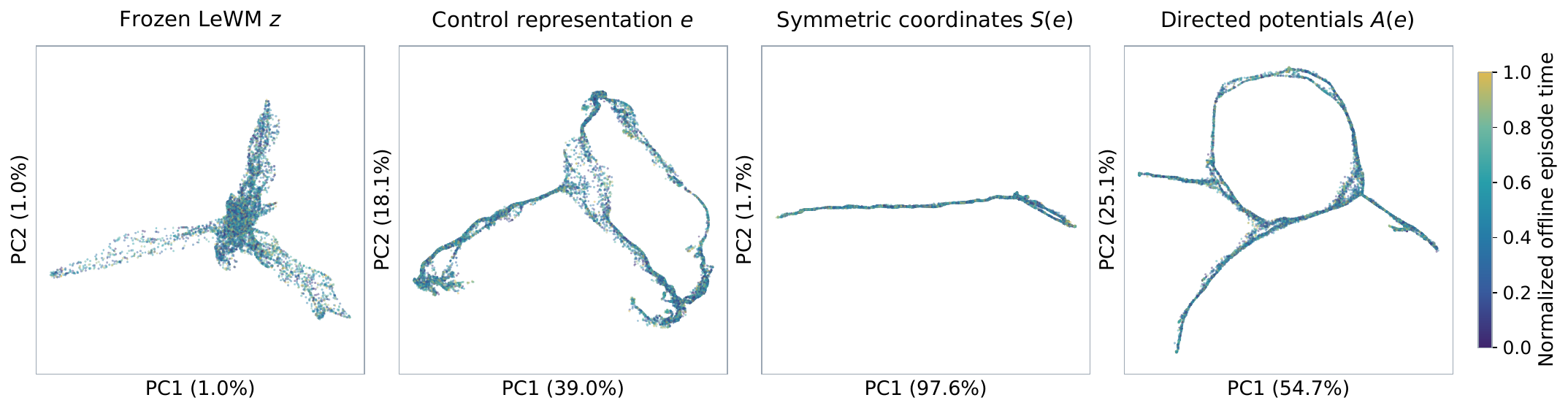}
        \caption{PointMaze}
        \label{fig:app_four_spaces_pointmaze}
    \end{subfigure}

    \vspace{5pt}

    \begin{subfigure}[t]{0.80\textwidth}
        \centering
        \includegraphics[width=\textwidth]
        {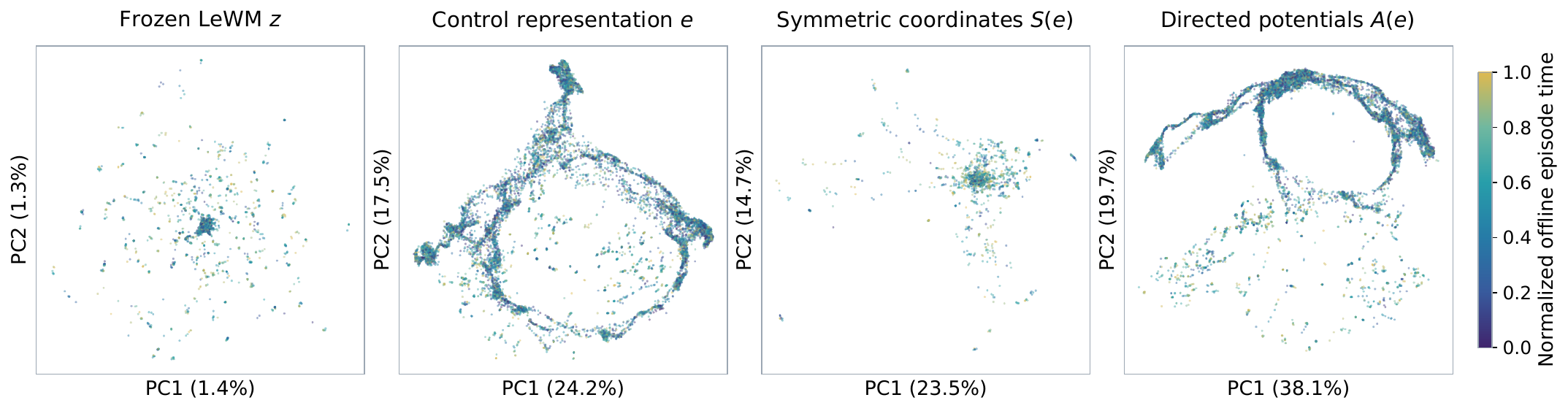}
        \caption{AntMaze-Medium}
        \label{fig:app_four_spaces_antmaze}
    \end{subfigure}

    \caption{
        \textbf{Learned representation spaces on PointMaze and
        AntMaze-Medium.}
        For each task, the panels show frozen LeWM features $z$,
        the learned control representation $e$, symmetric coordinates
        $S(e)$, and directed potentials $A(e)$.
        Colors indicate normalized offline episode time.
        PCA is fitted independently in each space, so orientations and
        scales are not directly comparable across panels or tasks.
    }
    \label{fig:app_four_spaces_navigation}
\end{figure*}

\begin{figure*}[p]
    \centering

    \begin{subfigure}[t]{0.80\textwidth}
        \centering
        \includegraphics[width=\textwidth]
        {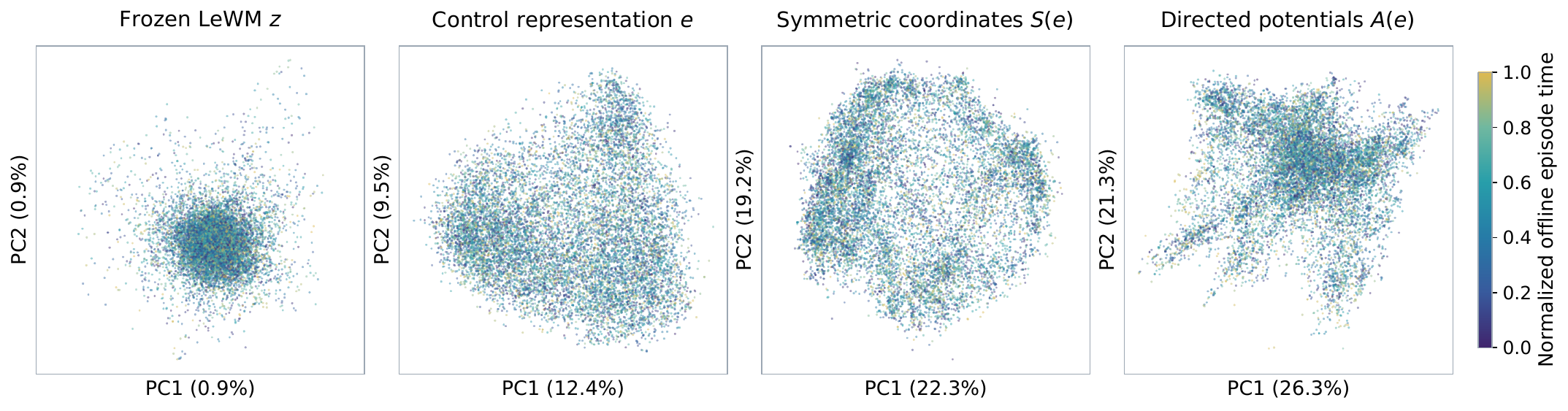}
        \caption{Cube-Double}
        \label{fig:app_four_spaces_cube_double}
    \end{subfigure}

    \vspace{5pt}

    \begin{subfigure}[t]{0.80\textwidth}
        \centering
        \includegraphics[width=\textwidth]
        {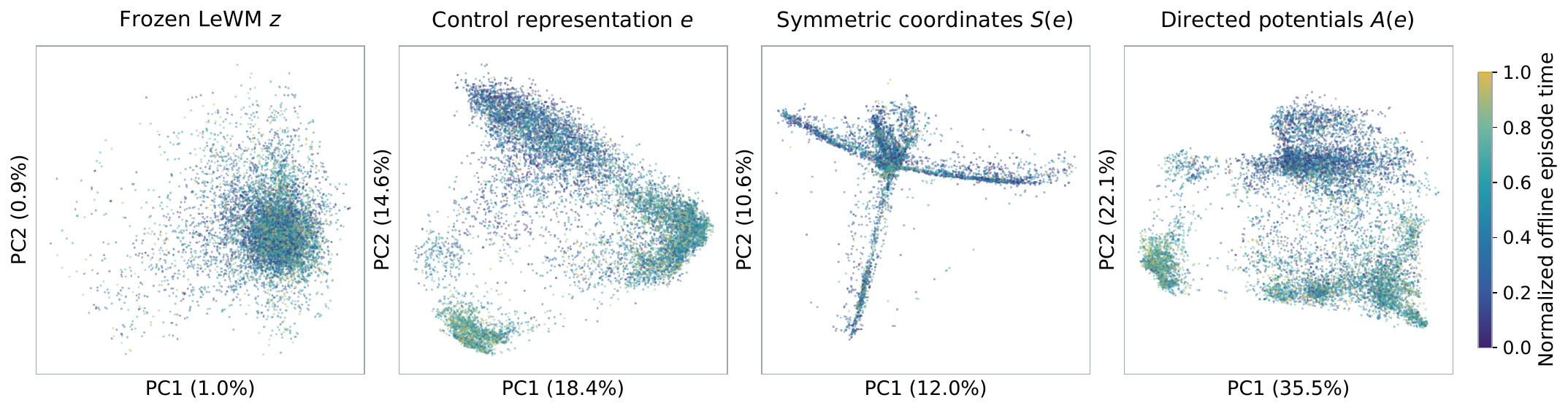}
        \caption{Scene}
        \label{fig:app_four_spaces_scene}
    \end{subfigure}

    \caption{
        \textbf{Learned representation spaces on Cube-Double and Scene.}
        The visualization follows the same construction as
        Figure~\ref{fig:app_four_spaces_navigation}.
        Each space is projected independently onto its first two
        principal components and colored by normalized offline episode
        time.
    }
    \label{fig:app_four_spaces_manipulation}
\end{figure*}

\begin{figure*}[p]
    \centering

    \begin{subfigure}[t]{0.80\textwidth}
        \centering
        \includegraphics[width=\textwidth]
        {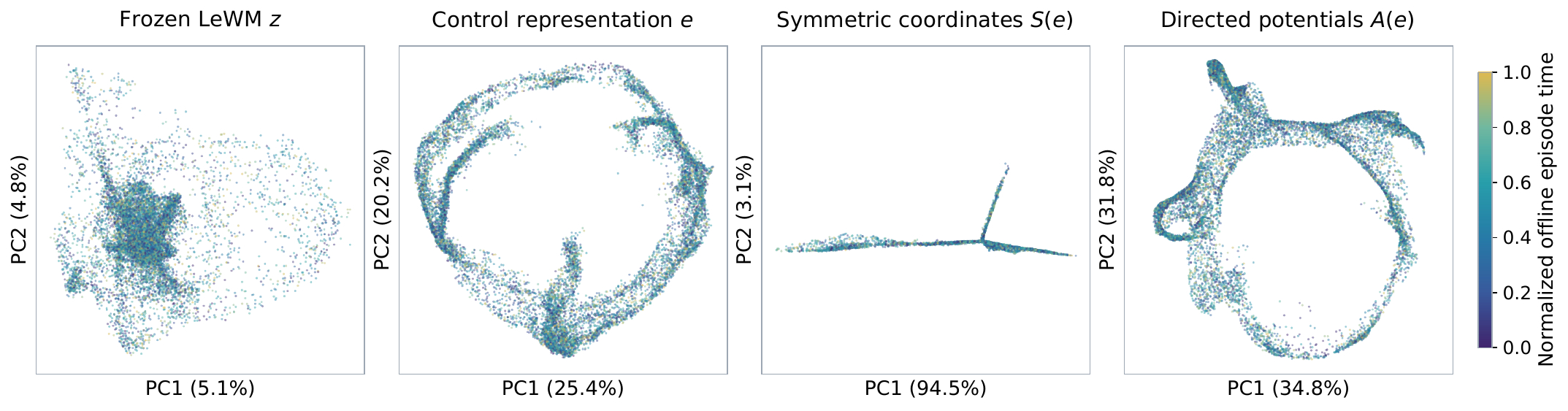}
        \caption{HumanoidMaze-Medium}
        \label{fig:app_four_spaces_humanoidmaze}
    \end{subfigure}

    \vspace{5pt}

    \begin{subfigure}[t]{0.80\textwidth}
        \centering
        \includegraphics[width=\textwidth]
        {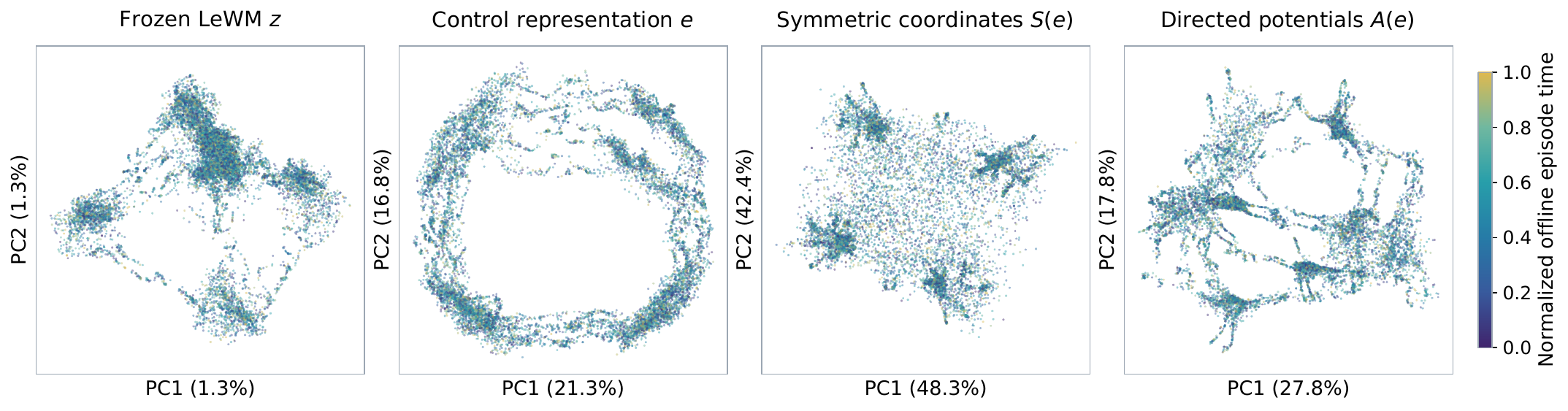}
        \caption{Puzzle-3x3}
        \label{fig:app_four_spaces_puzzle}
    \end{subfigure}

    \caption{
        \textbf{Learned representation spaces on HumanoidMaze-Medium
        and Puzzle-3x3.}
        From left to right, each task shows frozen LeWM features $z$,
        the learned control representation $e$, symmetric coordinates
        $S(e)$, and directed potentials $A(e)$.
        PCA is fitted independently in each space.
    }
    \label{fig:app_four_spaces_humanoid_puzzle}
\end{figure*}

Across the ten tasks, the learned representation does not converge to
a single common projected geometry.
Navigation, reaching, manipulation, and puzzle environments produce
different organizations in $e$, $S(e)$, and $A(e)$.
In several tasks, the leading principal components of the frozen LeWM
features explain only a small fraction of the variance, while the
learned control and distance-head representations exhibit substantially
more structured low-dimensional organization.
The symmetric and directed heads also differ markedly across tasks,
consistent with their distinct roles in the quasimetric and with the
task-dependent directional statistics reported in the main text.
These visualizations therefore support interpreting DTRC through the
induced state--goal temporal geometry rather than through any single
low-dimensional representation shape.

\subsection{Additional Temporal Diagnostics}
\label{app:additional_temporal_diagnostics}

The main text reports detailed temporal-distance diagnostics on PushT.
Here we provide the corresponding results for TwoRoom, Reacher, and
OGBench-Cube under the same held-out, offset-balanced evaluation
protocol.
For each task, we evaluate 20,000 state--future-goal pairs, with
200 pairs at every integer temporal offset from 1 to 100 transitions.
All distances are computed with the EMA target network at 200,000
training updates in the full learned representation.
Recorded temporal offset is used as a reference for trajectory duration
and is not treated as ground-truth optimal reaching cost.

Each figure reports three properties.
The left panel compares forward temporal distance with recorded temporal
offset.
The middle panel compares forward and reverse temporal costs together
with the shared symmetric contribution of the same DTRC model.
The right panel reports one-step temporal progress,
\[
P_{t,g}^{(1)}
=
d(e_t,e_g)-d(e_{t+1},e_g).
\]
These diagnostics use the same definitions as
Section~\ref{sec:temporal_structure}.

\begin{figure*}[p]
    \centering

    \includegraphics[width=0.96\textwidth]
    {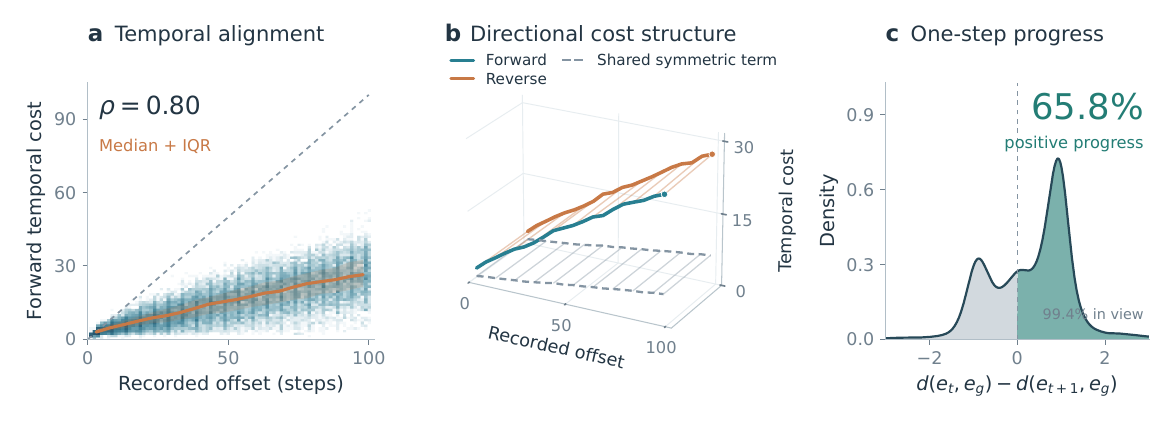}

    \vspace{6pt}

    \includegraphics[width=0.96\textwidth]
    {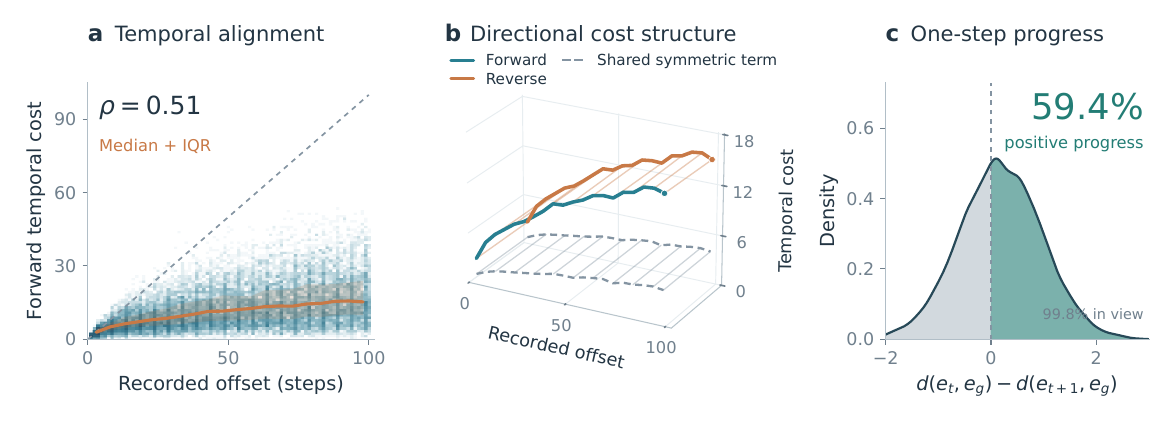}

    \caption{
        \textbf{Additional temporal diagnostics on TwoRoom and Reacher.}
        Each task is evaluated on 20,000 held-out state--future-goal pairs
        with balanced temporal offsets from 1 to 100 transitions.
        The panels show temporal structure across recorded offsets,
        directional state--goal cost, and one-step temporal progress.
        The shared symmetric term is computed from the same trained DTRC
        model and is not a separately trained baseline.
    }
    \label{fig:app_diagnostics_tworoom_reacher}
\end{figure*}

\begin{figure*}[p]
    \centering
    \includegraphics[width=0.96\textwidth]
    {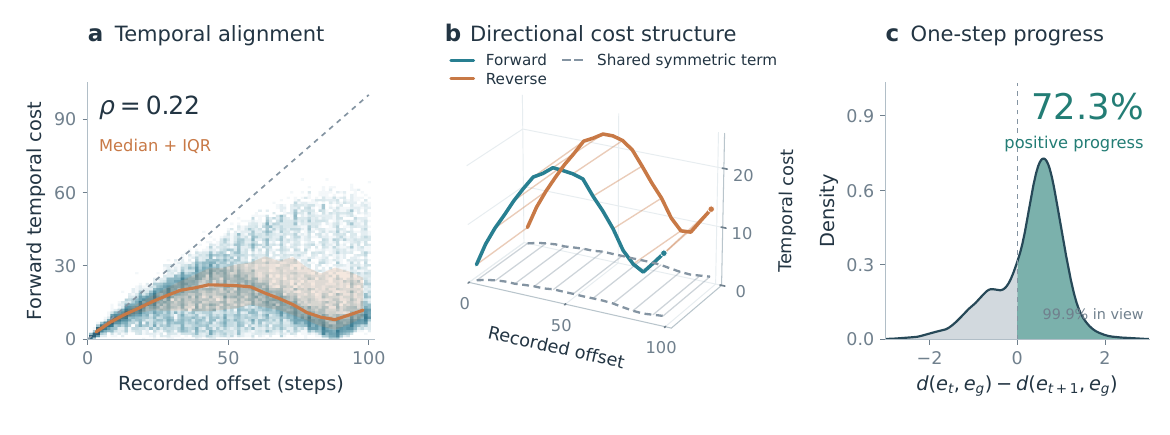}
    \caption{
        \textbf{Additional temporal diagnostics on OGBench-Cube.}
        The evaluation uses the same held-out, offset-balanced protocol
        as the main text.
        The panels show temporal structure across recorded offsets,
        directional state--goal cost, and one-step temporal progress.
    }
    \label{fig:app_diagnostics_cube}
\end{figure*}

The diagnostics reveal distinct long-range structure across tasks.
TwoRoom retains a strong rank association between temporal distance and
recorded offset, with $\rho=0.80$, while Reacher shows a weaker
association of $\rho=0.51$ and Cube shows a substantially weaker
association of $\rho=0.22$.
The forward and reverse curves are also much closer on these tasks than
on PushT, consistent with the task-dependent directional statistics
reported in the main text.
Despite these differences, positive one-step progress remains above
$50\%$ on all three tasks: $65.8\%$ on TwoRoom, $59.4\%$ on Reacher,
and $72.3\%$ on Cube.
These results reinforce the distinction between local actionable
temporal structure and global agreement with recorded trajectory
duration.

\paragraph{Additional OGBench tasks.}
We further apply the same held-out, offset-balanced diagnostic protocol
to the six additional visual OGBench tasks.
For each task, we evaluate 20,000 state--future-goal pairs, with 200
pairs at every integer temporal offset from 1 to 100 transitions.
All distances are computed with the EMA target network at 200,000
training updates in the full learned representation.
The figures report temporal structure across recorded offsets,
directional state--goal cost, and one-step temporal progress using the
same definitions as in the main text.

\begin{figure*}[p]
    \centering

    \begin{subfigure}[t]{0.96\textwidth}
        \centering
        \includegraphics[width=\textwidth]
        {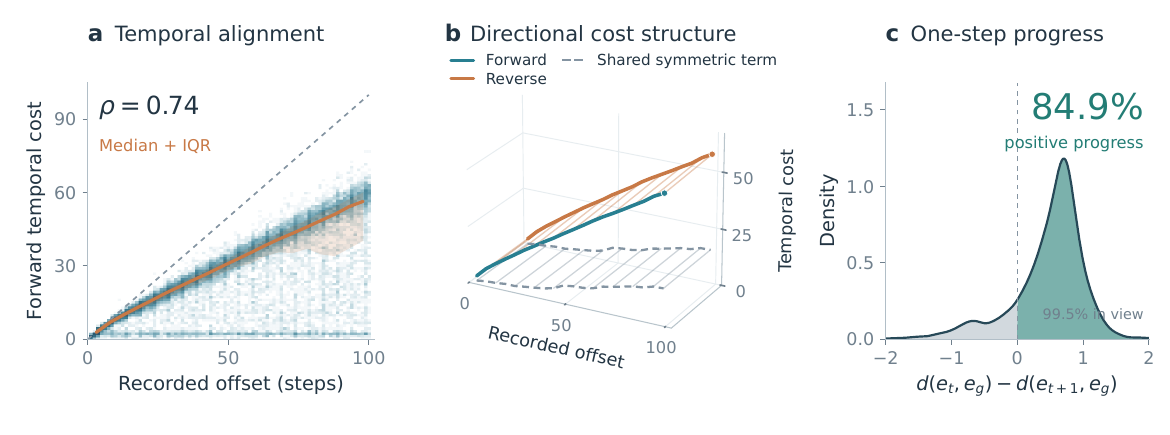}
        \caption{PointMaze}
        \label{fig:app_diagnostics_pointmaze}
    \end{subfigure}

    \vspace{6pt}

    \begin{subfigure}[t]{0.96\textwidth}
        \centering
        \includegraphics[width=\textwidth]
        {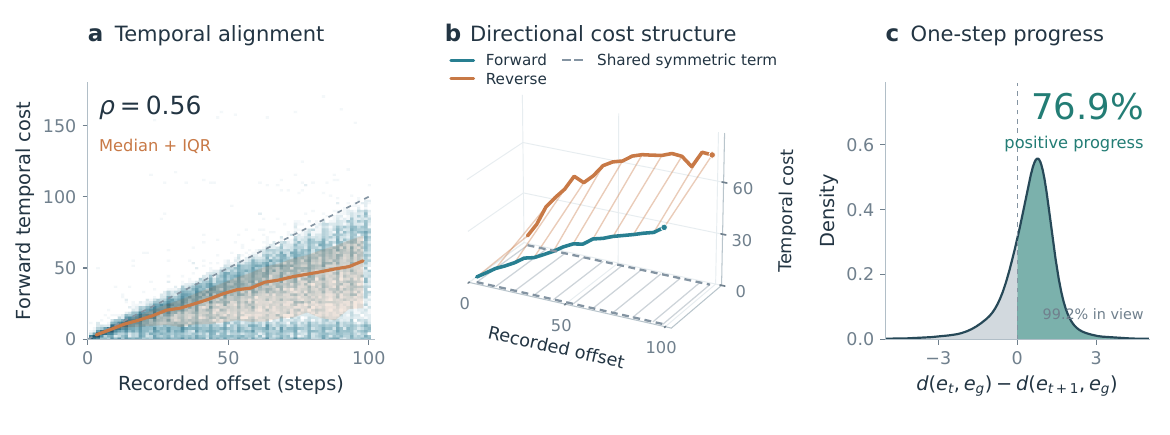}
        \caption{AntMaze-Medium}
        \label{fig:app_diagnostics_antmaze}
    \end{subfigure}

    \caption{
        \textbf{Temporal diagnostics on PointMaze and AntMaze-Medium.}
        The panels show temporal structure across recorded offsets,
        directional state--goal cost, and one-step temporal progress.
        Forward, reverse, and shared symmetric curves are computed from
        the same trained DTRC model.
    }
    \label{fig:app_diagnostics_navigation}
\end{figure*}

\begin{figure*}[p]
    \centering

    \begin{subfigure}[t]{0.96\textwidth}
        \centering
        \includegraphics[width=\textwidth]
        {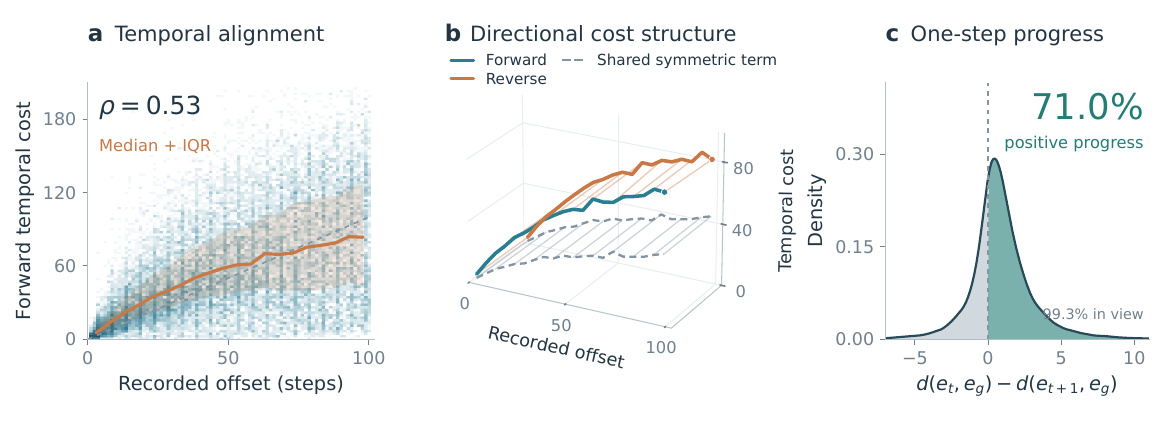}
        \caption{Cube-Double}
        \label{fig:app_diagnostics_cube_double}
    \end{subfigure}

    \vspace{6pt}

    \begin{subfigure}[t]{0.96\textwidth}
        \centering
        \includegraphics[width=\textwidth]
        {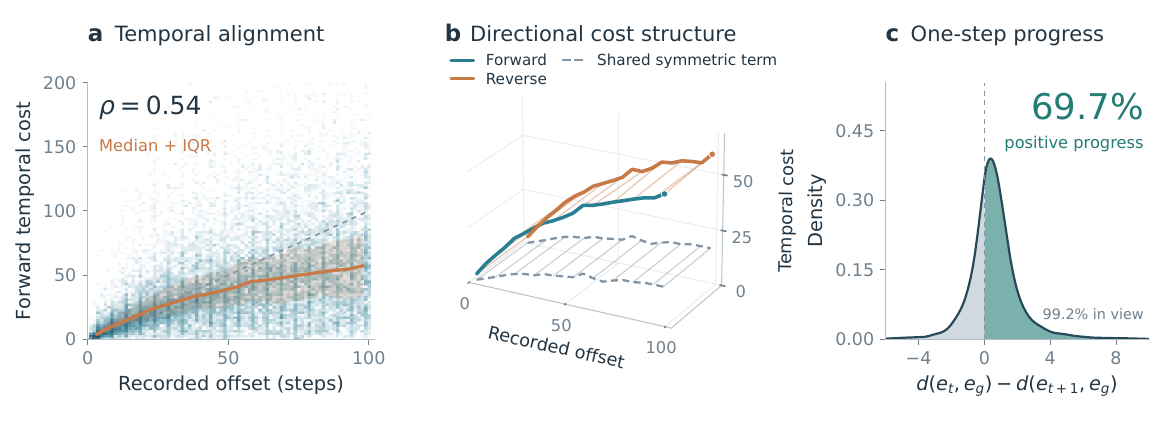}
        \caption{Scene}
        \label{fig:app_diagnostics_scene}
    \end{subfigure}

    \caption{
        \textbf{Temporal diagnostics on Cube-Double and Scene.}
        The evaluation follows the same held-out, offset-balanced
        protocol used for the main-text representation diagnostics.
        Recorded offset is used as a reference for trajectory duration
        rather than as ground-truth optimal reaching cost.
    }
    \label{fig:app_diagnostics_manipulation}
\end{figure*}

\begin{figure*}[p]
    \centering

    \begin{subfigure}[t]{0.96\textwidth}
        \centering
        \includegraphics[width=\textwidth]
        {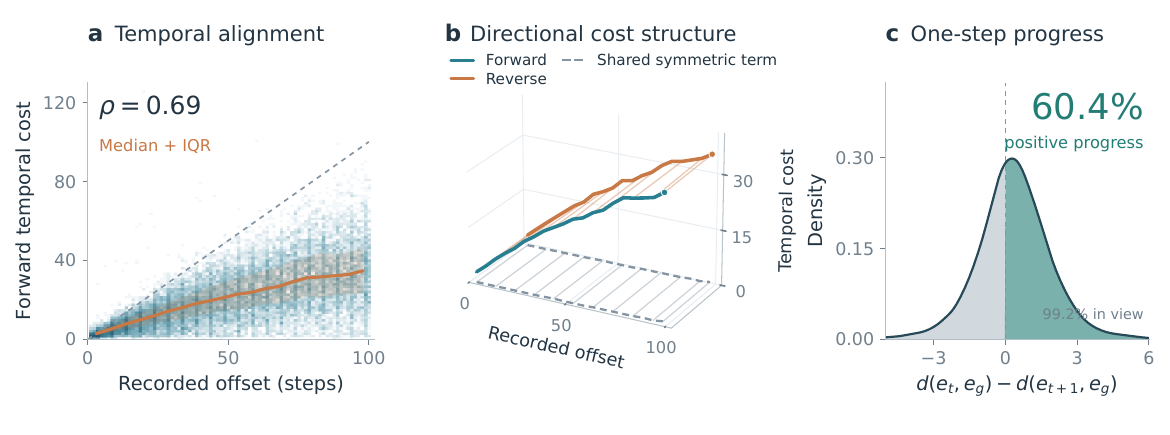}
        \caption{HumanoidMaze-Medium}
        \label{fig:app_diagnostics_humanoidmaze}
    \end{subfigure}

    \vspace{6pt}

    \begin{subfigure}[t]{0.96\textwidth}
        \centering
        \includegraphics[width=\textwidth]
        {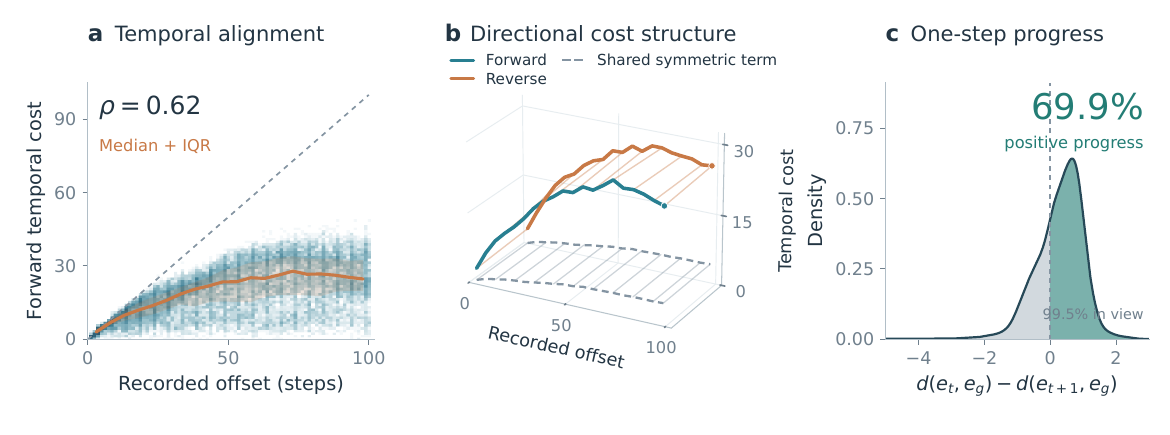}
        \caption{Puzzle-3x3}
        \label{fig:app_diagnostics_puzzle}
    \end{subfigure}

    \caption{
        \textbf{Temporal diagnostics on HumanoidMaze-Medium and
        Puzzle-3x3.}
        Each task is evaluated on the same offset-balanced cohort used
        throughout the additional temporal diagnostics.
    }
    \label{fig:app_diagnostics_humanoid_puzzle}
\end{figure*}

Across the six additional tasks, the relation between temporal distance
and recorded trajectory duration remains task-dependent.
Spearman correlation over offsets 1--100 ranges from $0.53$ on
Cube-Double to $0.74$ on PointMaze.
The directional curves also exhibit different degrees of
forward--reverse separation across tasks, further supporting the
task-dependent use of the order-sensitive component.
In contrast, positive one-step progress remains above $60\%$ on every
task, ranging from $60.4\%$ on HumanoidMaze-Medium to $84.9\%$ on
PointMaze.
Together with the four LeWM tasks, these results show that
transition-level temporal progress is observed across all ten evaluated
control problems without requiring a common long-range or directional
geometry.

\subsection{Quasimetric Decomposition}
\label{app:quasimetric_decomposition}

We further examine the two components of the directed temporal
quasimetric along recorded trajectories.
For each task, we evaluate pairwise distances between states from a
recorded episode and index source and goal states by normalized episode
time.
We separately visualize the symmetric metric contribution, the
order-sensitive directed contribution, and their sum.
These matrices provide a qualitative view of the structure induced by
the two components of Eq.~\ref{eq:directed_quasimetric}.

\begin{figure*}[p]
    \centering

    \begin{subfigure}[t]{0.98\textwidth}
        \centering
        \includegraphics[width=\textwidth]
        {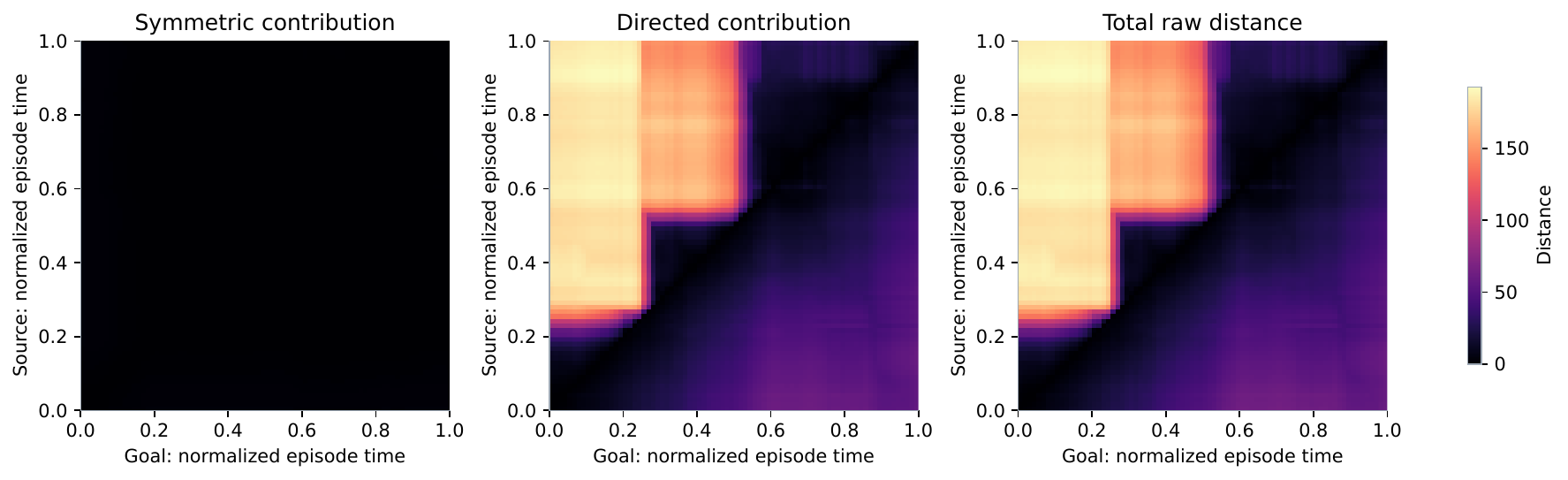}
        \caption{PushT}
        \label{fig:app_quasimetric_pusht}
    \end{subfigure}

    \vspace{6pt}

    \begin{subfigure}[t]{0.98\textwidth}
        \centering
        \includegraphics[width=\textwidth]
        {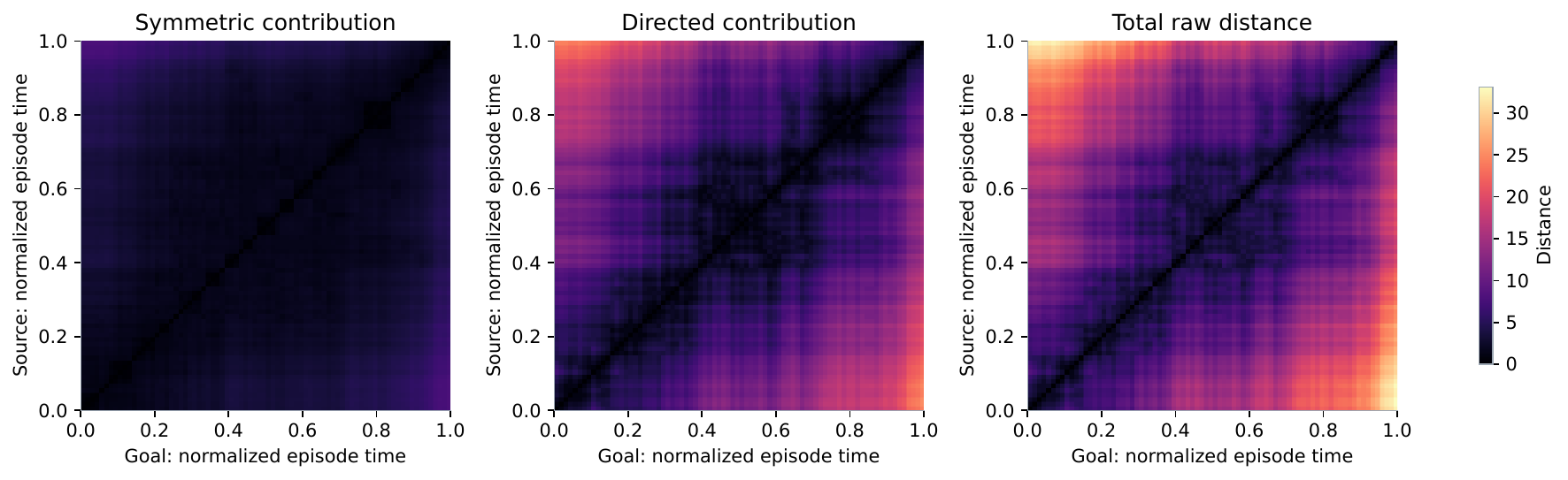}
        \caption{TwoRoom}
        \label{fig:app_quasimetric_tworoom}
    \end{subfigure}

    \caption{
        \textbf{Pairwise decomposition of the learned temporal distance
        on PushT and TwoRoom.}
        Rows index source states and columns index goal states by
        normalized episode time.
        From left to right, each row shows the symmetric metric
        contribution, the order-sensitive directed contribution, and
        their total temporal distance.
        Each task uses its own distance scale.
    }
    \label{fig:app_quasimetric_pusht_tworoom}
\end{figure*}

\begin{figure*}[p]
    \centering

    \begin{subfigure}[t]{0.98\textwidth}
        \centering
        \includegraphics[width=\textwidth]
        {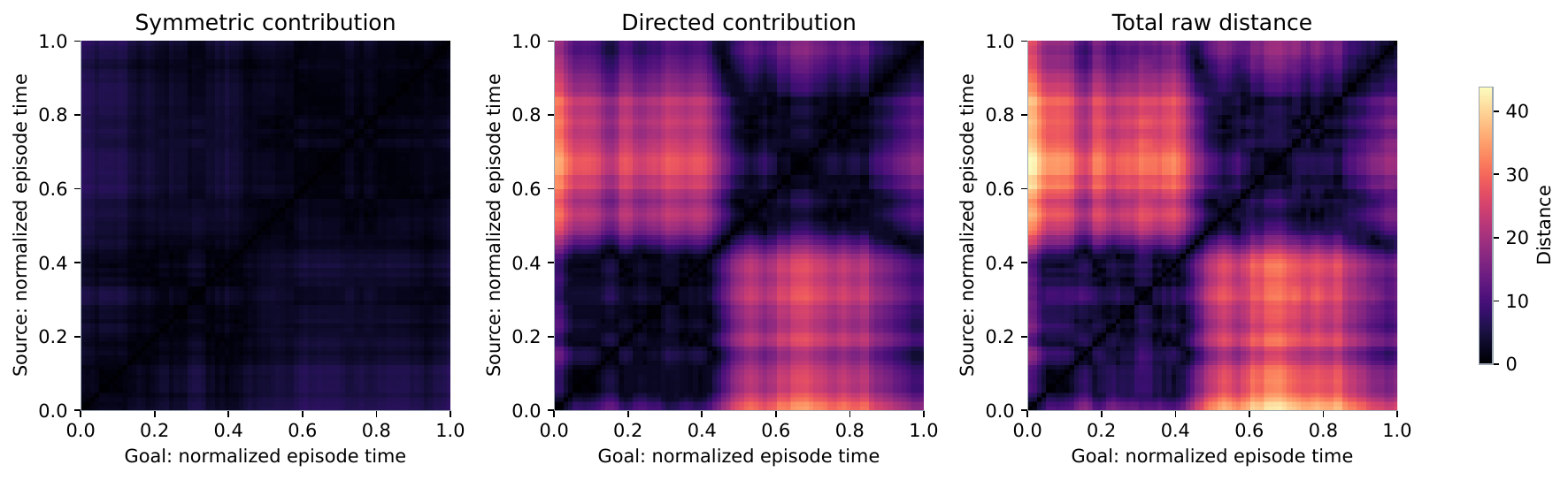}
        \caption{Reacher}
        \label{fig:app_quasimetric_reacher}
    \end{subfigure}

    \vspace{6pt}

    \begin{subfigure}[t]{0.98\textwidth}
        \centering
        \includegraphics[width=\textwidth]
        {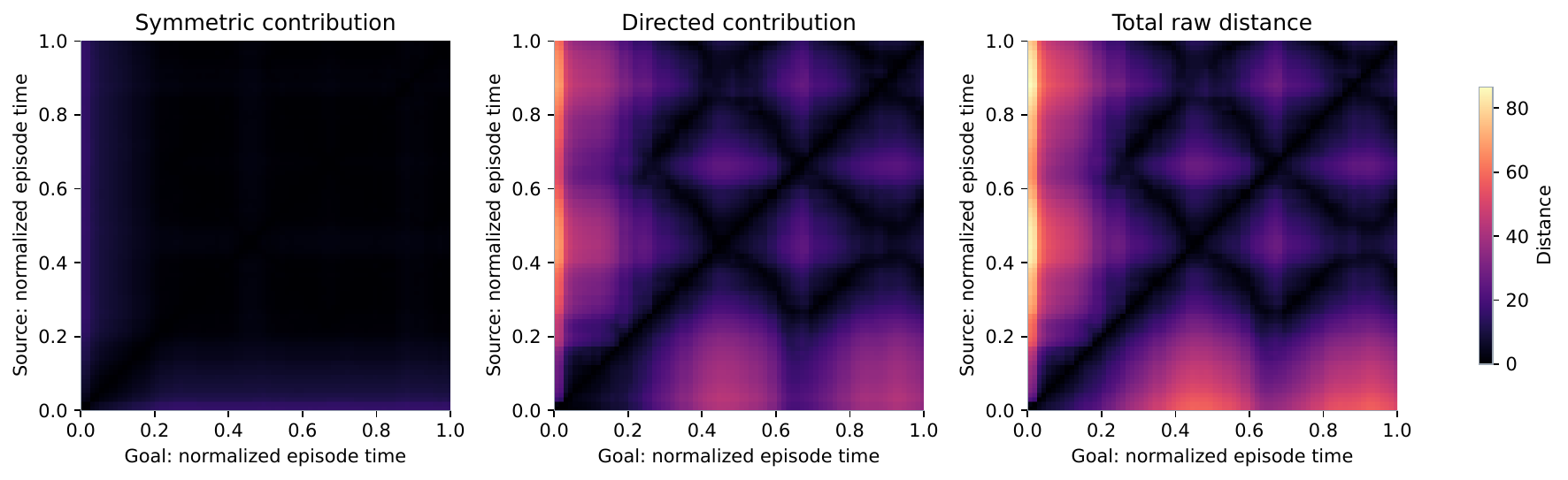}
        \caption{OGBench-Cube}
        \label{fig:app_quasimetric_cube}
    \end{subfigure}

    \caption{
        \textbf{Pairwise decomposition of the learned temporal distance
        on Reacher and OGBench-Cube.}
        Rows index source states and columns index goal states by
        normalized episode time.
        From left to right, each row shows the symmetric metric
        contribution, the order-sensitive directed contribution, and
        their total temporal distance.
        Each task uses its own distance scale.
    }
    \label{fig:app_quasimetric_reacher_cube}
\end{figure*}

The decomposition varies substantially across tasks.
On PushT, the pairwise structure of the total temporal distance is
dominated by the order-sensitive component, consistent with the strong
forward--reverse asymmetry observed in the main-text diagnostics.
TwoRoom, Reacher, and Cube exhibit different relative contributions and
pairwise structures.
The directed component therefore provides task-dependent
order-sensitive structure rather than imposing a common asymmetric
pattern across environments.
The symmetric component remains invariant to source--goal reversal,
while asymmetry in the total distance arises from the directed
component by construction.

\paragraph{Additional OGBench tasks.}
We provide the same quasimetric decomposition for the six additional
visual OGBench tasks.
As in the four LeWM tasks, each matrix evaluates pairwise temporal
distance along a recorded trajectory and separates the symmetric metric
contribution from the order-sensitive directed contribution.

\begin{figure*}[p]
    \centering

    \begin{subfigure}[t]{0.98\textwidth}
        \centering
        \includegraphics[width=\textwidth]
        {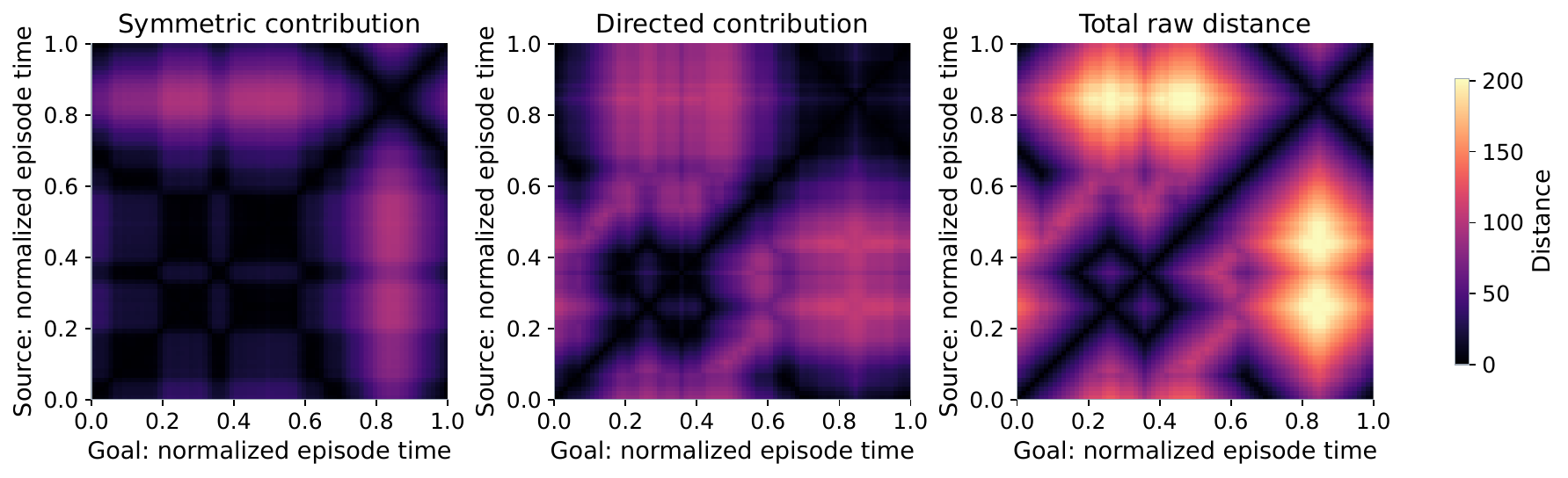}
        \caption{PointMaze}
        \label{fig:app_quasimetric_pointmaze}
    \end{subfigure}

    \vspace{6pt}

    \begin{subfigure}[t]{0.98\textwidth}
        \centering
        \includegraphics[width=\textwidth]
        {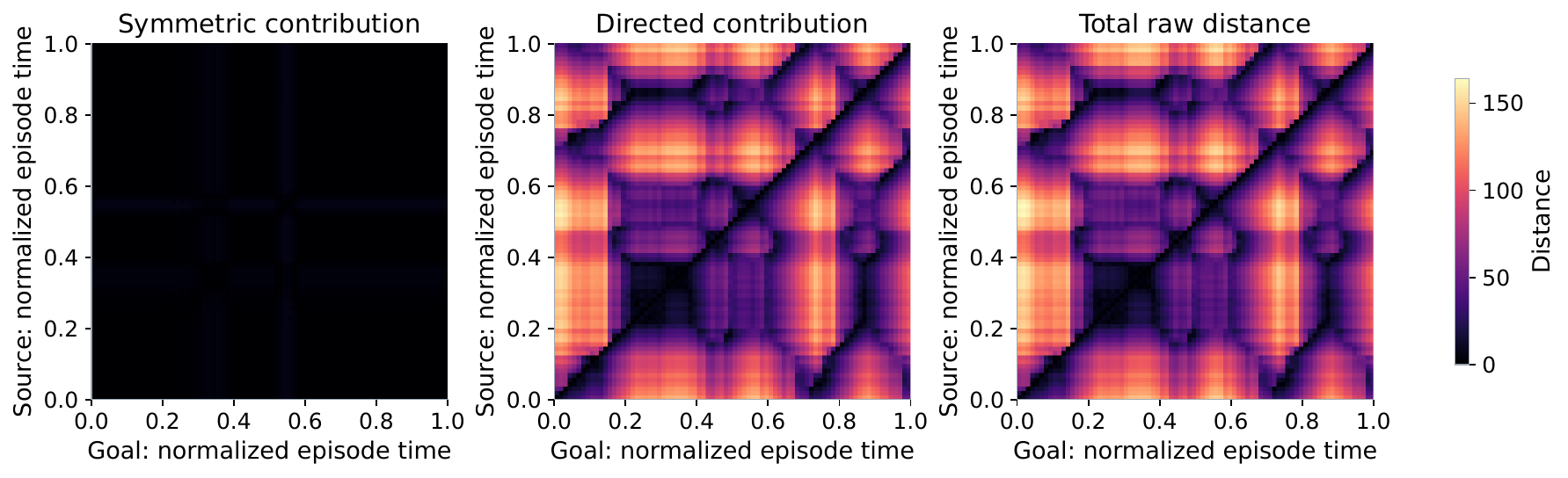}
        \caption{AntMaze-Medium}
        \label{fig:app_quasimetric_antmaze}
    \end{subfigure}

    \caption{
        \textbf{Pairwise decomposition of the learned temporal distance
        on PointMaze and AntMaze-Medium.}
        Rows index source states and columns index goal states by
        normalized episode time.
        From left to right, each row shows the symmetric metric
        contribution, the order-sensitive directed contribution, and
        their total temporal distance.
        Each task uses its own distance scale.
    }
    \label{fig:app_quasimetric_navigation}
\end{figure*}

\begin{figure*}[p]
    \centering

    \begin{subfigure}[t]{0.98\textwidth}
        \centering
        \includegraphics[width=\textwidth]
        {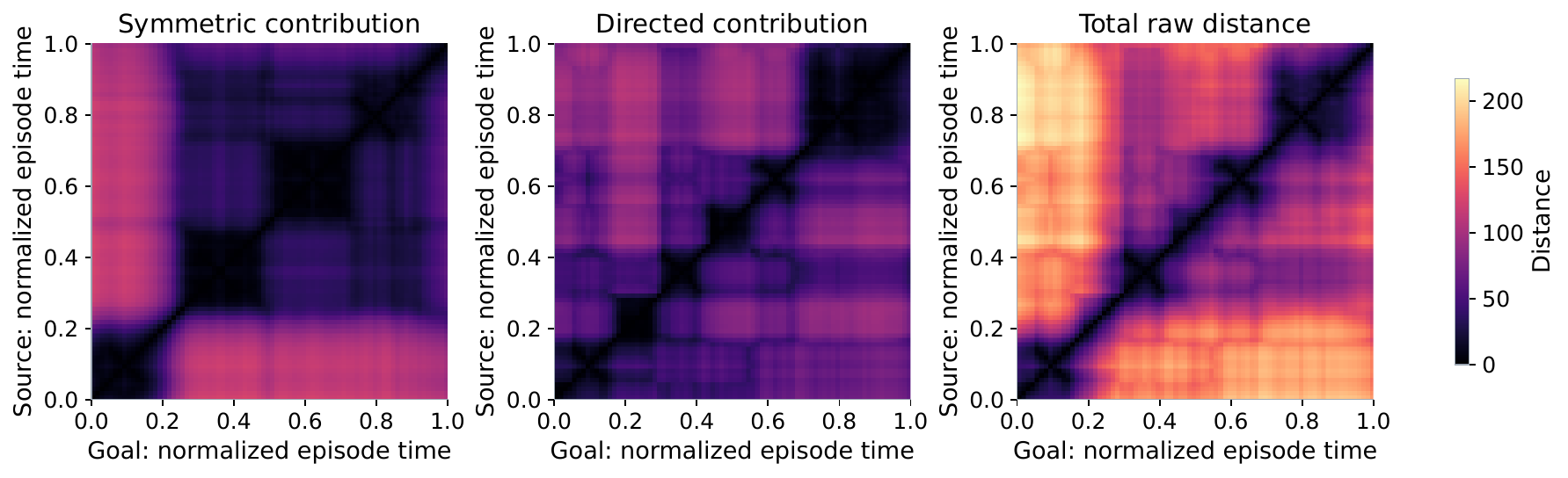}
        \caption{Cube-Double}
        \label{fig:app_quasimetric_cube_double}
    \end{subfigure}

    \vspace{6pt}

    \begin{subfigure}[t]{0.98\textwidth}
        \centering
        \includegraphics[width=\textwidth]
        {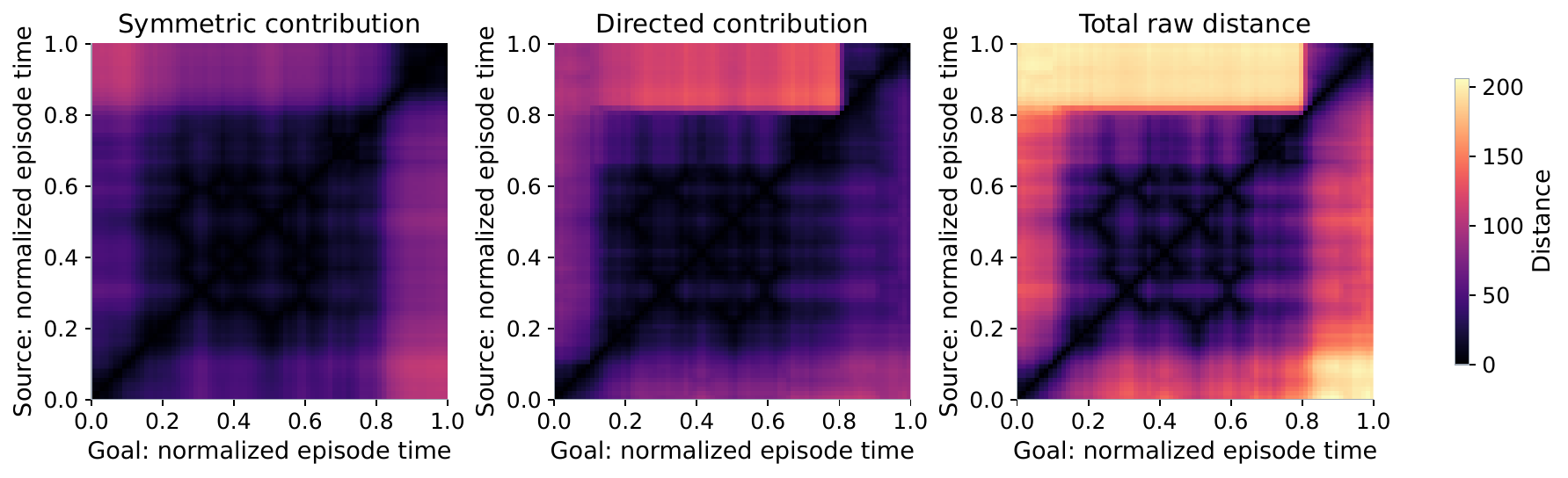}
        \caption{Scene}
        \label{fig:app_quasimetric_scene}
    \end{subfigure}

    \caption{
        \textbf{Pairwise decomposition of the learned temporal distance
        on Cube-Double and Scene.}
        The visualization follows the same construction as
        Figure~\ref{fig:app_quasimetric_navigation}.
        Symmetric and directed contributions are evaluated separately
        before forming the total temporal distance.
    }
    \label{fig:app_quasimetric_manipulation}
\end{figure*}

\begin{figure*}[p]
    \centering

    \begin{subfigure}[t]{0.98\textwidth}
        \centering
        \includegraphics[width=\textwidth]
        {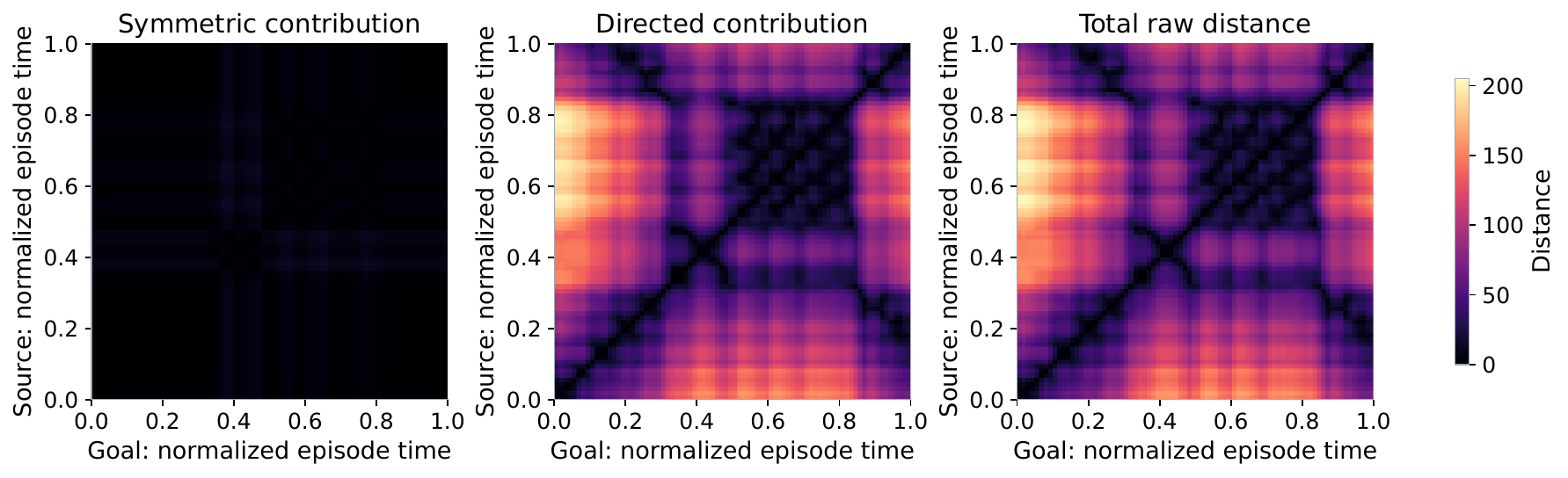}
        \caption{HumanoidMaze-Medium}
        \label{fig:app_quasimetric_humanoidmaze}
    \end{subfigure}

    \vspace{6pt}

    \begin{subfigure}[t]{0.98\textwidth}
        \centering
        \includegraphics[width=\textwidth]
        {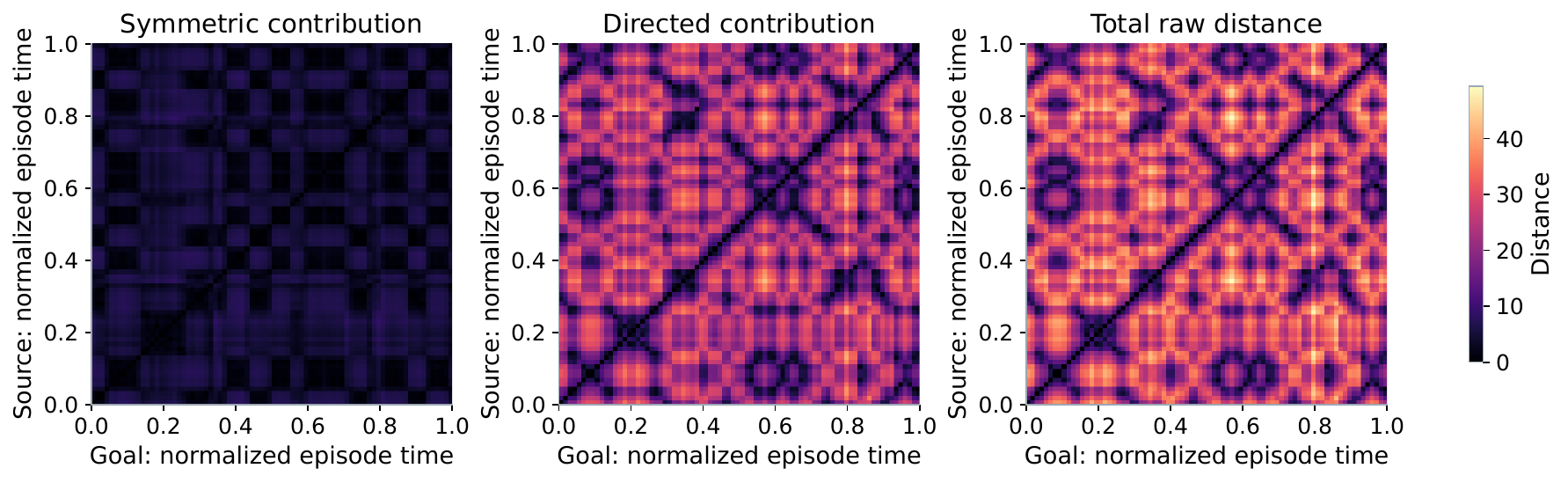}
        \caption{Puzzle-3x3}
        \label{fig:app_quasimetric_puzzle}
    \end{subfigure}

    \caption{
        \textbf{Pairwise decomposition of the learned temporal distance
        on HumanoidMaze-Medium and Puzzle-3x3.}
        Rows and columns index source and goal states by normalized
        episode time.
        Each task uses its own distance scale.
    }
    \label{fig:app_quasimetric_humanoid_puzzle}
\end{figure*}

Across the additional tasks, the relative contributions of the
symmetric and directed components vary substantially.
PointMaze, Cube-Double, and Scene exhibit visible structure in both
components, while AntMaze-Medium, HumanoidMaze-Medium, and Puzzle-3x3
place comparatively greater structure in the order-sensitive
component.
The total distance inherits these task-specific decompositions rather
than following a common pairwise pattern across environments.
Together with the four LeWM tasks, these visualizations show that the
metric--residual construction provides a common structural form while
allowing the learned balance between symmetric and directed geometry to
adapt to the control problem.

\subsection{Goal-Relative Directional Landscapes}
\label{app:goal_relative_landscapes}

We further visualize the directional temporal cost around a fixed
recorded goal.
For each task, we evaluate the temporal distance from every sampled
state to the reference goal and in the reverse direction from the goal
to each state.
We also visualize the directional difference
$d(e_g,e)-d(e,e_g)$.
Distances are computed in the full learned representation.
PCA is used only to position states for visualization.

\begin{figure*}[p]
    \centering

    \begin{subfigure}[t]{0.98\textwidth}
        \centering
        \includegraphics[width=\textwidth]
        {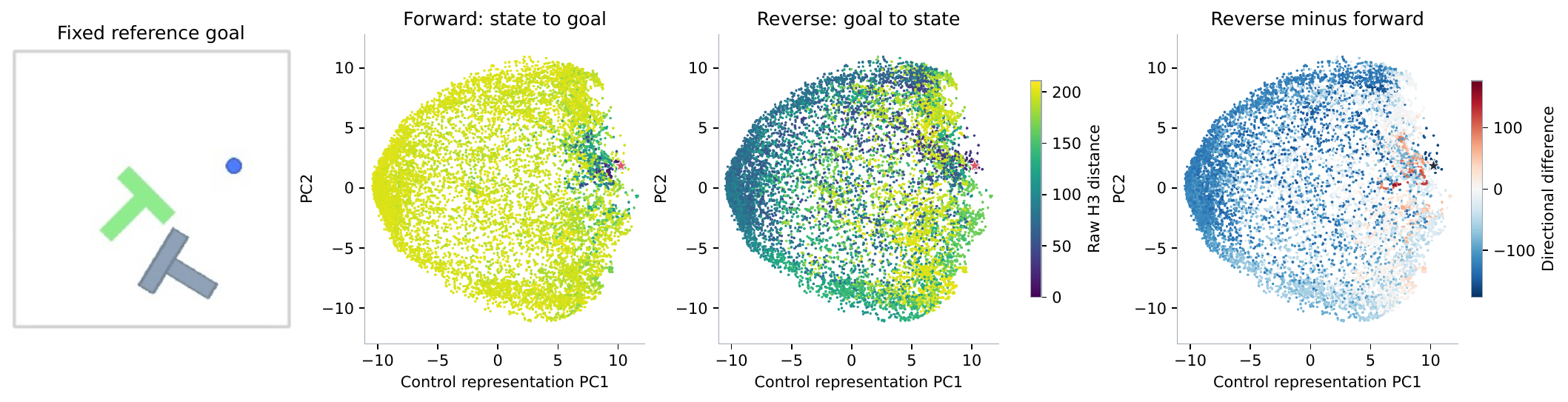}
        \caption{PushT}
        \label{fig:app_goal_landscape_pusht}
    \end{subfigure}

    \vspace{6pt}

    \begin{subfigure}[t]{0.98\textwidth}
        \centering
        \includegraphics[width=\textwidth]
        {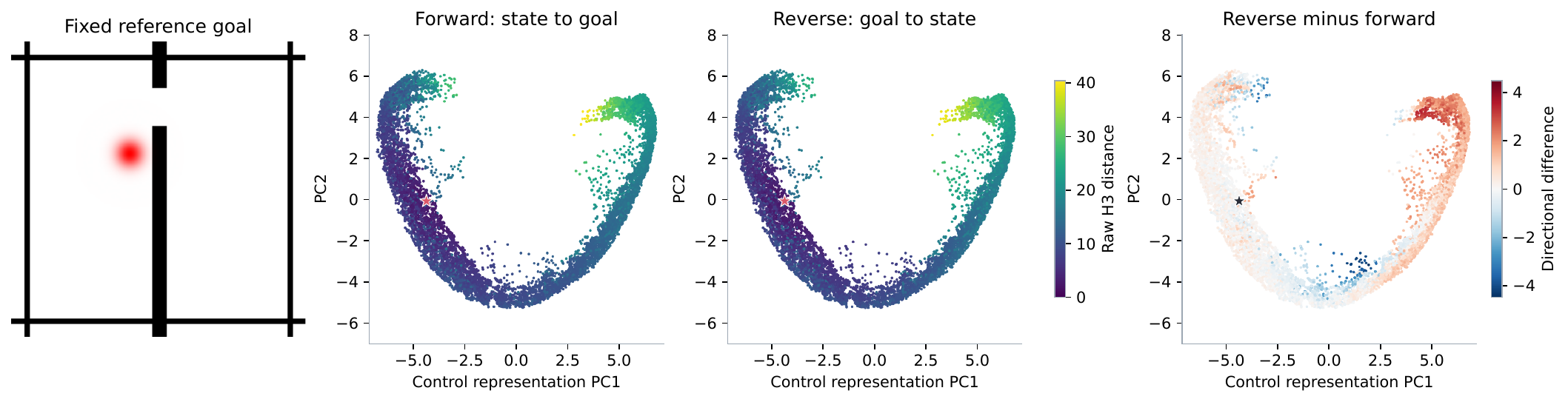}
        \caption{TwoRoom}
        \label{fig:app_goal_landscape_tworoom}
    \end{subfigure}

    \caption{
        \textbf{Goal-relative directional landscapes on PushT and
        TwoRoom.}
        For a fixed recorded goal, the panels show the goal observation,
        forward temporal distance from each state to the goal, reverse
        temporal distance from the goal to each state, and the
        reverse--forward difference.
        PCA is used only for visualization; temporal distances are
        evaluated in the full learned representation.
    }
    \label{fig:app_goal_landscape_pusht_tworoom}
\end{figure*}

\begin{figure*}[p]
    \centering

    \begin{subfigure}[t]{0.98\textwidth}
        \centering
        \includegraphics[width=\textwidth]
        {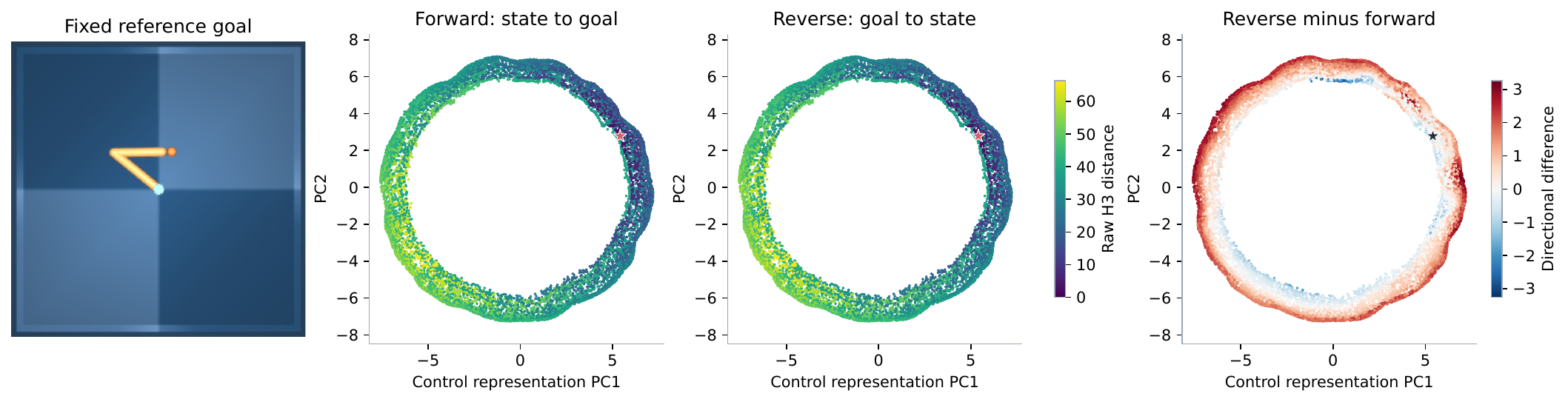}
        \caption{Reacher}
        \label{fig:app_goal_landscape_reacher}
    \end{subfigure}

    \vspace{6pt}

    \begin{subfigure}[t]{0.98\textwidth}
        \centering
        \includegraphics[width=\textwidth]
        {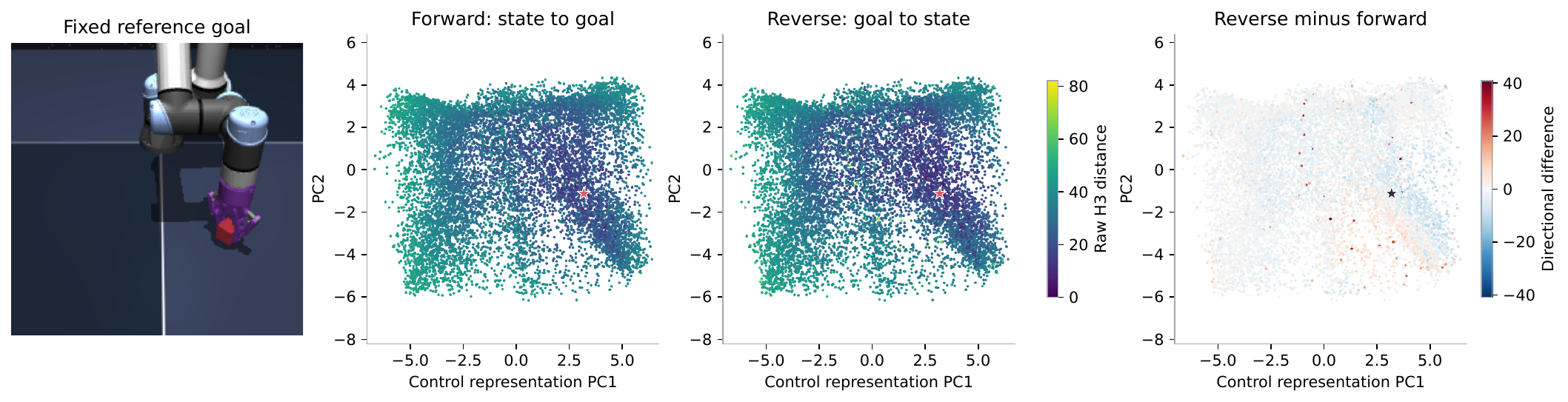}
        \caption{OGBench-Cube}
        \label{fig:app_goal_landscape_cube}
    \end{subfigure}

    \caption{
        \textbf{Goal-relative directional landscapes on Reacher and
        OGBench-Cube.}
        The visualization follows the same construction as
        Figure~\ref{fig:app_goal_landscape_pusht_tworoom}.
        The forward and reverse landscapes are evaluated for the same
        fixed goal, and the final panel shows their difference.
    }
    \label{fig:app_goal_landscape_reacher_cube}
\end{figure*}

The directional landscapes vary substantially across tasks.
PushT exhibits large spatial differences between forward and reverse
costs, consistent with its pronounced forward--reverse asymmetry in the
main-text diagnostics.
TwoRoom, Reacher, and Cube show substantially weaker or more localized
differences.
These visualizations therefore provide a spatial view of the
task-dependent directional structure captured by the quasimetric.
They complement the aggregate forward--reverse statistics reported in
Section~\ref{sec:temporal_structure}.

\paragraph{Additional OGBench tasks.}
We provide the same fixed-goal directional visualizations for the six
additional OGBench tasks.
For each task, we evaluate forward temporal distance from sampled states
to a fixed recorded goal, reverse temporal distance from the goal to the
same states, and their difference.
All distances are computed in the full learned representation.
PCA is used only to position states for visualization.

\begin{figure*}[p]
    \centering

    \begin{subfigure}[t]{0.98\textwidth}
        \centering
        \includegraphics[width=\textwidth]
        {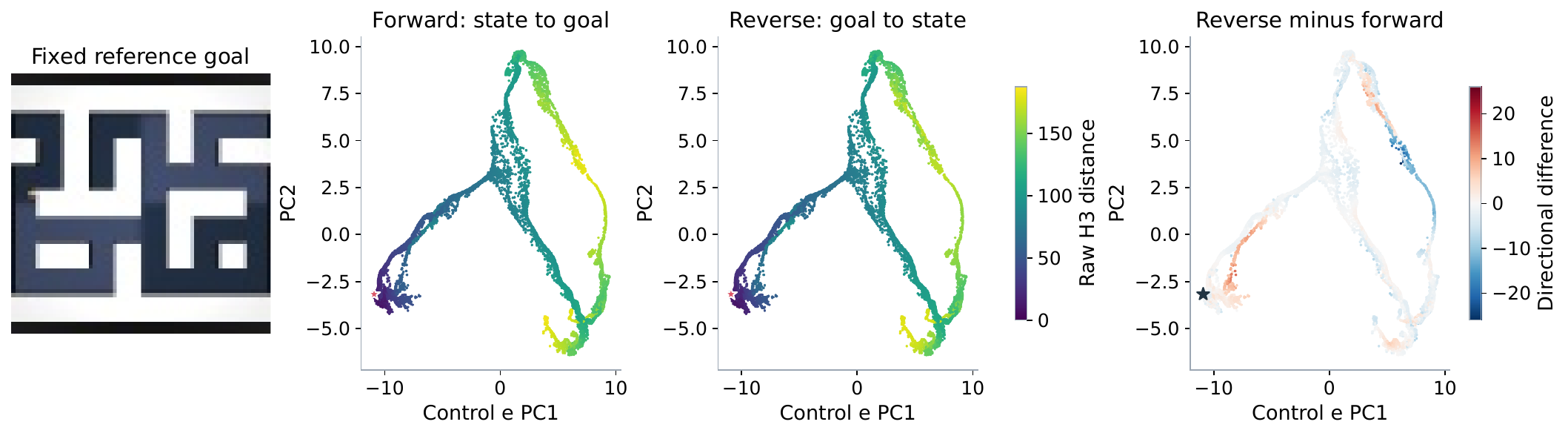}
        \caption{PointMaze}
        \label{fig:app_goal_landscape_pointmaze}
    \end{subfigure}

    \vspace{6pt}

    \begin{subfigure}[t]{0.98\textwidth}
        \centering
        \includegraphics[width=\textwidth]
        {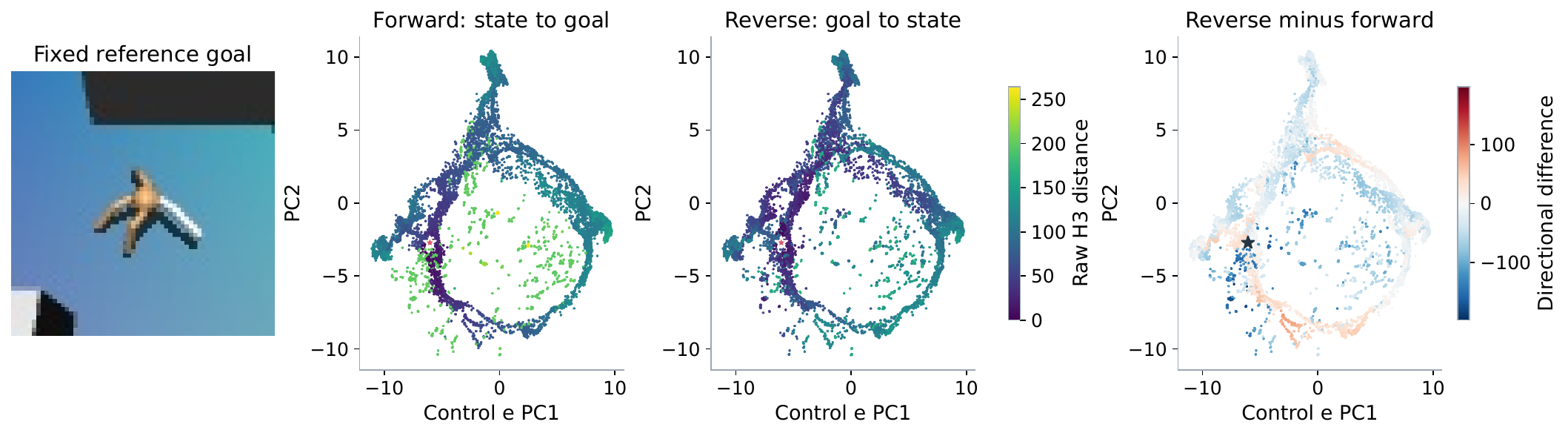}
        \caption{AntMaze-Medium}
        \label{fig:app_goal_landscape_antmaze}
    \end{subfigure}

    \caption{
        \textbf{Goal-relative directional landscapes on PointMaze and
        AntMaze-Medium.}
        For a fixed recorded goal, the panels show the goal observation,
        forward temporal distance from each state to the goal, reverse
        temporal distance from the goal to each state, and the
        reverse--forward difference.
        PCA is used only for visualization.
    }
    \label{fig:app_goal_landscape_navigation}
\end{figure*}

\begin{figure*}[p]
    \centering

    \begin{subfigure}[t]{0.98\textwidth}
        \centering
        \includegraphics[width=\textwidth]
        {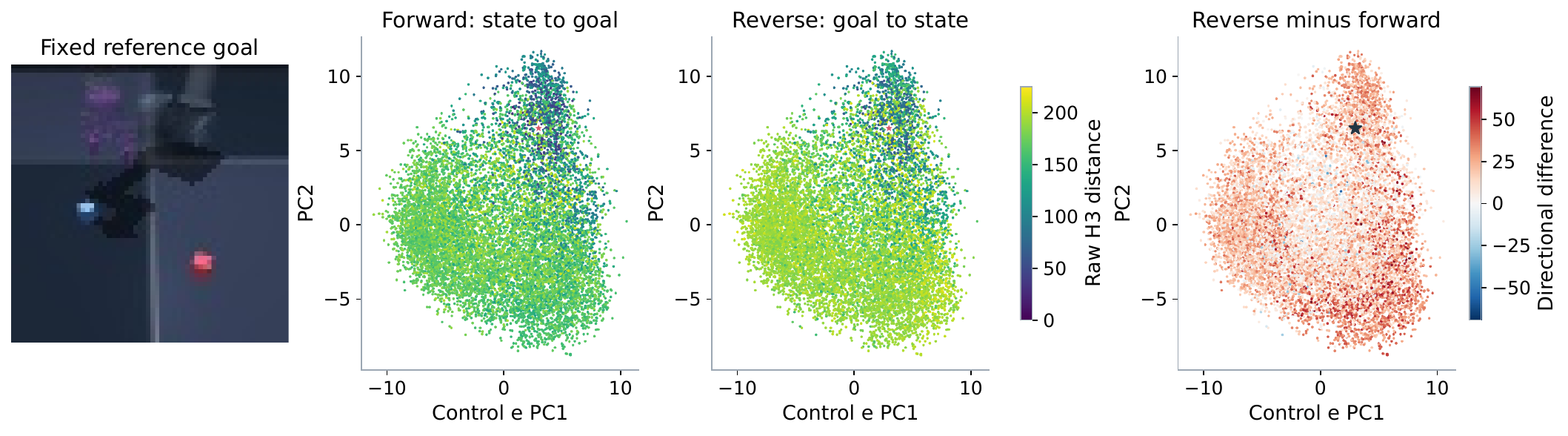}
        \caption{Cube-Double}
        \label{fig:app_goal_landscape_cube_double}
    \end{subfigure}

    \vspace{6pt}

    \begin{subfigure}[t]{0.98\textwidth}
        \centering
        \includegraphics[width=\textwidth]
        {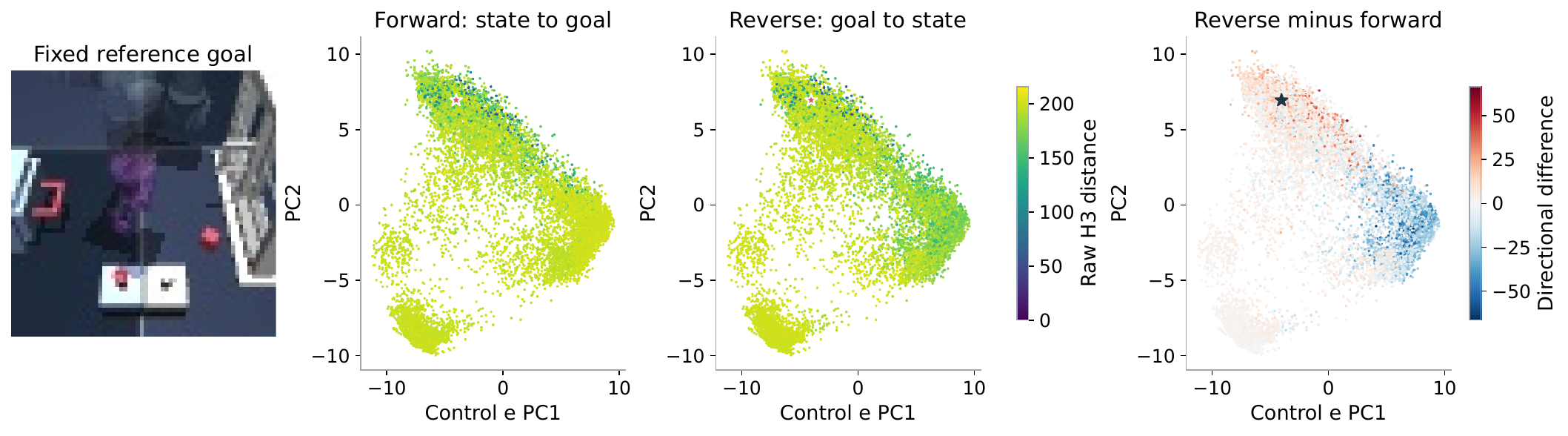}
        \caption{Scene}
        \label{fig:app_goal_landscape_scene}
    \end{subfigure}

    \caption{
        \textbf{Goal-relative directional landscapes on Cube-Double
        and Scene.}
        Forward and reverse temporal distances are evaluated for the
        same fixed goal, and the final panel shows their difference.
        Distances are computed in the full learned representation.
    }
    \label{fig:app_goal_landscape_manipulation}
\end{figure*}

\begin{figure*}[p]
    \centering

    \begin{subfigure}[t]{0.98\textwidth}
        \centering
        \includegraphics[width=\textwidth]
        {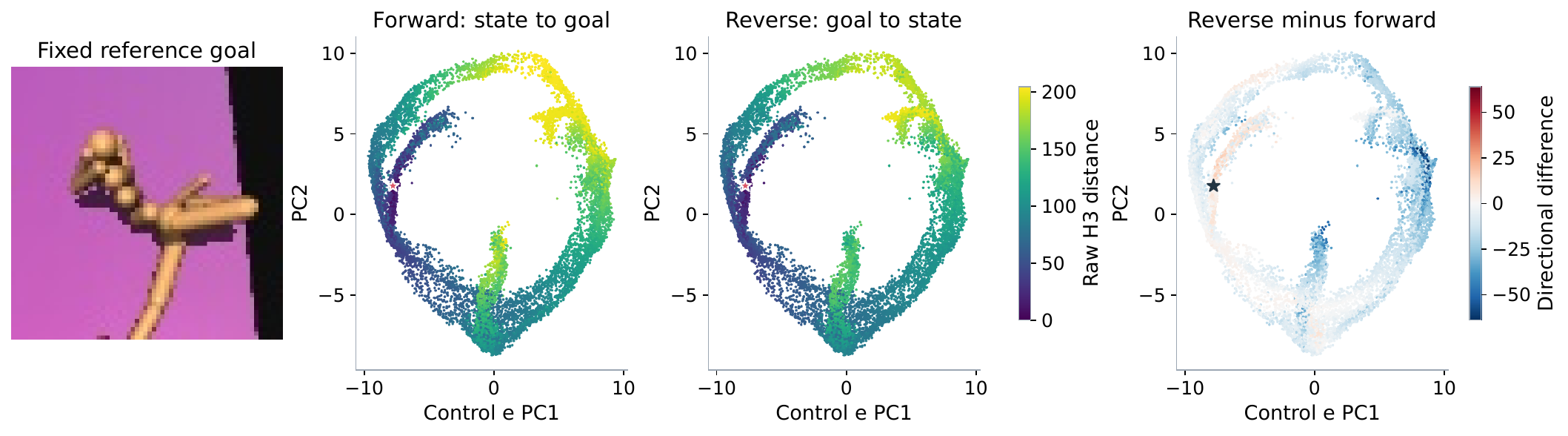}
        \caption{HumanoidMaze-Medium}
        \label{fig:app_goal_landscape_humanoidmaze}
    \end{subfigure}

    \vspace{6pt}

    \begin{subfigure}[t]{0.98\textwidth}
        \centering
        \includegraphics[width=\textwidth]
        {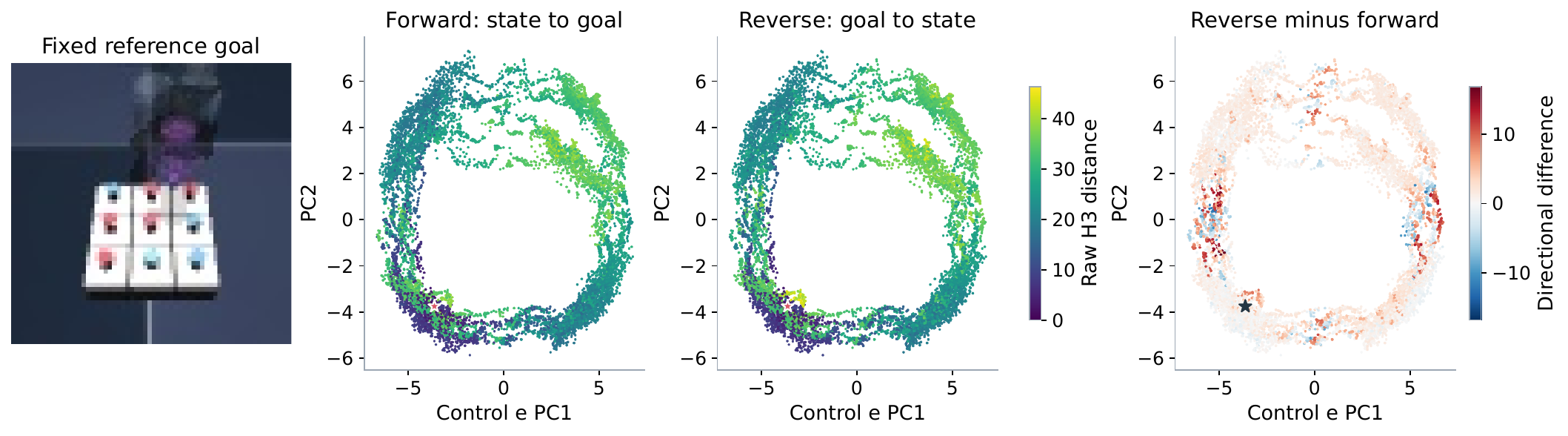}
        \caption{Puzzle-3x3}
        \label{fig:app_goal_landscape_puzzle}
    \end{subfigure}

    \caption{
        \textbf{Goal-relative directional landscapes on
        HumanoidMaze-Medium and Puzzle-3x3.}
        The visualization follows the same construction as
        Figure~\ref{fig:app_goal_landscape_navigation}.
    }
    \label{fig:app_goal_landscape_humanoid_puzzle}
\end{figure*}

The additional tasks further illustrate that directional structure is
not uniform across control problems.
Some environments exhibit pronounced spatial differences between
forward and reverse temporal costs, while others remain closer to
symmetric around the selected reference goal.
The spatial pattern of the reverse--forward difference also varies
substantially across tasks.
Together with the aggregate diagnostics, these visualizations show that
the directed quasimetric provides capacity for order-sensitive
state--goal evaluation without imposing a fixed directional pattern
across environments.

\clearpage
\section{Training Dynamics}
\label{app:training_dynamics}

We visualize the evolution of the learned DTRC representation during
training on the four LeWM tasks.
For each task, we show the control representation $e$, symmetric
coordinates $S(e)$, and directed potentials $A(e)$ at
0, 1,000, 10,000, 20,000, 60,000, and 200,000 training updates.
Points are colored by normalized offline episode time.
These visualizations are intended to illustrate the development of
representation structure during optimization rather than to provide
quantitative convergence criteria.

\begin{figure*}[p]
    \centering

    \begin{subfigure}[t]{0.98\textwidth}
        \centering
        \includegraphics[width=\textwidth]
        {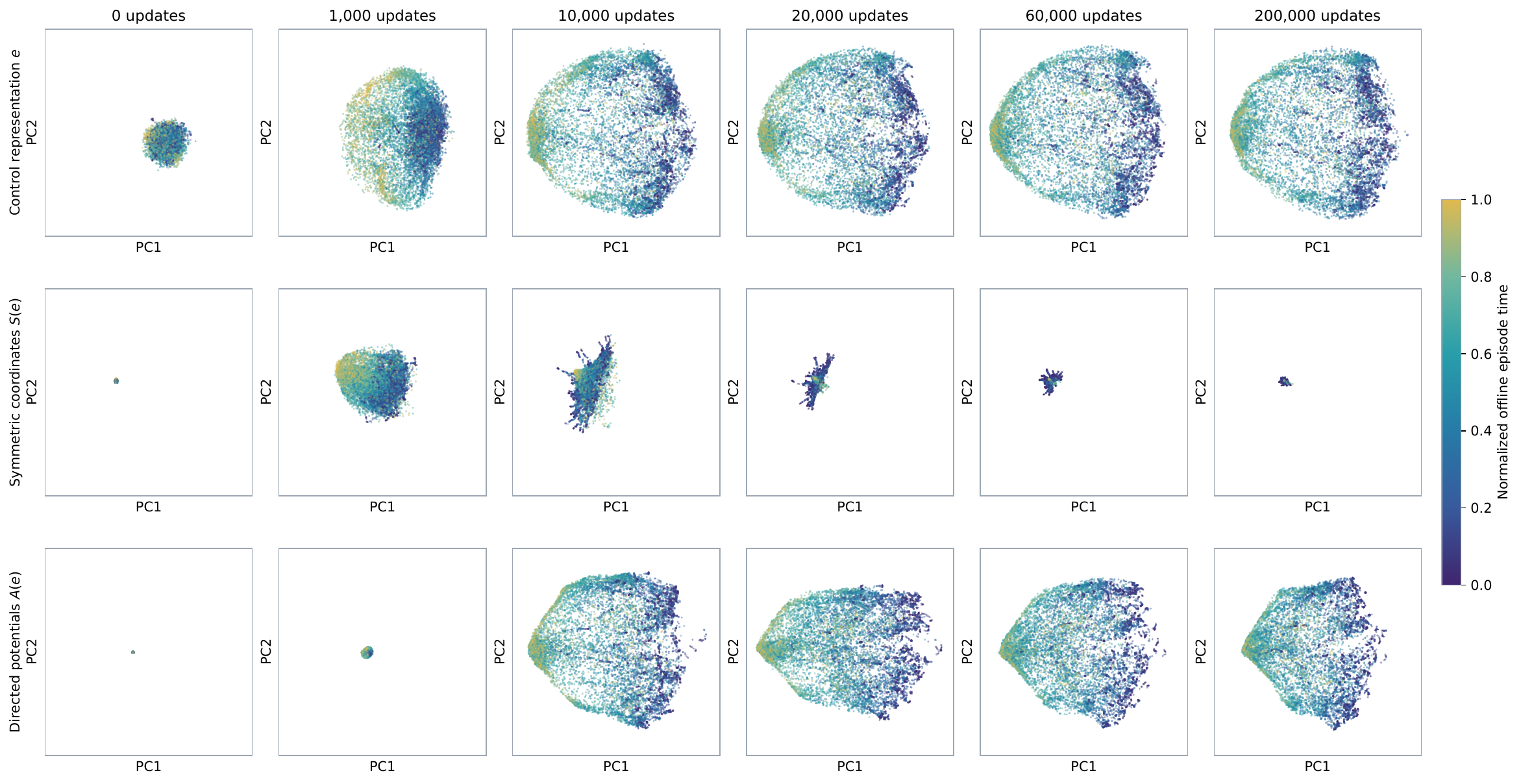}
        \caption{PushT}
        \label{fig:app_evolution_pusht}
    \end{subfigure}

    \vspace{6pt}

    \begin{subfigure}[t]{0.98\textwidth}
        \centering
        \includegraphics[width=\textwidth]
        {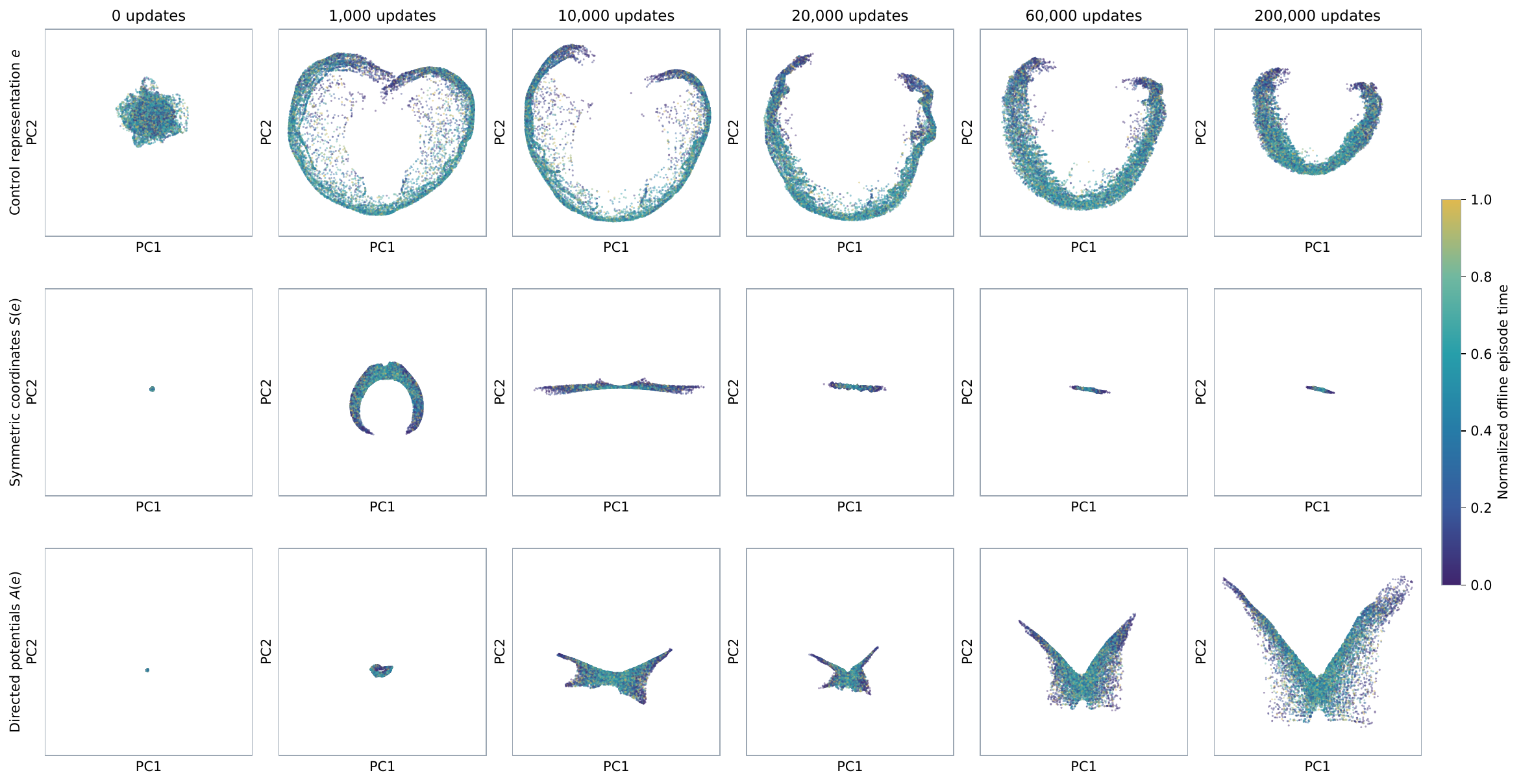}
        \caption{TwoRoom}
        \label{fig:app_evolution_tworoom}
    \end{subfigure}

    \caption{
        \textbf{Evolution of the learned representation on PushT and
        TwoRoom.}
        Columns correspond to training checkpoints and rows show the
        control representation $e$, symmetric coordinates $S(e)$, and
        directed potentials $A(e)$.
        Colors indicate normalized offline episode time.
    }
    \label{fig:app_evolution_pusht_tworoom}
\end{figure*}

\begin{figure*}[p]
    \centering

    \begin{subfigure}[t]{0.98\textwidth}
        \centering
        \includegraphics[width=\textwidth]
        {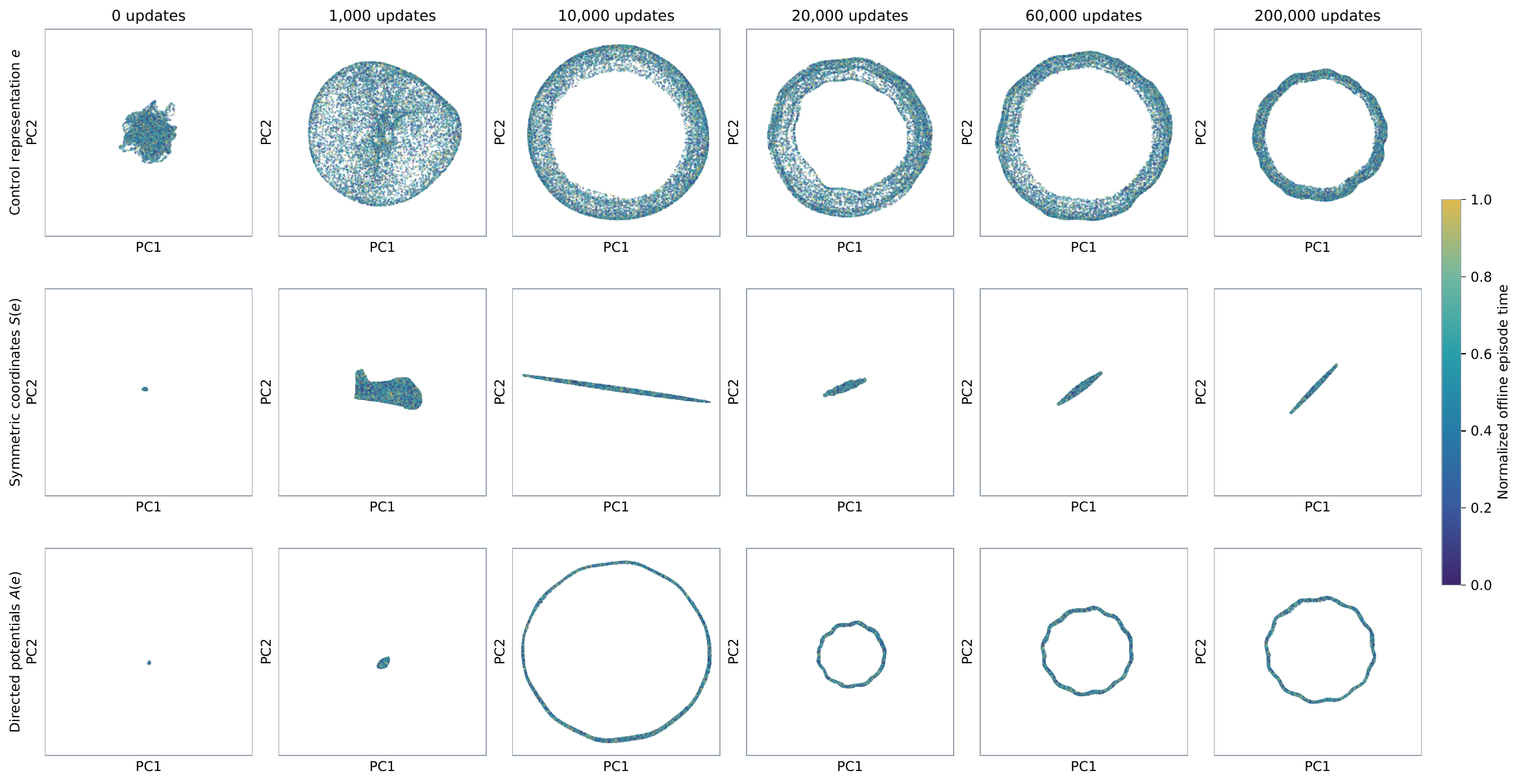}
        \caption{Reacher}
        \label{fig:app_evolution_reacher}
    \end{subfigure}

    \vspace{6pt}

    \begin{subfigure}[t]{0.98\textwidth}
        \centering
        \includegraphics[width=\textwidth]
        {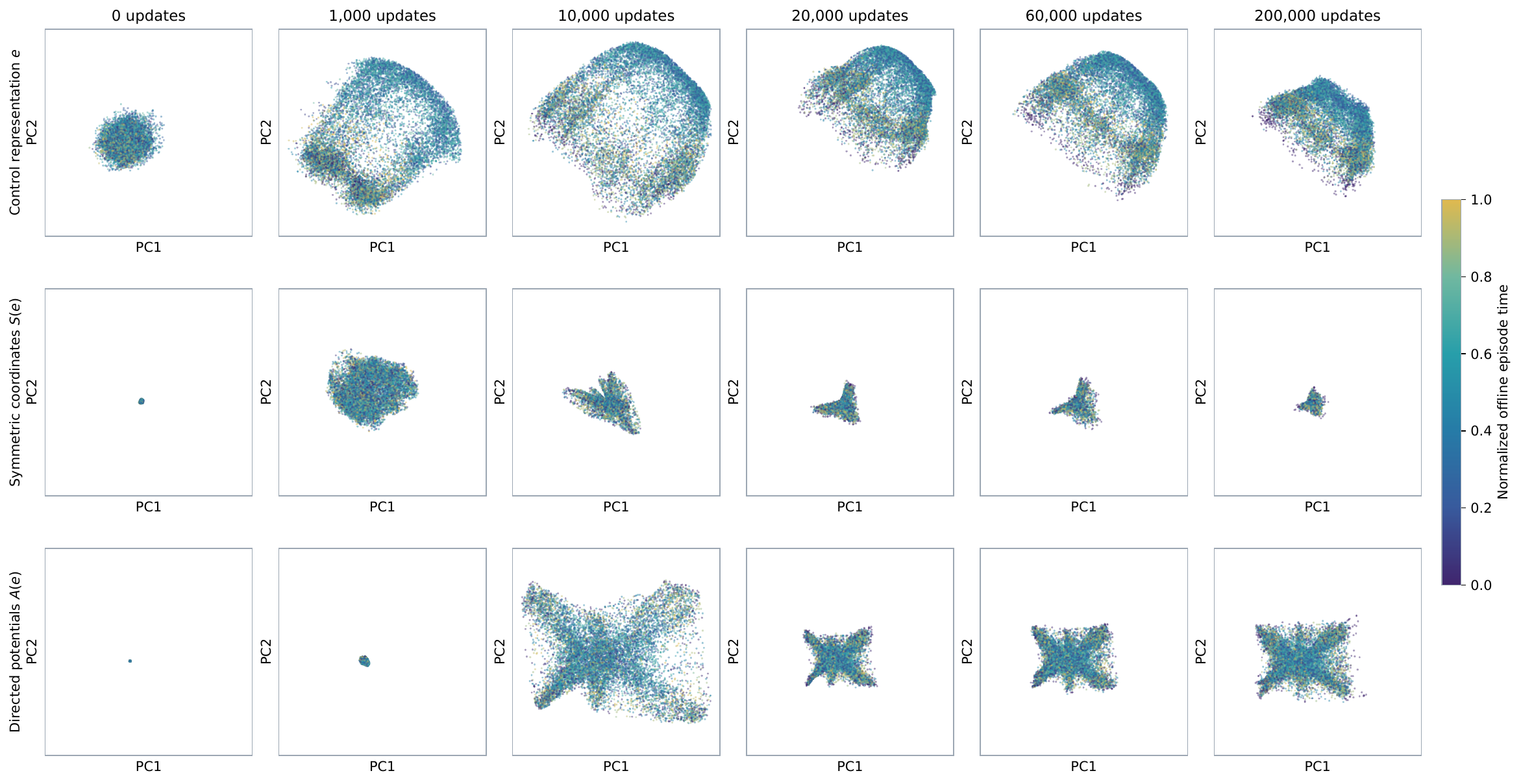}
        \caption{OGBench-Cube}
        \label{fig:app_evolution_cube}
    \end{subfigure}

    \caption{
        \textbf{Evolution of the learned representation on Reacher and
        OGBench-Cube.}
        Columns correspond to training checkpoints and rows show the
        control representation $e$, symmetric coordinates $S(e)$, and
        directed potentials $A(e)$.
        Colors indicate normalized offline episode time.
    }
    \label{fig:app_evolution_reacher_cube}
\end{figure*}

The learned structures emerge progressively during optimization and
differ substantially across tasks.
On TwoRoom, the control representation develops a curved geometry while
the directed potentials expand into a distinct order-sensitive
structure.
Reacher develops an approximately cyclic control representation and
directed-potential geometry.
PushT and Cube exhibit different non-cyclic organizations, consistent
with the task-dependent final representations reported in
Section~\ref{app:representation_spaces}.
These visualizations indicate that DTRC does not converge toward a
single common projected geometry across environments.

\paragraph{Additional OGBench tasks.}
We further visualize representation evolution on the six additional
OGBench tasks using the same checkpoints and PCA protocol.
These results show that task-specific geometric structure also emerges
progressively beyond the four LeWM tasks.

\begin{figure*}[p]
    \centering

    \begin{subfigure}[t]{0.98\textwidth}
        \centering
        \includegraphics[width=\textwidth]
        {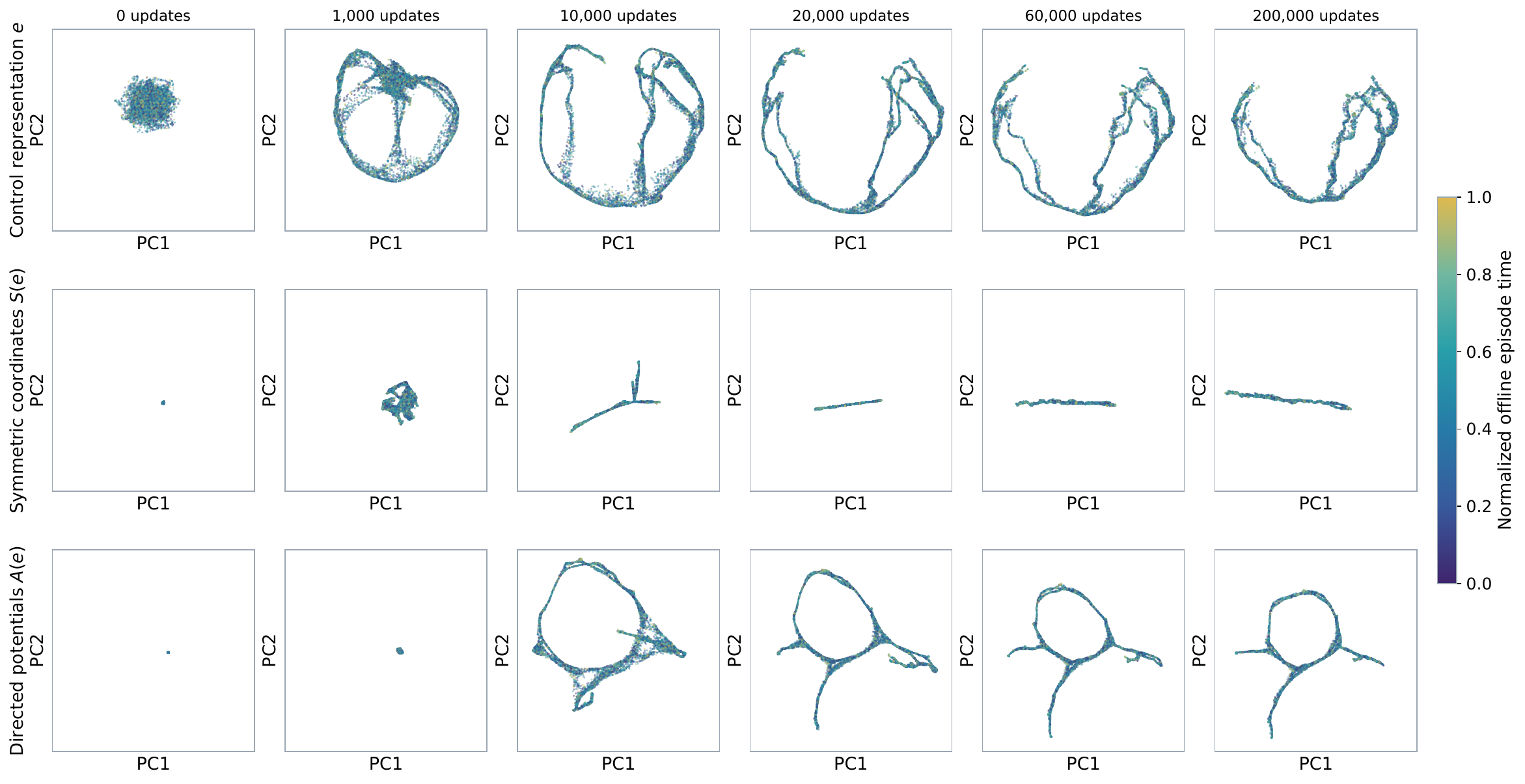}
        \caption{PointMaze}
        \label{fig:app_evolution_pointmaze}
    \end{subfigure}

    \vspace{6pt}

    \begin{subfigure}[t]{0.98\textwidth}
        \centering
        \includegraphics[width=\textwidth]
        {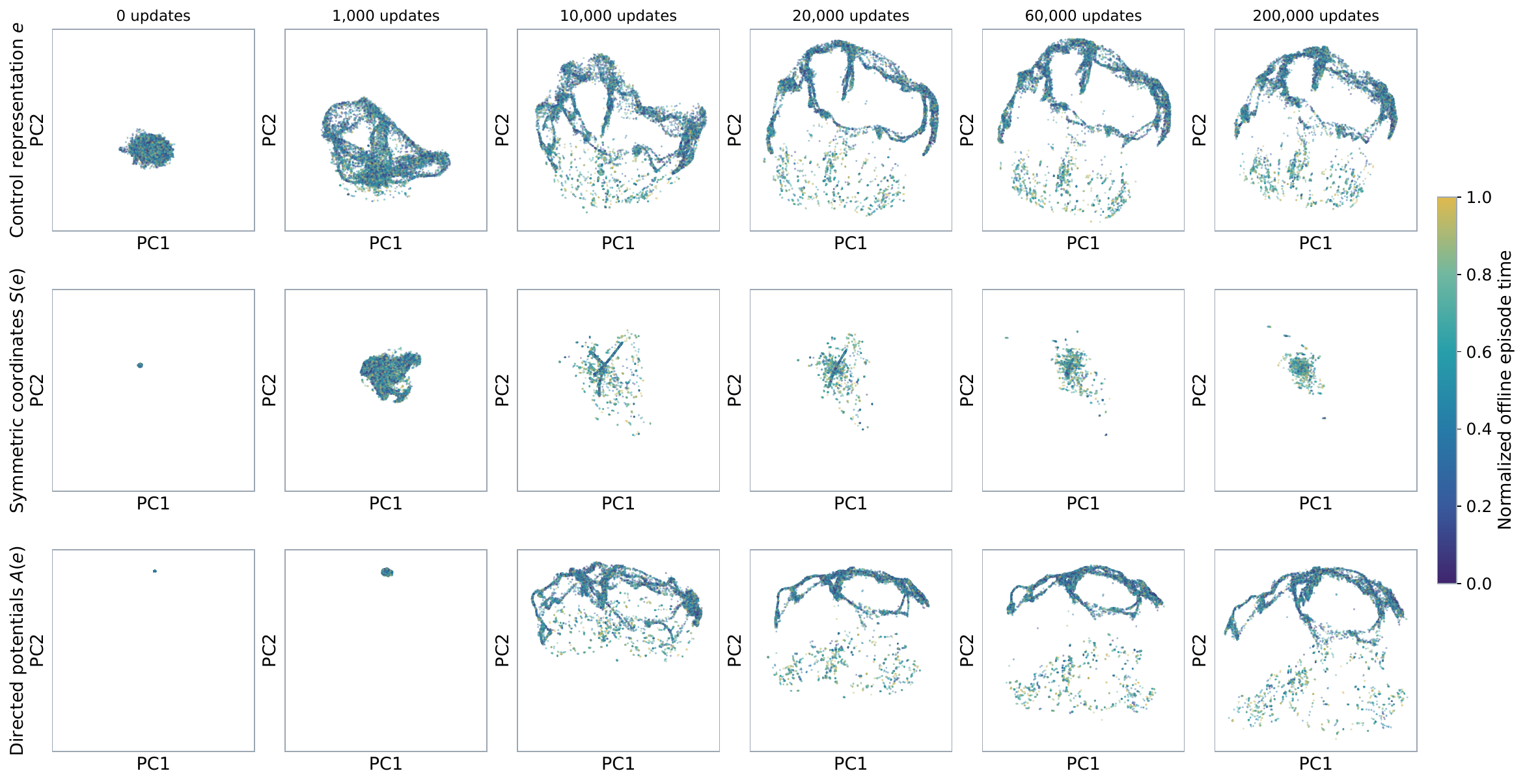}
        \caption{AntMaze-Medium}
        \label{fig:app_evolution_antmaze}
    \end{subfigure}

    \caption{
        \textbf{Evolution of the learned representation on PointMaze
        and AntMaze-Medium.}
        Columns correspond to training checkpoints and rows show the
        control representation $e$, symmetric coordinates $S(e)$, and
        directed potentials $A(e)$.
        Colors indicate normalized offline episode time.
        PCA is fitted independently at each checkpoint, so orientation
        and scale are not directly comparable across columns.
    }
    \label{fig:app_evolution_navigation}
\end{figure*}

\begin{figure*}[p]
    \centering

    \begin{subfigure}[t]{0.98\textwidth}
        \centering
        \includegraphics[width=\textwidth]
        {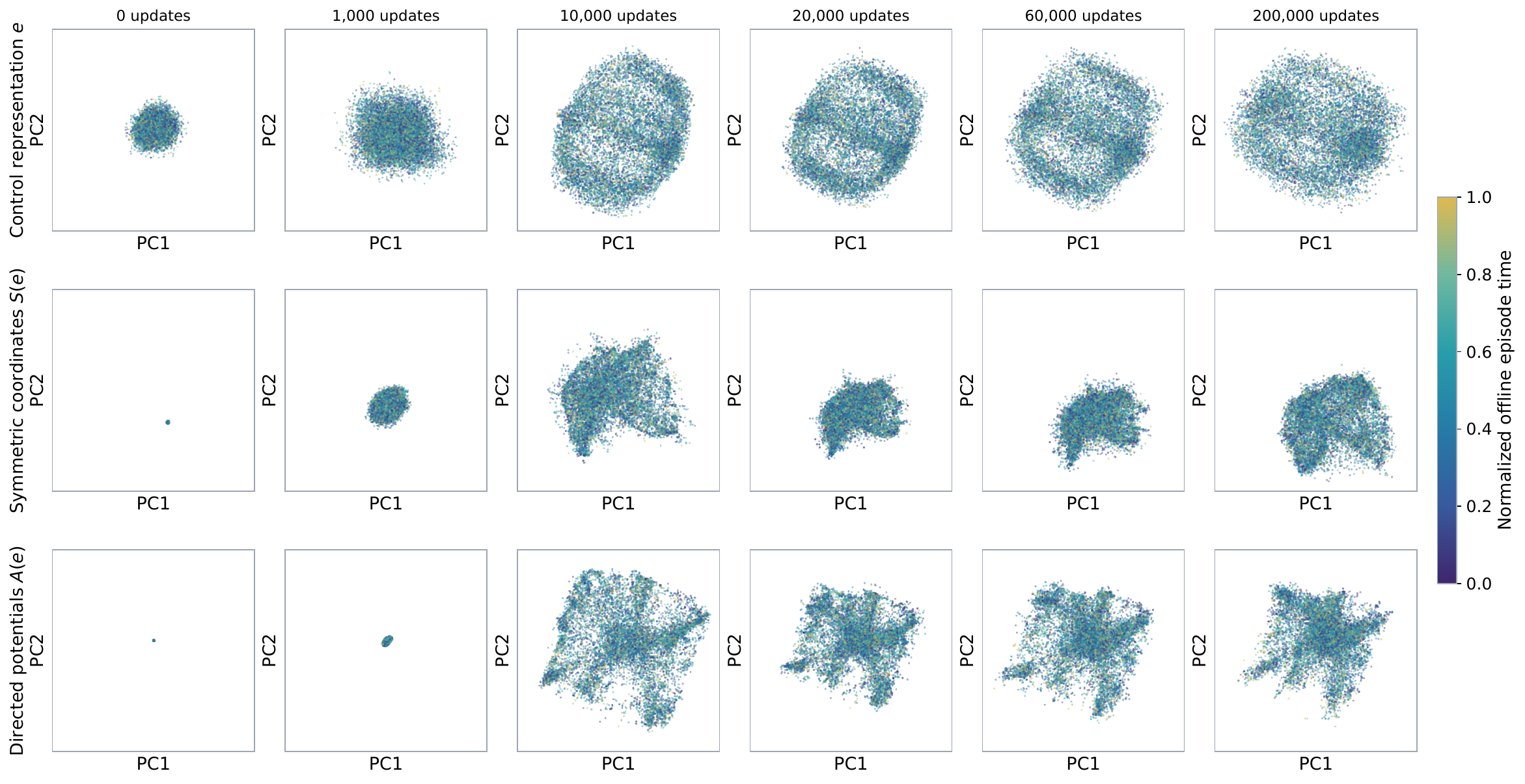}
        \caption{Cube-Double}
        \label{fig:app_evolution_cube_double}
    \end{subfigure}

    \vspace{6pt}

    \begin{subfigure}[t]{0.98\textwidth}
        \centering
        \includegraphics[width=\textwidth]
        {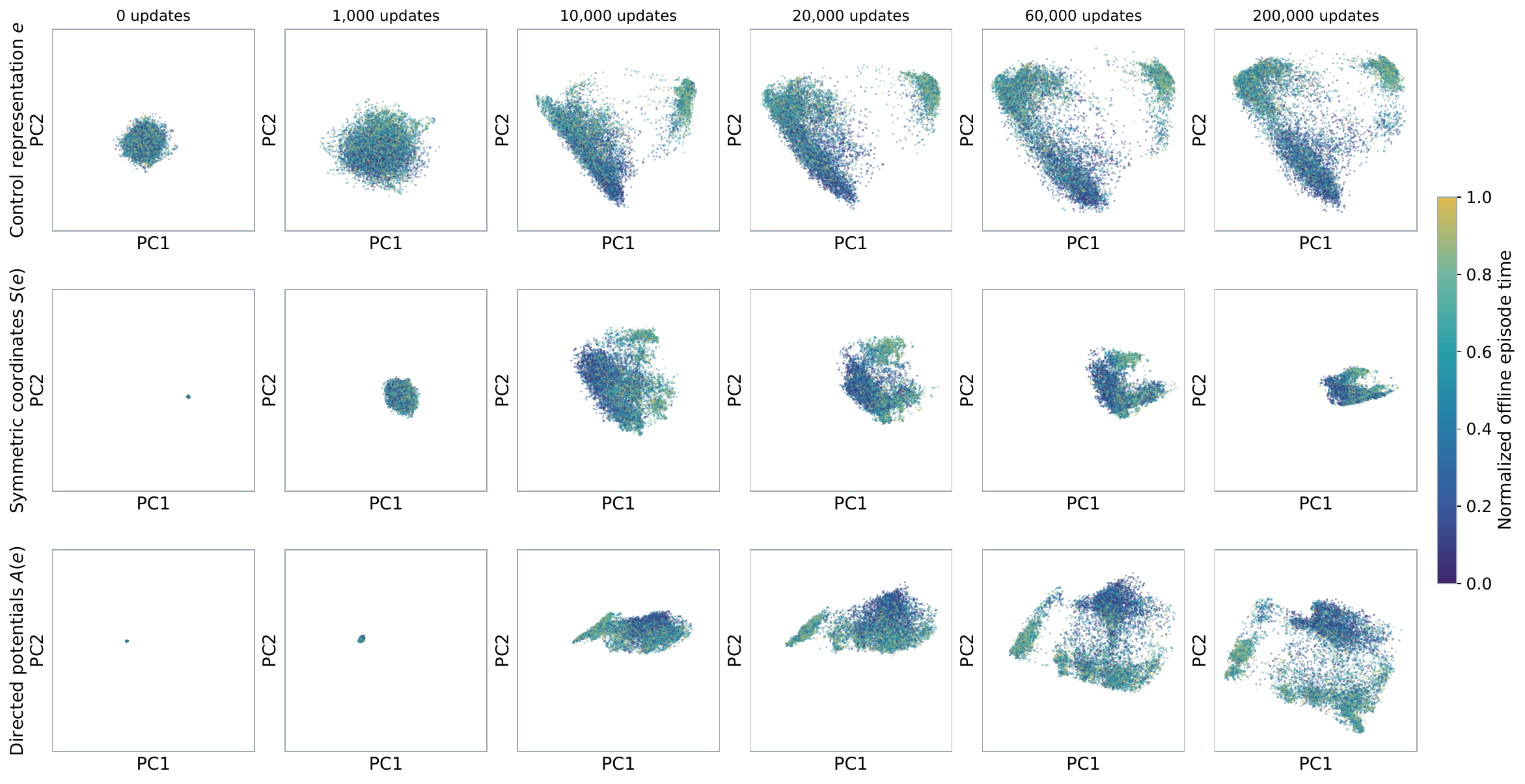}
        \caption{Scene}
        \label{fig:app_evolution_scene}
    \end{subfigure}

    \caption{
        \textbf{Evolution of the learned representation on Cube-Double
        and Scene.}
        The visualization follows the same checkpoint and PCA protocol
        as Figure~\ref{fig:app_evolution_navigation}.
    }
    \label{fig:app_evolution_manipulation}
\end{figure*}

\begin{figure*}[p]
    \centering

    \begin{subfigure}[t]{0.98\textwidth}
        \centering
        \includegraphics[width=\textwidth]
        {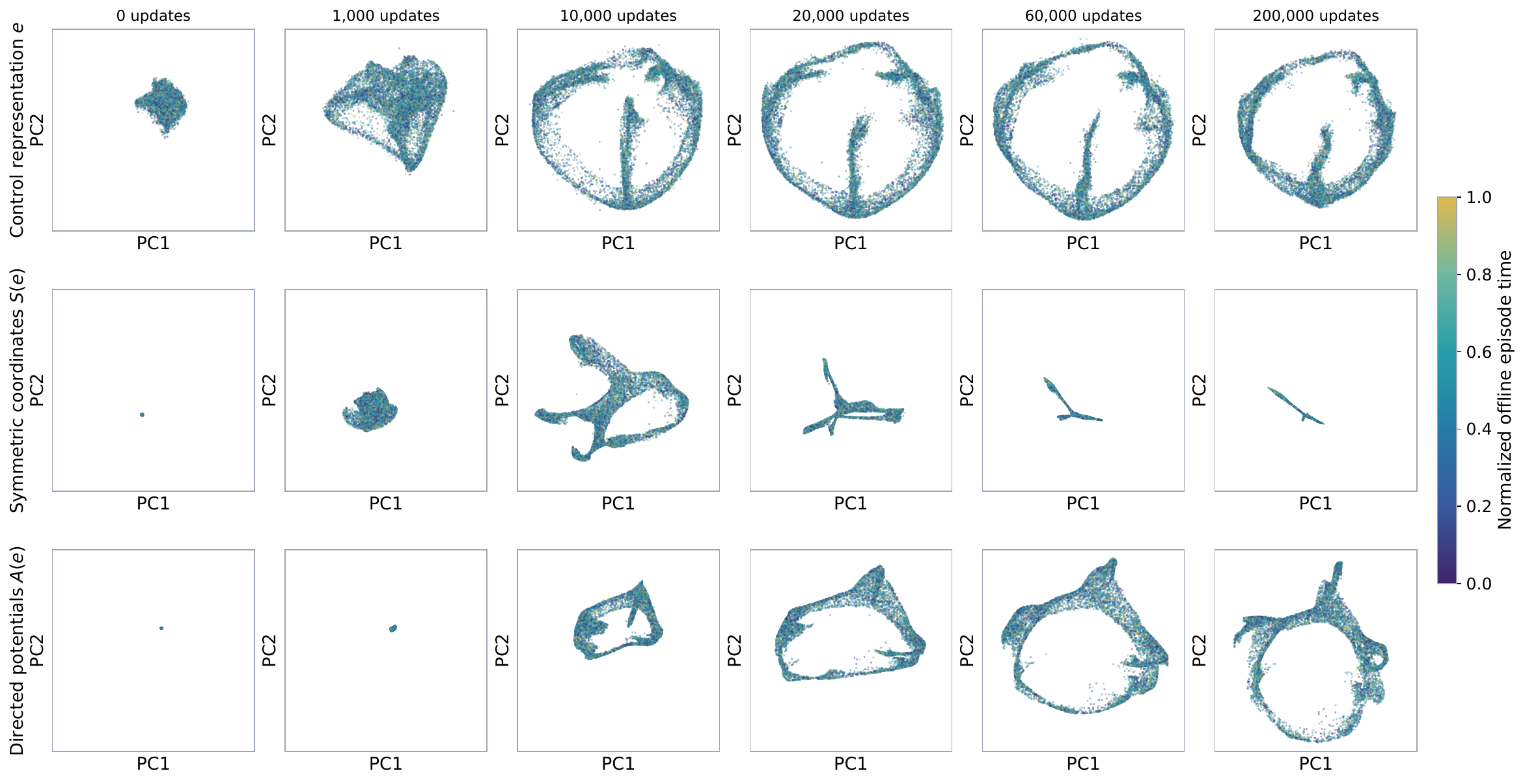}
        \caption{HumanoidMaze-Medium}
        \label{fig:app_evolution_humanoidmaze}
    \end{subfigure}

    \vspace{6pt}

    \begin{subfigure}[t]{0.98\textwidth}
        \centering
        \includegraphics[width=\textwidth]
        {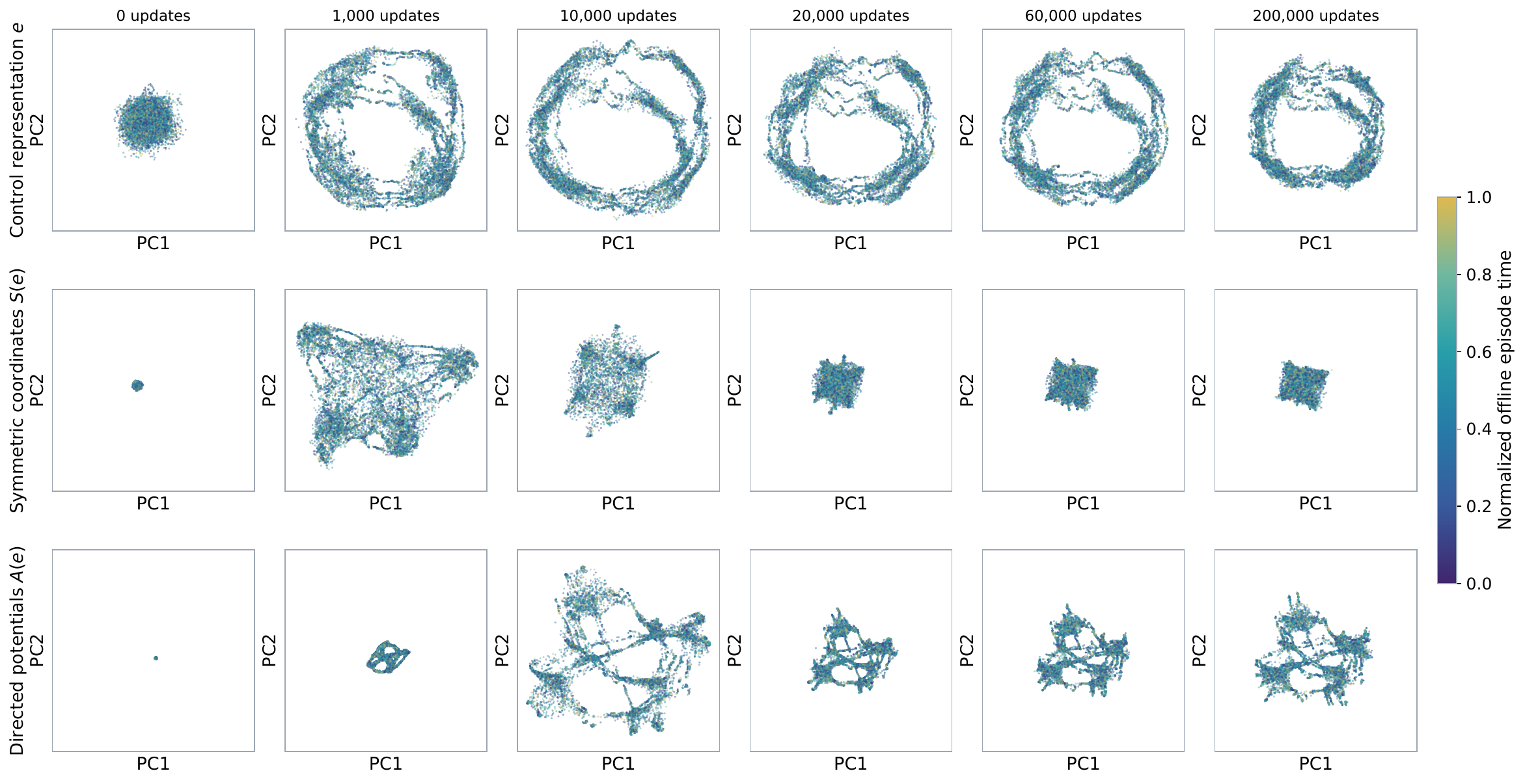}
        \caption{Puzzle-3x3}
        \label{fig:app_evolution_puzzle}
    \end{subfigure}

    \caption{
        \textbf{Evolution of the learned representation on
        HumanoidMaze-Medium and Puzzle-3x3.}
        Columns correspond to training checkpoints and rows show
        $e$, $S(e)$, and $A(e)$.
        PCA is fitted independently at each checkpoint.
    }
    \label{fig:app_evolution_humanoid_puzzle}
\end{figure*}

Across the additional tasks, representation structure develops
progressively during optimization while remaining strongly
task-dependent.
PointMaze and AntMaze-Medium develop distinct nonlinear organizations
in the control representation and directed potentials.
Cube-Double and Puzzle-3x3 form different structured manifolds, while
Scene and HumanoidMaze-Medium exhibit more diffuse projected
organizations.
The symmetric and directed heads also evolve differently across tasks.
These visualizations are therefore consistent with the final
task-dependent geometries reported in the representation analyses,
rather than with convergence toward a single common projected form.

\end{document}